%% file: main-arxiv.tex
\documentclass{article} %
\PassOptionsToPackage{table}{xcolor} %
\usepackage{arxiv,times}

\input{math_commands.tex}

\usepackage{xcolor}             %

\usepackage{amsmath}
\usepackage{amssymb}

\usepackage{algorithm}
\usepackage{algpseudocode}      %

\usepackage{booktabs}           %
\usepackage{multirow}           %
\usepackage{array}
\usepackage{float}              %

\usepackage{graphicx}           %
\usepackage{adjustbox}           %
\usepackage{subcaption}         %

\usepackage[most]{tcolorbox}
\usepackage{enumitem}
\usepackage{wrapfig}
\usepackage{url}
\usepackage[hidelinks]{hyperref}
\usepackage{soul}
\sethlcolor{yellow}

\providecommand{\na}{\textcolor{gray}{--}}   %
\providecommand{\na}{\textcolor{gray}{--}}

\title{Beyond 4 Bits: Data-Centric Post-Training Quantization and Recovery}

\prepfinalcopy
\author{%
  \textbf{Rhea Sanjay Sukthanker}$^{1}$\quad
  \textbf{Riccardo Grazzi}$^{2}$\quad
  \textbf{James Hensman}$^{2}$\quad
  \textbf{Pashmina Cameron}$^{1}$\\[6pt]
  {\normalsize $^{1}$Applied Sciences Group (ASG), Microsoft\qquad
  $^{2}$Microsoft Research}\\[6pt]
  {\small\texttt{rsukthanker@microsoft.com}\quad
  \texttt{riccardo.grazzi.4@gmail.com}}\\
  {\small\texttt{jameshensman@microsoft.com}\quad
  \texttt{pcameron@microsoft.com}}%
}

\begin{document}

\maketitle

\input{sections/abstract}
\input{sections/introduction}
\input{sections/related_work}

\input{sections/methodology}
\input{sections/experiments_v2}
\input{sections/best_practices}

\section*{Use of AI Tools}
All ideas presented in this paper are original to the authors or build upon the
related work cited herein. AI-based tools were used only in an assistive
capacity: to sharpen the clarity of the authors' initial drafts, to help typeset
the paper in \LaTeX{}, to rephrase sections, to improve the presentation of figures and tables, to
correct grammatical errors and make sentences more salient, and to act as a
coding assistant that implements and documents the authors' ideas and results. All technical content,
claims, and conclusions were conceived, verified, and approved by the authors.

\section*{Acknowledgments}
We are deeply grateful to Hari VK and Andrey Rybalchenko (Microsoft Research, Cambridge) for their
extensive work to make application of incoherence rotations automatic and robust. This work allowed us to reliably test the effect of incoherent rotations on
a large number of dense and mixture-of-experts transformer models with different architectures. The work underpins our rotated scalar baselines and the
incoherence-rotated frame in which our vector quantizers operate on the MoE models, and it was
essential for extending this study from dense models to Qwen3-30B-A3B and OLMoE-1B-7B.

\bibliographystyle{preprint}
\bibliography{references}
\appendix
\input{sections/appendix_v2}
\end{document}

%% file: math_commands.tex
\usepackage{amsmath,amsfonts,bm}

\def\eqref#1{equation~\ref{#1}}

\def\1{\bm{1}}

\DeclareMathAlphabet{\mathsfit}{\encodingdefault}{\sfdefault}{m}{sl}
\SetMathAlphabet{\mathsfit}{bold}{\encodingdefault}{\sfdefault}{bx}{n}

%% file: sections/abstract.tex
\begin{abstract}
Post-training quantization (PTQ) aims at compressing the weights of a large
trained model using a limited compute and data budget. Down to 4 bits,
state-of-the-art PTQ methods are near-lossless; below this limit, the tested scalar PTQ
configurations collapse to near-random accuracy and model quality becomes acutely sensitive to
choices such as the quantization representation and the calibration data. At such
low bitwidths, recovery fine-tuning after quantization becomes crucial, and its
effectiveness depends heavily on these same choices. Yet there is little
systematic evidence on how the quantization method, the quality and task
alignment of calibration data, and the design of the recovery procedure interact
to determine the best final 2-bit model. We present a systematic study of 2-bit
quantization and recovery fine-tuning across multiple model families, scales, and
data distributions, and distill our findings into a practical, architecture-dependent set of best
practices for efficiently obtaining an accurate 2-bit model. This recipe, which combines a vector codebook representation, a small task-aligned calibration mixture, and a short codebook-only distillation stage, restores $63$--{$68$}$\%$ average accuracy across seven question-answering tasks at 2 bits (a $71$--$73\%$ \textsc{bf16} ceiling), versus the ${\sim}29\%$ random-accuracy floor of {the evaluated} scalar PTQ {pipelines},
at roughly half the compute budget of {a full-model QAT baseline we studied. Given the same calibration and recovery data, EfficientQAT, a trained scalar quantization baseline, avoids the collapse. However, at a matched rate it trails the vector recipe on generation on every model ($2.8$--$10.9$ AvgGen; within noise on Qwen3-MoE) and on question answering on three out of the four studied models ($2.7$--$7.8$ AvgQA). We also find that the best recovery data depends on the calibration data, an interaction confirmed on a held-out model. We provide a systematic, architecture-aware map of the 2-bit design space, showing that representation, calibration data, and recovery data must be chosen jointly.}
\end{abstract}

%% file: sections/introduction.tex
\setlength{\parfillskip}{0pt plus 0.8\textwidth}
\vspace{-6mm}
\section{Introduction}

\begin{figure*}[t]
\centering
\begin{subfigure}[t]{0.48\textwidth}
  \centering
  \parbox[c][0.19\textheight][c]{\linewidth}{%
    \centering
    \includegraphics[
      width=\linewidth,
      height=0.19\textheight,
      keepaspectratio
    ]{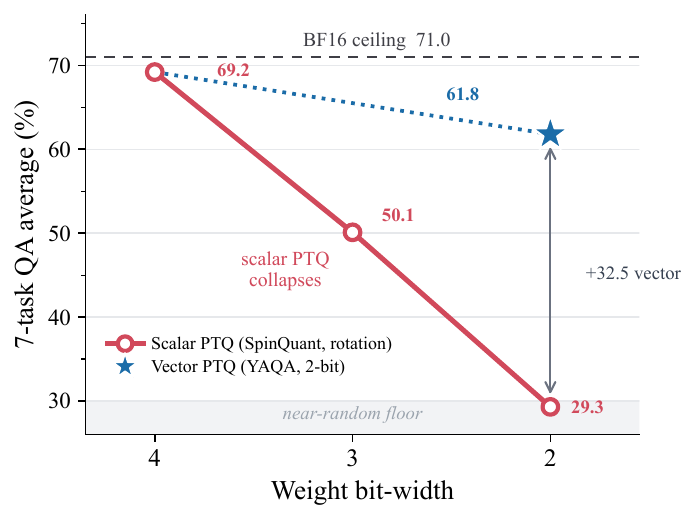}\par}
  \caption{The 2-bit cliff.}
  \label{fig:teaser-cliff}
\end{subfigure}
\hfill
\begin{subfigure}[t]{0.48\textwidth}
  \centering
  \parbox[c][0.19\textheight][c]{\linewidth}{%
    \centering
    \includegraphics[
      width=\linewidth,
      height=0.19\textheight,
      keepaspectratio
    ]{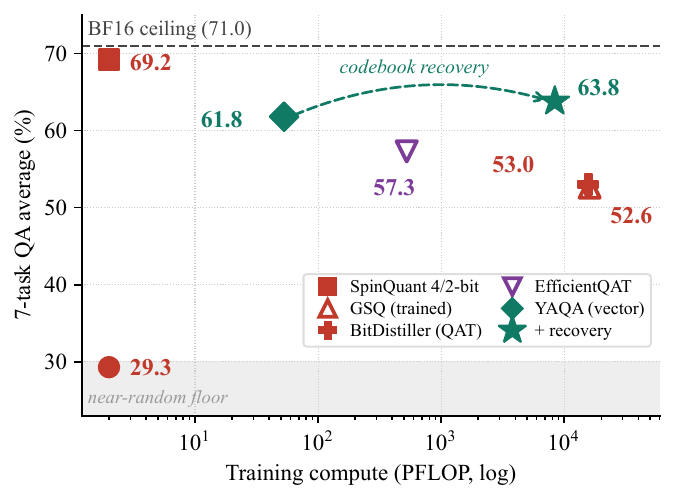}\par}
  \caption{Trained baselines underperform vector PTQ.}
  \label{fig:teaser-recovery}
\end{subfigure}

\caption{\textbf{The 2-bit cliff and how we mitigate it.}
\textbf{(a)} On Phi-4-mini (7-task QA average; \textsc{bf16} $71.0$), scalar
PTQ (SpinQuant{, web calibration}) falls with bitwidth, whereas a 2-bit vector representation (YAQA{, math+web $0.6/0.4$ calibration, no recovery}) stays far
higher ($61.8$).
\textbf{(b)} At 2 bits, every scalar method we evaluate trails the vector model
($61.8$), including trained GSQ ($52.6$) and BitDistiller ($53.0$){; EfficientQAT reaches $57.3$ at ${\sim}526$ PFLOP (Table~\ref{tab:rev-eqat})}; a short
codebook-only recovery on the 2-bit YAQA model (PTQ$+$RFT) reaches $63.8$ at
roughly half their compute ($8{,}349$ vs $15{,}717$--$16{,}136$ PFLOP). {Persistent memory on Phi-4-mini is $1.0$\,GB for PTQ, $2.2$\,GB for GSQ, $9.0$\,GB for recovery, and $53.8$\,GB for full QAT (Table~\ref{tab:memory-tradeoff}). Per-method compute metrics are tabulated} in Table~\ref{tab:app-compute-all}.}
\label{fig:teaser}
\end{figure*}

Quantization is a preferred route to serving Large Language Models (LLMs) under
constrained memory and latency budgets \citep{frantar2023gptq,lin2023awq}, and at 4 bits (W4A16) it is nearly free with state-of-the-art methods such as QuaRot \citep{ashkboos2024quarot} and SpinQuant \citep{liu2024spinquant}. Quantization maps a continuous range of weight values onto a small signed discrete grid ($q \in \{-2, -1, 0, 1\}$ in the symmetric 2-bit case), with the goal of minimizing the error this discretization introduces \citep{gholami2022survey}. Current post-training quantization (PTQ) methods typically compress a billion-parameter pretrained model with a few forward passes on a small calibration set on a single GPU
\citep{frantar2023gptq}, shrinking its memory footprint by about $4\times$ at 4 bits (relative to \textsc{bf16}). This
preserves accuracy to within a fraction of a point of the full-precision model,
all without retraining \citep{frantar2023gptq,lin2023awq,shao2023omniquant}. Pushing to 3 or 2 bits would roughly halve the footprint
again and is the natural next target. It is also where the standard scalar PTQ methods break \citep{egiazarian2024aqlm,chee2023quip,vptq2024}, as illustrated in Figure~\ref{fig:teaser}a.

Scalar post-training quantization (PTQ) exhibits a sharp performance cliff below 4 bits (Figure~\ref{fig:teaser}a). Methods that are nearly lossless at 4 bits, including QuaRot \citep{ashkboos2024quarot} and SpinQuant \citep{liu2024spinquant}, degrade to near-random accuracy at 2 bits. This failure is particularly severe for reasoning models and on generative tasks, which additionally exhibit pronounced overthinking and repetitive looping at lower precisions \citep{lotfi2026quantized}. Existing approaches attempt to mitigate this collapse by trading additional computation for model quality. One line of work retains the PTQ framework but replaces scalar quantization with richer, codebook-based representations \citep{vanbaalen2024gptvq,vptq2024,egiazarian2024aqlm,tseng2024quip}. Another replaces calibration-only PTQ with learning-based optimization, such as quantization-aware training (QAT) \citep{liu2023llmqat} or learned scalar grids \citep{dadgarnia2026gsq}. Learning-based methods, however, require substantially more data and computation \citep{liu2023llmqat,du2024bitdistiller,dadgarnia2026gsq}. Even after incurring these costs, 2-bit models continue to lag far behind their 4-bit counterparts, and no existing approach fully closes the gap.

We argue that a \emph{trinity} of choices governs the quality of a 2-bit model,
choices that practitioners usually treat as implementation details and mention
only in passing: the quantization \emph{representation}, the \emph{calibration data {type used}}, and the \emph{recovery fine-tuning procedure}. At 4 bits, these three choices hardly matter. Below 4 bits, they become decisive and are strongly coupled{: the best recovery data flips with the calibration data (Section~\ref{sec:exp-recovery}, Figure~\ref{fig:rev-interaction})}. Hence studying any \textit{single choice} in isolation gives a misleading picture. As a result, no principled recipe exists for building the best possible 2-bit model; this work aims at bridging this gap.

We perform a systematic study of 2-bit PTQ and recovery methods across multiple model
families and scales, and distill it into a concrete recipe. Moving from a scalar
to a vector representation recovers most of the collapsed accuracy at a fixed
compute budget. Aligning the calibration data with the target {tasks} recovers more{, primarily on generation}. A lightweight
recovery stage that updates only the codebook parameters with a distillation loss
\citep{hinton2015distilling} pushes performance even closer to 4-bit quality at a
fraction of the compute of QAT and learning-based methods
\citep{du2024bitdistiller,dadgarnia2026gsq}
(Figure~\ref{fig:teaser}b). Together these choices yield a 2-bit model {that retains 90\% of \textsc{bf16} average question-answering accuracy (63.8 vs.\ 71.0 on Phi-4-mini)} while using roughly half the training compute of the {full-model QAT baseline (BitDistiller)}. Our study spans five models across dense and MoE architectures: three primary models (Phi-4-mini, Llama-3.1-8B, Qwen3-MoE), with OLMoE and Qwen3-8B as auxiliary checks. {Our primary goal is to propose} an empirically grounded, {actionable} recipe that gets the most out of existing quantizers. Our key contributions are:

\begin{itemize}
    \item \textbf{Representation drives the 2-bit cliff.}
    At a fixed compute budget, the {studied} scalar PTQ methods collapse to the ${\sim}29\%$
    random-accuracy floor in every family, while a vector codebook restores a
    usable model (average question-answering accuracy ${\approx}\,63$--{$66$}$\%$).
    {Trained scalar methods such as EfficientQAT, given the same calibration and recovery data, do not collapse, but at a matched rate EfficientQAT trails the vector codebook on generation by up to $10.9$ points of average generation accuracy.}

    \item \textbf{Generation retention is strongly model-dependent.} At 2-bit vector
    quantization, QA retention is ${\sim}90\%$ everywhere, but per-model {average} generation
    retention spans $51\%$ (Phi-4-mini)
    to $62\%$ (Llama) to {$100\%$} (Qwen3-MoE). Hence, architecture and model scale, more than the quantizer, shape whether a 2-bit model can still reason and code.

    \item \textbf{Calibration and recovery are decisive {and coupled}.} {Calibration composition moves generation far more than question answering, and the best recovery data depends on the calibration data (Figure~\ref{fig:rev-interaction}).} A codebook-only distillation stage further narrows the generation gap to 4-bit at a fraction of QAT compute
    (Figure~\ref{fig:teaser}b).

\end{itemize}
{Our core findings distill} an actionable recipe for {developing and evaluating the best possible} 2-bit models:
quantize with a vector representation, calibrate on a small {mixture matched to the target tasks}, and recover with a short codebook-only
distillation stage {whose data is chosen together with the calibration data}. It targets the limited compute and memory regime where a vector
representation is deployable and only a small accuracy loss is acceptable, and its
gains hold across {the} model families and scales {we study}. We release our code and, upon acceptance, will also release the Hessian mixtures and checkpoints.

%% file: sections/related_work.tex
\section{Related Work}

\textbf{Post-Training Scalar Quantization.}
Post-training quantization (PTQ) compresses a pretrained model from a small
calibration corpus without retraining. \textbf{GPTQ}~\citep{frantar2023gptq}
sets the near-lossless 4-bit weight-only baseline via approximate second-order
error minimization; \textbf{AWQ}~\citep{lin2023awq} and
\textbf{SmoothQuant}~\citep{xiao2023smoothquant} rescale salient channels or
migrate activation difficulty to weights, and rotation methods
(\textbf{QuaRot}~\citep{ashkboos2024quarot},
\textbf{SpinQuant}~\citep{liu2024spinquant}) enforce weight incoherence before
rounding. These scalar methods are near-lossless at 4 bits but collapse at 2
bits, where small uniform grids cannot represent LLM weight distributions. We
take this collapse as our starting point.

\textbf{Ultra-Low-Bit and Vector Quantization.}
Vector and codebook quantization jointly encode weights for better
rate-distortion below 3 bits. \textbf{QuIP}~\citep{chee2023quip} and
\textbf{QuIP\#}~\citep{tseng2024quip} pair incoherence preprocessing with
lattice codebooks and fine-tuning; \textbf{AQLM}~\citep{egiazarian2024aqlm},
\textbf{GPTVQ}~\citep{vanbaalen2024gptvq}, \textbf{VPTQ}~\citep{vptq2024}, and
\textbf{QTIP}~\citep{tseng2024qtip} push extreme compression through additive,
hierarchical, or trellis-coded codebooks, and
\textbf{YAQA}~\citep{tseng2026yaqa} attains state-of-the-art 2-bit quality via a
Fisher-weighted rounding objective. Others target 1-bit compression
(\textbf{SqueezeLLM}~\citep{kim2023squeezellm},
\textbf{BiLLM}~\citep{huang2024billm}, \textbf{PB-LLM}~\citep{shang2023pbllm},
\textbf{OneBit}~\citep{xu2024onebit}). These works study representation in
isolation; we instead analyze how calibration data and recovery interact with
the quantizer, narrowing the gap to \textsc{bf16} at a fraction of QAT compute.

\textbf{Learning-Based PTQ.}
Between calibration-only PTQ and full QAT, some methods \emph{learn} the mapping
to a quantization grid. \textbf{OmniQuant}~\citep{shao2023omniquant} learns
clipping thresholds block-wise, and \textbf{GSQ}~\citep{dadgarnia2026gsq} learns
scalar-grid assignments through a Gumbel-Softmax relaxation. Because grid
learning uses gradient signal but neither backpropagates through the full network
nor runs multi-epoch training, we treat GSQ as an intermediate-cost baseline that
isolates grid learning from representation choice.

\textbf{Calibration Data.}
Most PTQ methods estimate statistics from 128--512 generic C4 or WikiText-2
samples, assuming content is interchangeable
\citep{frantar2023gptq,lin2023awq,shao2023omniquant,egiazarian2024aqlm}.
\citet{williams2023calibration} first showed task-performance sensitivity, but
only at 4 bits and mainly for web data. We extend this to 2 bits and show that
{calibration composition moves generation far more than question answering (Section~\ref{sec:exp-calibration}). Our sweeps vary composition at a fixed budget; they do not separate data quality from quantity, which we leave for future work.}

\textbf{Recovery for Quantized Models.}
Lightweight recovery narrows the post-quantization gap without full QAT, usually
via knowledge distillation~\citep{hinton2015distilling}:
\textbf{LLM-QAT}~\citep{liu2023llmqat},
\textbf{BitDistiller}~\citep{du2024bitdistiller}, and
\textbf{EfficientQAT}~\citep{chen2024efficientqat} update full weight matrices
over multiple epochs and ignore codebook structure, while
\textbf{LoRA}/\textbf{QLoRA}~\citep{hu2022lora,dettmers2023qlora} reintroduce
high-precision parameters that inflate the effective bitwidth. Our default recipe
instead fine-tunes only the quantizer's native parameters (the vector codebook, or
the scalar scale and zero-point), staying exactly at the target bitwidth and
converging in a few steps (a LoRA variant is in
Appendix~\ref{app:lora-recovery}). {We evaluate EfficientQAT as a trained scalar baseline on the same data as our recovery runs (Table~\ref{tab:rev-eqat}) and} further {study the} interaction between
calibration composition and recovery {in a crossed design (Figure~\ref{fig:rev-interaction})}, motivating a jointly designed
calibration--recovery pipeline.

%% file: sections/methodology.tex
\section{Background and Methodology}
\label{sec:method}

We build the PTQ-RFT pipeline, post-training quantization followed by recovery
fine-tuning, from three ingredients. We first fix notation for the two quantization
families we use, scalar and vector quantization, and identify for each the small
set of parameters that recovery later updates
(Sections~\ref{sec:method-scalar}--\ref{sec:method-vector}). We then contrast the
two ways to obtain a low-bit model, namely post-training quantization (PTQ),
which fits each layer locally to the full-precision model, and quantization-aware
training (QAT), which reshapes the whole network end to end
(Section~\ref{sec:method-ptqqat}). This contrast motivates our recovery stage
(Section~\ref{sec:method-recovery}), a short fine-tuning step that stays
\emph{parameter-matched} to the quantized format and therefore does not raise the
effective bitwidth.

\subsection{Scalar Quantization}
\label{sec:method-scalar}

Scalar quantization maps each weight independently onto a shared
one-dimensional grid of $2^b$ levels. In the general \emph{asymmetric} form,
round-to-nearest (RTN) over $[\mathrm{min},\mathrm{max}]$ uses a scale $s$ and
integer zero-point $z$:
\begin{equation}
\begin{aligned}
s &= \frac{\mathrm{max}-\mathrm{min}}{2^b-1},
\qquad
z = \operatorname{round}\!\left(-\frac{\mathrm{min}}{s}\right),
\\[2pt]
q &= \operatorname{clip}\!\left(
    \operatorname{round}\!\left(\frac{x}{s}\right)+z,\,
    0,\,2^b-1
\right),
\qquad
\hat{x}=s(q-z).
\end{aligned}
\end{equation}
The zero-point aligns the grid so that zero is represented exactly. Symmetric
quantization instead fixes $z$ and uses
$s=\mathrm{max}/2^{b-1}$ with $\hat{x}=sq$, yielding a simpler offset-free
kernel at the cost of one degree of freedom. Because RTN treats weights
independently, its error grows rapidly at low bitwidths.

\textbf{Error-aware PTQ.}
Methods such as GPTQ \citep{frantar2023gptq} account for the effect of weight
error on a layer's outputs. For $y=Wx$, calibration distribution $X$, and input
Hessian $H_X=\mathbb{E}_{x\sim X}[xx^\top]$, they approximately solve
\begin{equation}
\min_{\hat W\in\mathcal{Q}}
\mathbb{E}_{x\sim X}\!\big\|(W-\hat W)x\big\|_2^2
=
\min_{\hat W\in\mathcal{Q}}
\operatorname{Tr}\!\big[
    (W-\hat W)H_X(W-\hat W)^\top
\big],
\label{eq:hessian-weighted-quantization}
\end{equation}
where $\mathcal{Q}$ is the feasible set defined by the quantization format.
GPTQ greedily quantizes coordinates while updating the remaining weights to
compensate for accumulated error. This objective is a local second-order proxy for the full-model KL divergence;
methods such as KronQ \citep{lee2026kronq} improve the proxy using structured
curvature.

The same formulation enables inexpensive calibration-mixture ablations. If
$H_j$ is estimated independently on domain $j$, then a mixture with proportions
$\{\alpha_j\}$ has
\begin{equation}
H_{\boldsymbol{\alpha}}
=
\sum_j \alpha_j H_j,
\qquad
\alpha_j\geq 0,
\qquad
\sum_j\alpha_j=1.
\label{eq:hessian-mixture}
\end{equation}
Thus, calibration composition can be varied by reweighting precomputed
Hessians. Methods such as QAT and GSQ generally require a separate optimization run for
each mixture.

Scalar methods differ primarily in how they determine the grid parameters.
These are the parameters recovered by the recovery fine-tuning stage. It freezes the integer codes
$q$ at the target bitwidth and fine-tunes only $s$ in the symmetric case or
$(s,z)$ in the asymmetric case, without introducing large high-precision weight
updates.

\subsection{Vector Quantization}
\label{sec:method-vector}

Vector quantization changes the representation rather than only the precision
of individual weights. It partitions a layer into $d$-dimensional sub-vectors
and maps each to an entry in a learned codebook
$\mathcal{C}=\{c_k\}_{k=1}^K\subset\mathbb{R}^d$. Reshaping $W$ into vectors
$\{w_i\}_{i=1}^N$ with assignments $a_i\in\{1,\ldots,K\}$ gives
\begin{equation}
\min_{\mathcal{C},\,\{a_i\}}
\sum_{i=1}^N \left\|w_i-c_{a_i}\right\|_2^2,
\qquad
\hat w_i=c_{a_i}.
\end{equation}
By representing points in $\mathbb{R}^d$ rather than values on a scalar grid,
vector quantization exploits correlations among weights and remains effective
at bitwidths where scalar formats degrade. Its effective bitwidth includes the
per-vector indices and the amortized codebook storage.
{A rate-distortion argument explains this. For Gaussian weights at $2$ bits per weight, the optimal fixed-rate scalar (Lloyd--Max) quantizer has mean-squared error $0.1175\,\sigma^2$~\citep{max1960quantizing,lloyd1982least}, against the Shannon bound $\sigma^2 2^{-2R}=0.0625\,\sigma^2$~\citep{shannon1959coding,cover2006elements} ($2.7$\,dB apart). A scalar grid cannot close this gap at a fixed rate; an 8-dimensional codebook at the same rate is not restricted to a product of scalar grids and can close part of it~\citep{gersho1992vector,tseng2024qtip}. On rotated (near-Gaussian) Llama-3.1-8B weights, the measured 2-bit error is $0.118$ for the optimal scalar quantizer and $0.080$ for an 8-dimensional codebook (Table~\ref{tab:rev-rd}).}

\begin{table}[t]
\centering
\caption{Normalized MSE $\|W-\hat W\|^2/\|W\|^2$ at 2 bits per weight (mean over 20 matrices; Appendix~\ref{app:rev-rd-sec}). Gaussian: optimal scalar $0.1175$, Shannon bound $0.0625$.}
\label{tab:rev-rd}
\scriptsize
\renewcommand{\arraystretch}{0.92}
\setlength{\tabcolsep}{3.5pt}
\begin{tabular}{l cc cc cccc}
\toprule
& \multicolumn{2}{c}{Kurtosis} & \multicolumn{2}{c}{Scalar g128 rounding} & \multicolumn{4}{c}{Optimal codebook, dimension} \\
\cmidrule(lr){2-3}\cmidrule(lr){4-5}\cmidrule(lr){6-9}
Model & raw & rotated & raw & rotated & 1 & 2 & 4 & 8 \\
\midrule
Phi-4-mini   & 6.36 & 3.25 & 0.264 & 0.250 & 0.124 & 0.114 & 0.101 & 0.079 \\
Llama-3.1-8B & 5.70 & 3.01 & 0.281 & 0.250 & 0.118 & 0.109 & 0.098 & 0.080 \\
Qwen3-8B     & 4.59 & 3.00 & 0.266 & 0.250 & 0.118 & 0.109 & 0.097 & 0.079 \\
\bottomrule
\end{tabular}
\end{table}

Vector PTQ can use the same error-aware formulation as
Equation~\ref{eq:hessian-weighted-quantization}, with $\mathcal{Q}$ replaced by
the set of matrices representable by the chosen codebooks. VPTQ
\citep{vptq2024} uses channel-independent second-order optimization to solve
this layerwise reconstruction problem efficiently. YAQA
\citep{tseng2026yaqa} instead derives its curvature from the
full-model KL divergence and uses a Kronecker-factored approximation to guide
adaptive rounding, including over trellis-coded quantization sets. The
calibration-mixture construction in Equation~\ref{eq:hessian-mixture} similarly applies to vector PTQ. 

For vector formats, the recovery stage recovers the continuous codebook $\mathcal{C}$
while holding the discrete assignments $\{a_i\}$ fixed. Methods such as PV-tuning
\citep{malinovskii2024pvtuning} additionally optimize the assignments. Extending parameter-efficient recovery to {optimize} discrete codebook or trellis
assignments is a promising direction for future work.

\subsection{Post-Training Quantization versus Quantization-Aware Training}
\label{sec:method-ptqqat}

PTQ and QAT optimize fundamentally different objectives. Layer-wise PTQ methods,
including GPTQ~\citep{frantar2023gptq}, KronQ~\citep{lee2026kronq}, and
YAQA~\citep{tseng2026yaqa}, minimize a local Hessian-weighted reconstruction
error. This keeps each quantized layer close to its full-precision counterpart,
leaving a small residual error that parameter-efficient fine-tuning can correct.

\begin{wraptable}{r}{0.55\columnwidth}
\vspace{-\baselineskip}
\centering
\scriptsize
\renewcommand{\arraystretch}{0.9}
\caption{Persistent memory (GB) and estimated H100-80GB counts. PTQ and
PTQ-RFT report 2-bit/4-bit configurations; H100 counts are ordered
as~Phi/Llama/Qwen.}
\label{tab:memory-tradeoff}
\resizebox{\linewidth}{!}{%
\begin{tabular}{@{}lcccc@{}}
\toprule
Method & \shortstack{Phi-4\\(GB)} & \shortstack{Llama-8B\\(GB)}
& \shortstack{Qwen-MoE\\(GB)} & \shortstack{H100s\\(P/L/Q)} \\
\midrule
PTQ & $1.0/1.9$ & $2.0/4.0$ & $7.6/15.3$ & $1/1/1$ \\
PTQ+RFT & $9.0/10.0$ & $18.8/20.8$ & $71.5/79.1$ & $1/1/1^{\ast}$ \\
GSQ (2-bit) & $2.2$ & $5.0$ & $17.0^{\dagger}$ & $1/1/4^{\dagger}$ \\
Full QAT & $53.8$ & $112.4$ & $427.0$ & $1/2/6$ \\
\bottomrule
\end{tabular}%
}

\parbox{\linewidth}{\scriptsize
Counts are persistent-state estimates and exclude activations, temporary
buffers, and communication workspaces.
$^{\ast}$The 4-bit configuration occupies $79.1$\,GB and requires additional
runtime capacity in practice.
$^{\dagger}$Aggregate across four expert-sharded GPUs ($4.3$\,GB per GPU).}
\vspace{-4mm}
\end{wraptable}

QAT instead optimizes the end-to-end objective through the quantizer, typically
using the straight-through estimator (STE)~\citep{liu2023llmqat},
$\partial x_q/\partial x \approx 1$. Weights and intermediate representations
may therefore {diverge} from the full-precision model if doing so improves the
task loss. This flexibility requires model-wide backpropagation and optimizer
state, making QAT substantially more expensive than layer-wise PTQ.

Table~\ref{tab:memory-tradeoff} compares persistent memory for Phi-4-mini,
Llama-3.1-8B, and Qwen3-30B-A3B, using $P=3.84$B, $8.03$B, and $30.5$B
parameters, respectively, and provides an approximate H100s count. An ideal $b$-bit PTQ checkpoint requires $bP/8$
bytes. PTQ followed by Recovery Fine-Tuning (RFT) requires approximately $2P+bP/8+12P/128$ bytes for the \textsc{bf16}
teacher, quantized model, and scale optimizer state. Full QAT requires
approximately $14P$ bytes for the \textsc{bf16} teacher, student weights and gradients,
and two FP32 Adam moments; an optional FP32 master copy adds $4P$ bytes.

GSQ~\citep{dadgarnia2026gsq} lies between these regimes. It refines one
transformer block at a time, avoiding a separate teacher and full-model optimizer
state, but stores four \textsc{bf16} assignment logits per optimized weight. Its logits,
gradients, Lion momentum, and reconstructed weights require approximately $26N$
bytes for a block with $N$ optimized weights. This yields $2.2$\,GB for
Phi-4-mini, $5.0$\,GB for Llama-3.1-8B, and $17.0$\,GB for Qwen3-30B-A3B
across four expert-sharded GPUs.

Overall, GSQ optimizes a discrete assignment for each weight in the active block,
whereas PTQ+RFT freezes weights and assignments and updates only continuous
quantizer parameters. PTQ+RFT therefore pays for a full-precision teacher but
retains a much smaller trainable state.

\vspace{-2mm}
\subsection{Recovery Fine-Tuning}
\label{sec:method-recovery}

Recovery fine-tuning (RFT) restores accuracy lost to low-bit quantization without the
cost of full QAT. In its base form, it freezes the quantized backbone and updates
only the parameters already present in the quantized representation: the scales
$s$ and zero-points $z$ for scalar formats and the codebook $\mathcal{C}$ and assignments $\mathcal{A}$ for vector formats. Since
assignments are discrete and difficult to optimize, we only consider the codebook $\mathcal{C}$ for recovery fine-tuning. This
preserves the target storage format and effective bitwidth while applying
uniformly to scalar and vector quantizers.

\textbf{Base objective.}
We train against the full-precision model as teacher, using logit-level distillation. Let
$z_t$ and $z_s$ denote the teacher and student logits, respectively. Our objective
combines a temperature-scaled forward KL divergence with next-token
cross-entropy:
\begin{equation}
\mathcal{L}
=
\alpha\,\mathcal{L}_{\mathrm{CE}}
+
(1-\alpha)\,\mathcal{L}_{\mathrm{KD}},
\qquad
\mathcal{L}_{\mathrm{KD}}
=
\tau^2
\operatorname{KL}\!\left(
\sigma\!\left(\tfrac{z_t}{\tau}\right)
\,\big\|\,
\sigma\!\left(\tfrac{z_s}{\tau}\right)
\right),
\label{eqn:base-loss}
\end{equation}
where $\sigma(\cdot)$ is the softmax, $\tau$ is the distillation temperature, and
$\alpha\in[0,1]$ balances the two terms. This objective uses only model logits and
introduces no additional trainable parameters. We use it for all base experiments;
alternative divergences, temperatures, and top-$k$ logit truncation appear
in Appendix~\ref{app:recovery-ablation}.

\textbf{Adapter-based recovery.} Recovery can also add parameter-efficient modules such as LoRA adapters~\citep{hu2022lora}. Prior work recovers frozen quantized models with low-rank adapters (QLoRA~\citep{dettmers2023qlora}, QA-LoRA~\citep{xu2023qalora}, LoftQ~\citep{li2024loftq}); these are complementary to quantizer-only recovery. Adapters, however, add deployment overhead and may remain in half precision unless merged and requantized, so we use quantizer-only recovery as our default {format and bit-width preserving} setting. Appendix~\ref{app:lora-recovery} evaluates joint \mbox{quantizer--adapter recovery}.

%% file: sections/experiments_v2.tex
\vspace{-3mm}
\section{Experiments and Results}
\label{sec:experiments}

We organize the evaluation around four factors that determine the viability
of weight-only 2-bit quantization: representation, optimization budget,
calibration data, and recovery parameterization. Specifically, we ask (i)~where scalar PTQ fails as precision drops from 4 to 2 bits; (ii)~whether vector codebooks avert this failure across dense and mixture-of-experts (MoE) architectures; (iii)~how learnable scalar PTQ and QAT compare with vector PTQ; and (iv)~how calibration and codebook recovery trade off language understanding, mathematical reasoning, and code generation.
\vspace{-2mm}
\subsection{Experimental setup}
\label{sec:exp-setup}

\textbf{Models and quantization.}
We evaluate Phi-4-mini-instruct~\citep{microsoft2025phi4mini} (3.8B, abbreviated
Phi-4-mini), Llama-3.1-8B~\citep{grattafiori2024llama3}, and
Qwen3-30B-A3B~\citep{yang2025qwen3}, a 30B-parameter MoE with approximately
3B active parameters. The no-recovery scalar baselines are
SpinQuant~\citep{liu2024spinquant} {(fixed Hadamard rotations, not learned; Table~\ref{tab:app-method-configs})} and QuaRot~\citep{ashkboos2024quarot}. At
2 bits, scalar weights use group size 128 with a
half-precision per-group scale, or approximately $2.125$ bits per quantized
weight ($2.12$ in the appendix tables; an asymmetric zero-point variant would be
$2.25$). The vector baselines are VPTQ~\citep{vptq2024}, which uses a
$65{,}536$-entry codebook over 8-dimensional vectors (2.00 bits/weight), and
YAQA~\citep{tseng2026yaqa}, a Fisher-weighted trellis quantizer (approximately
2.26 bits/weight). We also compare GSQ~\citep{dadgarnia2026gsq} as a group-128 scalar PTQ
method that optimizes discrete assignments layer by layer,
BitDistiller~\citep{du2024bitdistiller} as group-128 scalar QAT with
self-distillation{, and EfficientQAT~\citep{chen2024efficientqat} (w2g128, $\le\!2.25$ bits; using public code and hyperparameters, our calibration data and exact recovery sample; Appendix~\ref{app:rev-eqat})}. {These trained scalar methods} receive additional optimization unavailable to the headline
no-recovery PTQ methods; their comparisons test whether {learning-based scalar quantization},
rather than {a stronger vector} representation, is sufficient at 2 bits.
{We quantize attention and MLP projections, including every Qwen3-MoE expert, and
keep router gates, embeddings, norms, and the language-model head in
half precision (Appendix~\ref{app:router} quantizes the routers).}\\
{\textbf{Effective bits.} We count indices, codebooks, and scales: the vector methods use $2.09$--$2.33$ bits per weight and the scalar methods $2.12$--$2.25$ (Table~\ref{tab:app-method-configs}). On the MoEs, VPTQ shares one codebook per layer and projection type across experts.}
{Every method, protocol, and cost profile is specified in
Appendices~\ref{app:protocol}--\ref{app:profiling} and Table~\ref{tab:app-method-configs}.}

\textbf{Calibration and evaluation.}
We compare web, instruction, reasoning, code, and convex mixtures of these
calibration sources {(data card in Appendix~\ref{app:rev-provenance})}. {AvgQA averages MMLU~\citep{hendrycks2021mmlu} (5-shot), WinoGrande~\citep{sakaguchi2021winogrande}, LAMBADA~\citep{paperno2016lambada}, ARC-Easy, ARC-Challenge~\citep{clark2018arc}, HellaSwag~\citep{zellers2019hellaswag}, and PIQA~\citep{bisk2020piqa} (the last four length-normalized); AvgGen averages GSM8K~\citep{cobbe2021gsm8k} (5-shot) and HumanEval~\citep{chen2021humaneval} (pass@1); IFEval~\citep{zhou2023ifeval} uses prompt-strict accuracy.}

\textbf{Recovery.}
Recovery freezes the quantized backbone and updates only the separate vector
codebook. The reference run uses a \textsc{bf16} teacher, forward KL,
temperature $\tau=1$, full-vocabulary logits, and web recovery data. We vary
one factor at a time. This protocol tests codebook-restricted
post-quantization recovery rather than full QAT.
{Appendix~\ref{app:rev-provenance} lists every recovery run's base, data, and budget.}
\vspace{-2mm}
\subsection{Representation determines whether 2-bit PTQ works}
\label{sec:exp-representation}

\begin{table}[t]
\centering
\caption{\textbf{Quantization ladder.} Best QA and generation at each
precision under \emph{web calibration}, selected separately for QA and
generation, \emph{without recovery}. Scalar entries report the better of
SpinQuant (\textsuperscript{S}) and QuaRot (\textsuperscript{Q}) (on Qwen3-MoE, QuaRot uses expert-wise
rotation). Superscripts mark the
winning method.}
\label{tab:quantization-ladder}
\scriptsize
\renewcommand{\arraystretch}{0.9}
\setlength{\tabcolsep}{6pt}
\begin{adjustbox}{max width=\linewidth}
\begin{tabular}{l cc cc cc cc}
\toprule
& \multicolumn{2}{c}{\textsc{bf16}}
& \multicolumn{2}{c}{Scalar 4-bit}
& \multicolumn{2}{c}{Scalar 2-bit}
& \multicolumn{2}{c}{Vector 2-bit {(YAQA)}} \\
\cmidrule(lr){2-3}\cmidrule(lr){4-5}\cmidrule(lr){6-7}\cmidrule(lr){8-9}
Model & QA & Gen & QA & Gen & QA & Gen & QA & Gen \\
\midrule
Phi-4-mini    & 71.0 & 73.2 & 69.3\textsuperscript{Q} & 65.8\textsuperscript{S} & 29.3\textsuperscript{S} & 0.0 & {63.3} & 12.2 \\
Llama-3.1-8B  & 73.1 & 42.9 & 71.8\textsuperscript{S} & 38.0\textsuperscript{Q} & 29.0\textsuperscript{Q} & 0.0 & 65.1 & 9.6 \\
Qwen3-30B-A3B & 72.6 & 61.8 & 70.3\textsuperscript{S} & 66.2\textsuperscript{S} & 29.4\textsuperscript{S} & 0.0 & {66.0} & 54.0 \\
\bottomrule
\end{tabular}
\end{adjustbox}
\end{table}

\textbf{Scalar PTQ has a sharp 2-bit cliff.}
Table~\ref{tab:quantization-ladder} compresses the full per-task results into QA
and generation means. At 4 bits, the best scalar method stays within
1.3--2.3 QA points of \textsc{bf16}. At 2 bits, every evaluated scalar {PTQ} method
collapses to approximately 29\% QA and zero generation {(curves in Appendix~\ref{app:gen-cliff}), for both evaluated methods on all models; trained scalar methods avoid the collapse (Table~\ref{tab:vq-architecture}).}
Generation reveals a distinct MoE regime already at 4 bits: Phi-4-mini and
Llama lose 7.4 and 4.9 AvgGen points, whereas Qwen3-MoE gains 4.4 points
(66.2 versus 61.8). We treat the latter as full retention at the resolution
of this evaluation, not as evidence that quantization improves generation.

\textbf{Vector codebooks revive 2-bit models.}
Changing only the representation improves 2-bit QA from the random floor to
{63.3} with YAQA on Phi-4-mini, 65.1 with YAQA on Llama, and {66.0 with YAQA} on
Qwen3-MoE under web calibration. The effect on generation is
architecture-dependent: dense models retain little of their \textsc{bf16}
generation score, whereas Qwen3-MoE reaches 54.0 AvgGen with YAQA. Thus,
bitwidth alone is insufficient to predict failure; representation, architecture and scale jointly determine the 2-bit regime.
\vspace{-2mm}
\subsection{Vector PTQ outperforms trained scalar quantization}
\label{sec:exp-vector}

\begin{table*}[t]
\centering
\caption{\textbf{Cross-paradigm comparison at 2 bits.} Best AvgQA (left) and AvgGen (right).
{VPTQ and YAQA select, per metric, the best of the same calibration recipes (the five domains of Table~\ref{tab:app-vq-domain}; reason+web on OLMoE; \S\ref{app:best-recipes})}{; EfficientQAT (EQAT, w2g128) selects from its data grid (Table~\ref{tab:rev-eqat}). Dense VPTQ omits YAQA's model-wide rotation (Appendix~\ref{app:rev-vptq-rot}). Bold: best per model. $^{*}$GSQ without rotation.}}
\label{tab:vq-architecture}
\footnotesize
\renewcommand{\arraystretch}{0.9}
\setlength{\tabcolsep}{2.5pt}
\begin{tabular}{@{}l c ccc cc c ccc cc@{}}
\toprule
& \multicolumn{6}{c}{QA (AvgQA $\uparrow$)} & \multicolumn{6}{c}{Generation (AvgGen $\uparrow$)} \\
\cmidrule(lr){2-7}\cmidrule(lr){8-13}
Model & \textsc{bf16} & GSQ & BitDist. & {EQAT} & VPTQ & YAQA
& \textsc{bf16} & GSQ & BitDist. & {EQAT} & VPTQ & YAQA \\
\midrule
Phi-4-mini    & 71.0 & 52.6 & 53.0 & {57.3} & {57.6} & \textbf{63.3}
& 73.2 & 21.6 & 18.4 & {30.3} & {29.5} & \textbf{37.4} \\
Llama-3.1-8B  & 73.1 & 57.0 & 60.3 & {62.4} & {63.3} & {\textbf{65.1}}
& 42.9 & 20.0 & {20.3} & {22.3} & {19.7} & \textbf{26.6} \\
Qwen3-MoE     & 72.6 & {65.1$^{*}$} & 55.9 & {\textbf{69.0}} & {64.8} & {66.0}
& 61.8 & {50.4$^{*}$} & 45.7 & {59.0} & {53.0} & {\textbf{61.8}} \\
OLMoE-1B-7B   & {68.1} & {61.0$^{*}$} & 49.1 & {56.2} & {63.0} & {\textbf{64.0}}
& 32.4 & {13.6$^{*}$} & 0.9 & {9.7} & {15.2} & {\textbf{20.6}} \\
\bottomrule
\end{tabular}
\end{table*}

{\textbf{Matched-data trained scalar.} EfficientQAT on our data avoids the collapse and beats BitDistiller. It trails YAQA on generation on every model (within noise on Qwen3-MoE; Table~\ref{tab:rev-sig}) and on QA on the dense models and OLMoE ($2.7$--$7.8$ AvgQA). On Qwen3-MoE, its best-QA checkpoint leads by $3.0$ AvgQA but keeps little generation ($32.1$ vs.\ $54.0$ AvgGen for the two best-QA checkpoints; Tables~\ref{tab:rev-eqat}, \ref{tab:calibration-main}). Calibrating it on the all-4 mixture of the best GSQ and BitDistiller entries improves its generation, but not to YAQA's (Table~\ref{tab:rev-eqat}).}

\begin{table}[t]
\centering
\caption{\textbf{Paired significance of the main comparisons.} YAQA minus baseline (points), $95\%$ paired bootstrap interval below; checkpoints as in Table~\ref{tab:vq-architecture}. Evaluation details in Appendix~\ref{app:rev-sigtests}.}
\label{tab:rev-sig}
\scriptsize
\renewcommand{\arraystretch}{0.92}
\setlength{\tabcolsep}{2.6pt}
\begin{tabular}{l cc cc cc cc}
\toprule
 & \multicolumn{2}{c}{Phi-4-mini} & \multicolumn{2}{c}{Llama-3.1-8B} & \multicolumn{2}{c}{OLMoE-1B-7B} & \multicolumn{2}{c}{Qwen3-30B-A3B} \\
\cmidrule(lr){2-3}\cmidrule(lr){4-5}\cmidrule(lr){6-7}\cmidrule(lr){8-9}
Baseline & QA & Gen & QA & Gen & QA & Gen & QA & Gen \\
\midrule
GSQ & $10.4$ & $16.2$ & {$8.2$} & $6.9$ & $3.4$ & $5.7$ & {$0.7$} & {$10.2$} \\[-1.5pt]
 & {\tiny[$9.7$, $11.1$]} & {\tiny[$13.2$, $19.4$]} & {{\tiny[$7.5$, $8.8$]}} & {\tiny[$4.3$, $9.3$]} & {\tiny[$2.7$, $4.0$]} & {\tiny[$3.1$, $8.4$]} & {{\tiny[$0.01$, $1.4$]}} & {{\tiny[$6.1$, $14.2$]}} \\
BitDistiller & $10.5$ & $18.9$ & {$5.3$} & $6.7$ & $15.4$ & $18.9$ & {$9.0$} & {$17.3$} \\[-1.5pt]
 & {\tiny[$9.8$, $11.3$]} & {\tiny[$15.9$, $22.1$]} & {{\tiny[$4.7$, $6.0$]}} & {\tiny[$3.5$, $9.8$]} & {\tiny[$14.6$, $16.1$]} & {\tiny[$16.0$, $21.8$]} & {{\tiny[$8.2$, $9.7$]}} & {{\tiny[$12.6$, $22.0$]}} \\
EfficientQAT & $5.8$ & $7.4$ & {$2.6$} & $4.7$ & $8.2$ & $9.6$ & {$-2.9$} & {$3.2$} \\[-1.5pt]
 & {\tiny[$5.1$, $6.5$]} & {\tiny[$3.8$, $11.0$]} & {{\tiny[$2.0$, $3.3$]}} & {\tiny[$2.1$, $7.3$]} & {\tiny[$7.5$, $8.8$]} & {\tiny[$7.1$, $12.2$]} & {{\tiny[$-3.6$, $-2.3$]}} & {{\tiny[$-0.8$, $7.5$]}} \\
VPTQ & $5.5$ & $7.9$ & {$1.8$} & $6.7$ & $1.1$ & $5.2$ & {$1.1$} & {$5.9$} \\[-1.5pt]
 & {\tiny[$4.8$, $6.1$]} & {\tiny[$4.0$, $11.8$]} & {{\tiny[$1.2$, $2.4$]}} & {\tiny[$3.9$, $9.5$]} & {\tiny[$0.5$, $1.8$]} & {\tiny[$2.6$, $7.7$]} & {{\tiny[$0.3$, $1.7$]}} & {{\tiny[$1.7$, $10.2$]}} \\
\bottomrule
\end{tabular}
\end{table}

{\textbf{Noise floor and significance.} Recovery seeds move AvgGen by at most $1.0$ and AvgQA by at most $0.1$ point (standard deviation; Table~\ref{tab:rev-interaction}). Under a paired item-level bootstrap, every YAQA-versus-baseline gap on the dense models and OLMoE is significant (Table~\ref{tab:rev-sig}; all $p<0.001$ on the dense models). On Qwen3-MoE, YAQA leads every baseline except EfficientQAT, which leads on QA ($p<0.001$), and its Qwen3-MoE QA lead over GSQ is narrow ($+0.7$ [$0.01$, $1.37$], $p{=}0.045$). On Qwen3-MoE generation, YAQA and EfficientQAT are statistically tied ($+3.2$ [$-0.8$, $7.5$], $p{=}0.13$).}

Table~\ref{tab:vq-architecture} separates representation from optimization
budget. On Llama, YAQA exceeds GSQ by {8.1} AvgQA points and BitDistiller by
{4.8} points{; on the MoEs, where we run GSQ without rotation (its rotation breaks OLMoE), GSQ reaches $61.0$ and $65.1$ AvgQA on OLMoE and Qwen3-MoE, $3.0$ and $0.9$ below YAQA (Table~\ref{tab:vq-architecture}; Appendix~\ref{app:rev-checks}).}

{Within vector PTQ, YAQA beats VPTQ on Phi-4-mini and Llama by 5.7 and 1.8 AvgQA and 7.9 and 6.9 AvgGen. VPTQ runs without YAQA's model-wide rotation, which would cost it 2.6--12.1 AvgQA (Appendix~\ref{app:rev-vptq-rot}). YAQA also beats a bit-matched VPTQ control (Table~\ref{tab:app-bitmatch}) and VPTQ on both MoEs (Qwen3-MoE: 66.0 vs.\ 64.8 AvgQA, 61.8 vs.\ 53.0 AvgGen). With Hessian mixtures, which we ran for both methods only on Qwen3-MoE, VPTQ reaches 67.2 / 55.2 and YAQA 66.7 / 62.4 (Tables~\ref{tab:rev-vptq-shared}, \ref{tab:calibration-main}). Qwen3-8B (63.7 vs.\ 60.8) and 9 of 10 dense per-domain comparisons show the same ordering (Appendices~\ref{app:qwen3-8b}, \ref{app:vq-domains}).}
\vspace{-2mm}
\subsection{Calibration chiefly controls generation}
\label{sec:exp-calibration}

\begin{table}[t]
\centering
\caption{\textbf{Calibration composition primarily changes generation.}
QA/Gen means for representative 2-bit YAQA recipes; bold: best in the full sweep.
$\Delta_{\rm full}$ includes all recipes, including model-specific mixtures.}
\label{tab:calibration-main}
\scriptsize
\renewcommand{\arraystretch}{0.9}
\setlength{\tabcolsep}{6pt}
\begin{adjustbox}{max width=\linewidth}
\begin{tabular}{l cc cc cc}
\toprule
& \multicolumn{2}{c}{Phi-4-mini}
& \multicolumn{2}{c}{Llama-3.1-8B}
& \multicolumn{2}{c}{Qwen3-MoE} \\
\cmidrule(lr){2-3}\cmidrule(lr){4-5}\cmidrule(lr){6-7}
Recipe & QA & Gen & QA & Gen & QA & Gen \\
\midrule
Web                    & {\textbf{63.3}} & 12.2 & 65.1 & 9.6 & 66.0 & 54.0 \\
Reasoning              & 58.5 & \textbf{37.4} & 62.1 & \textbf{26.6} & 64.5 & 61.8 \\
Code                   & 55.9 & 18.7 & 60.8 & 22.6 & 62.8 & 48.9 \\
Reason+web (0.5/0.5)   & 62.6 & 24.7 & \textbf{65.7} & 17.8 & \textbf{66.7} & 56.5 \\
Webreason (0.7/0.3)    & 63.0 & 27.7 & 65.5 & 15.1 & 65.8 & 56.2 \\
All-4 ($0.25\times4$)  & 58.8 & 19.9 & 65.2 & 20.1 & 66.0 & 59.3 \\
RWI442 (0.4/0.4/0.2)   & 59.5 & 17.1 & 62.7 & 12.2 & 66.5 & \textbf{62.4} \\
\midrule
$\Delta_{\rm full}$    & {7.4} & 26.7 & 4.9 & {17.0} & 3.9 & 13.5 \\
\bottomrule
\end{tabular}
\end{adjustbox}
\end{table}

Calibration composition remains important after the representation is fixed.
Table~\ref{tab:calibration-main} retains representative recipes with complete task coverage across all three
models. QA varies by
{7.4} points on Phi-4-mini, 4.9 on Llama, and 3.9 on Qwen3-MoE. Generation is
substantially less stable, varying by 26.7, {17.0}, and 13.5 points,
respectively. Reasoning calibration maximizes Llama
generation, while broad mixtures remain competitive for QA. On Qwen3-MoE,
the targeted reasoning/web/instruction mixture RWI442 gives the best
generation score, while a balanced reasoning/web mixture gives the best QA.
The complete cross-model sweep appears in
Appendix~\ref{app:calibration-full}.

\textbf{Generation retention separates dense models from the MoE.}
Selecting {vector} methods separately for each metric, the strongest task-complete
QA/Gen results are {63.3}/37.4 on Phi-4-mini, {65.1}/26.6 on Llama, and
{66.0/61.8} on Qwen3-MoE. These correspond to 89\%/51\%, {89\%}/62\%, and
{91\%/100\%} of the respective \textsc{bf16} scores (AvgQA/AvgGen for this
task-complete selection), matching the {$51$--$100\%$} per-model \emph{generation}
retention band quoted in the introduction and Table~\ref{tab:recipe}. QA retention is therefore
similar across families, whereas generation retention spans more than a
factor of two, as the per-task retention heatmaps and QA--generation
scatterplots make explicit (Appendix~\ref{app:analyses},
Figures~\ref{fig:app-retention}--\ref{fig:app-qa-gen}). This result
makes architecture a first-order variable in 2-bit quantization. We note,
however, that a second MoE (OLMoE) retains far less generation than Qwen3-MoE
({$\approx\!64\%$} vs {$\approx\!100\%$} of \textsc{bf16}), so architecture strongly
influences, but does not by itself determine, generation retention {(completion-style HumanEval; chat-template retention in Table~\ref{tab:rev-vptq-shared}.)}
\vspace{-2mm}
\subsection{Codebook-only recovery helps}
\label{sec:exp-recovery}

\begin{wrapfigure}{r}{0.52\linewidth}
\centering
\includegraphics[width=\linewidth]{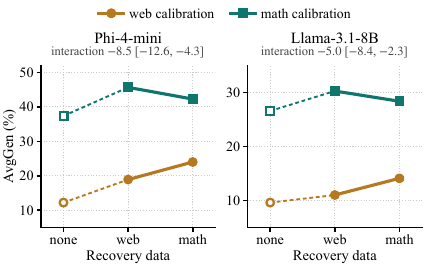}
\vspace{-1.6\baselineskip}
\caption{{\textbf{The best recovery data depends on the calibration data} (2-bit YAQA). Short codebook-only recovery; error bars: standard deviation over three recovery seeds (no-recovery points: one run). Subtitles: interaction with $95\%$ bootstrap interval; held-out Qwen3-8B: $-9.4$ [$-13.3$, $-5.3$], all three predictions hold (Table~\ref{tab:rev-interaction}).}}
\label{fig:rev-interaction}
\vspace{-0.6\baselineskip}
\end{wrapfigure}
Recovery improves capability without restoring full-precision parity
(Table~\ref{tab:recovery-summary}), and does so at a fraction of QAT compute,
placing recovered checkpoints on the {QA} accuracy--compute Pareto front
(Appendix~\ref{app:compute-figs}). On Phi-4-mini, codebook-only distillation
improves AvgQA {over each row's own base by 0.5--2.5 points (within noise on the web base; every generation gain is significant, Table~\ref{tab:rev-sig-calrec}), and on the reasoning-calibrated base improves GSM8K from 54.1 to 62.5 and HumanEval from 20.7 to 36.6 (Appendix~\ref{app:rev-provenance} relates Tables~\ref{tab:recovery-summary} and~\ref{tab:app-recovery-ofat}); on Llama, short and long recovery improves AvgQA by {$0.4$--$1.9$} and $1.6$--$1.8$ points (Tables~\ref{tab:rev-interaction}, \ref{tab:recovery-summary})}. Recovery data matters more than KD details: loss
direction, teacher precision, {temperature, and top-$k$ truncation (Appendix~\ref{app:rev-topk})} each move AvgQA
by at most 0.2, and a 4-bit teacher matches \textsc{bf16} (63.6 vs.\ 63.5). Qwen3-MoE
improves both AvgQA ($66.0\to67.8$) and HumanEval (${34.1}\to45.1$){, and broad web and mixture recovery data
outperform narrow ones}
(Table~\ref{tab:app-qwen-recovery}).
\begin{table}[tb]
\centering
\caption{\textbf{Best-recipe codebook recovery.} {Each 2-bit YAQA model before (base) and after (+rec.) long codebook-only recovery (bases and data in Table~\ref{tab:rev-provenance}).}}
\label{tab:recovery-summary}
\scriptsize
\renewcommand{\arraystretch}{0.72}
\setlength{\tabcolsep}{3pt}
\begin{tabular}{ll cc cc cc cc}
\toprule
& & \multicolumn{2}{c}{{AvgQA}} & \multicolumn{2}{c}{{GSM8K}} & \multicolumn{2}{c}{{HEval}} & \multicolumn{2}{c}{{IFEval}} \\
\cmidrule(lr){3-4}\cmidrule(lr){5-6}\cmidrule(lr){7-8}\cmidrule(lr){9-10}
Model & {Calibration recipe} & {base} & {+rec.} & {base} & {+rec.} & {base} & {+rec.} & {base} & {+rec.} \\
\midrule
\multirow{6}{*}{Phi-4-mini}
 & \textsc{bf16} ceiling     & \multicolumn{2}{c}{71.0} & \multicolumn{2}{c}{81.9} & \multicolumn{2}{c}{64.6} & \multicolumn{2}{c}{56.6} \\
 & web                       & {63.3} & 63.8 & {17.7} & 40.5 & {6.7} & 19.5 & {29.9} & 27.4 \\
 & reasoning                 & {58.5} & 61.0 & {54.1} & \textbf{62.5} & {20.7} & \textbf{36.6} & {32.2} & 27.7 \\
 & reason\_web               & {62.6} & 63.7 & {37.8} & 51.6 & {11.6} & 26.8 & {37.2} & \textbf{36.0} \\
 & webreason                 & {63.0} & \textbf{64.1} & {42.0} & 52.5 & {13.4} & 23.8 & {36.2} & 31.6 \\
 & rw64 (recommended)        & 61.8 & 63.8 & 44.9 & 57.7 & 8.5 & 28.7 & {34.0} & 34.2 \\
\midrule
\multirow{4}{*}{Llama-3.1-8B}
 & \textsc{bf16} ceiling     & \multicolumn{2}{c}{73.1} & \multicolumn{2}{c}{50.9} & \multicolumn{2}{c}{34.8} & \multicolumn{2}{c}{8.9} \\
 & {reason\_web}               & 65.7 & {\textbf{67.4}} & 22.1 & {\textbf{29.6}} & 13.4 & {15.9} & 7.4 & {8.5} \\
 & webreason                 & {65.5} & {67.3} & {18.5} & {25.9} & {11.6} & {17.7} & {7.2} & {5.5} \\
 & all-4                     & {65.2} & {66.8} & {20.1} & {26.5} & {20.1} & {\textbf{24.4}} & {9.1} & {\textbf{10.2}} \\
\midrule
\multirow{2}{*}{Qwen3-30B-A3B}
 & \textsc{bf16} ceiling     & \multicolumn{2}{c}{72.6} & \multicolumn{2}{c}{90.0} & \multicolumn{2}{c}{33.5} & \multicolumn{2}{c}{28.3} \\
 & {web}                       & 66.0 & \textbf{67.8} & {73.8} & \textbf{76.0} & {34.1} & \textbf{45.1} & 24.4 & 22.0 \\
\bottomrule
\end{tabular}
\end{table}

{\textbf{Calibration and recovery are coupled.} In a crossed design (Figure~\ref{fig:rev-interaction}; Table~\ref{tab:rev-interaction}), math recovery gives the higher AvgGen on the web-calibrated base and web recovery on the math-calibrated base, on Phi-4-mini and Llama-3.1-8B (intervals exclude zero) and on a held-out Qwen3-8B. The interaction is not an artifact of one benchmark, base, or sample: it persists on held-out tasks never used for selection ($-6.8$, $-4.0$, $-6.0$ on Phi-4-mini, Llama, Qwen3-8B), across independently re-quantized Phi-4-mini bases ($-9.6$ [$-15.8$, $-3.4$]) and disjoint recovery samples ($-8.5$ to $-5.3$), and in domain perplexity (Appendices~\ref{app:rev-crossed}, \ref{app:rev-robust}). Calibration remains the larger factor ($+16.8$ AvgGen on Llama); the design covers one domain pair at a short budget.}

\textbf{MoE robustness.}
Quantizing Qwen3-MoE's router to 2 bits changes accuracy by at most 0.2{, and with codebooks counted the MoE QA-retention advantage is small (91--94\% vs.\ 89\% on dense models; OLMoE YAQA $64.0/68.1$)}; its mechanism remains open
(Appendices~\ref{app:router}, \ref{app:olmoe}, and~\ref{app:limitations}).

%% file: sections/best_practices.tex
\vspace{-1.0em}
\section{A Practical Recipe for 2-Bit Models}
\label{sec:best-practices}

\begin{figure}[t!]
\centering
\begin{minipage}[c]{0.53\linewidth}
\centering
\captionsetup{type=table}
\caption{\textbf{2-bit decision guide.} Recommended choice and evidence per
decision (Section~\ref{sec:experiments}); AvgQA \mbox{unless noted}.}
\label{tab:recipe}
\scriptsize
\renewcommand{\arraystretch}{0.9}
\setlength{\tabcolsep}{2pt}
\begin{tabular}{@{}ll>{\raggedright\arraybackslash\leavevmode}p{0.45\linewidth}@{}}
\toprule
Decision & Recommendation & Evidence \\
\midrule
Representation   & Vector (not scalar)   & $29\!\to\!63$--{$66$} \\
Dense quantizer  & YAQA                  & {$+5.7$/$+1.8$} vs.\ VPTQ \\
MoE quantizer    & {YAQA}            & {$+1.2$ QA/$+8.8$ gen.\ vs VPTQ} \\
Calibration      & Hessian mixture       & moves gen.\ $\gg$ QA \\
Recovery         & Codebook-only KD      & ${63.0}\!\to\!64.1$ \\
{Recovery data}  & {pick with calib.} & {interaction $-5$ to $-9$} \\
KD knobs         & Defaults ($\tau{=}1$) & $\le\!0.2$ AvgQA \\
Teacher          & 4-bit (not \textsc{bf16})      & $63.6\!\approx\!63.5$ \\
Router           & keep fp16             & $<\!0.2$ (not cause) \\
Decoding         & Rep.\ penalty $1.3$   & {$40.5\!\to\!63.4$ MATH} \\
Gen.\ boost      & $+$LoRA (optional)    & HEval $\to\!32.9$ \\
Reporting        & split QA/gen          & ret.\ {$51$--$100\%$} \\
\bottomrule
\end{tabular}
\end{minipage}\hfill
\begin{minipage}[c]{0.44\linewidth}
\centering
\includegraphics[width=\linewidth]{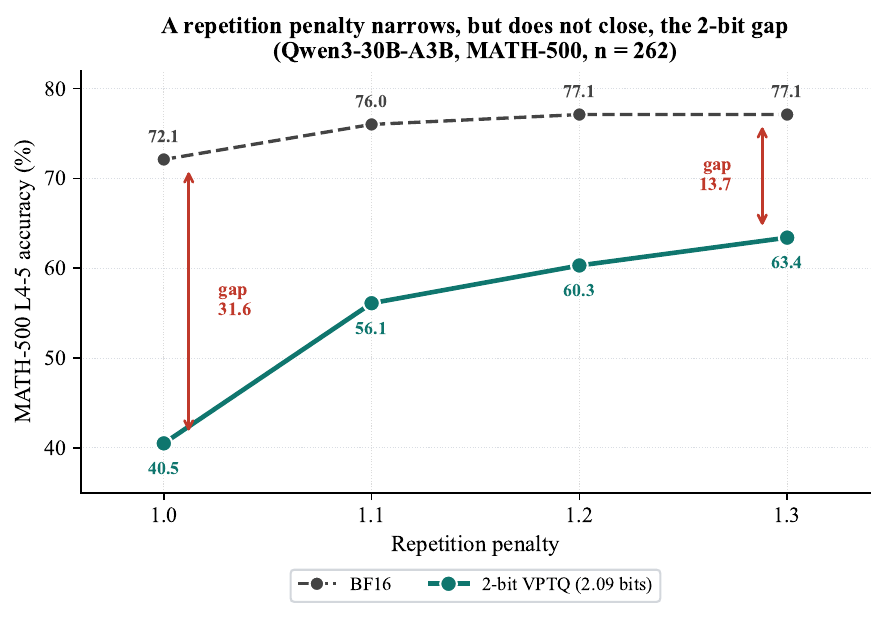}
\end{minipage}
\caption{\textbf{A repetition penalty {narrows, but does not close, the 2-bit gap}}
(Qwen3-30B-A3B, MATH-500 {levels 4--5, $n{=}262$}). {Applied at every precision, a penalty of $1.3$ raises the 2-bit model ($2.09$ bits) from $40.5$ to $63.4$ and \textsc{bf16} from $72.1$ to $77.1$, so the matched-decoding gap shrinks from $31.6$ to $13.7$ points (Table~\ref{tab:rev-math500}). Loop scores under the penalty are not measured in this study.}}
\label{fig:app-overthinking-recovery}
\end{figure}

The decision order is: \textbf{representation} first (vector, not scalar), then
\textbf{architecture} (YAQA for dense models {and MoEs}), then \textbf{data}
(task-aligned Hessian calibration and short codebook-only recovery with a
disposable 4-bit teacher{, recovery data chosen with the calibration}), as summarized in Table~\ref{tab:recipe}. Calibration
composition moves generation far more than QA, and the KD hyperparameters barely
matter once we choose the recovery data. Keep the MoE router in fp16 as a cheap
default, and report QA and generation separately, since generation retention
spans a much wider band ({$51$--$100\%$}) than QA. This single-GPU recipe yields
$63$--{$68$}$\%$ AvgQA at 2 bits across the models we study.

\textbf{Decoding matters at 2 bits.} A final lever acts at inference. On
reasoning models with an explicit thinking mode, 2-bit quantization amplifies an
\emph{overthinking} pathology: on MATH-500 {(2-bit Qwen3-MoE, $n{=}262$), 2 bits increase looping and cut accuracy from $72.1$ to $40.5$; a repetition penalty of $1.3$ restores $63.4$ but also helps \textsc{bf16}, so the gap shrinks from $31.6$ to $13.7$ (Figure~\ref{fig:app-overthinking-recovery}; Appendix~\ref{app:rev-math500}).}

\section{Conclusion}
\label{sec:conclusion}
\vspace{-3mm}
{Our study shows that the 2-bit cliff is primarily a representation failure rather than an inevitable consequence of bitwidth. Scalar PTQ collapses, whereas vector codebooks retain strong QA and, when calibration and recovery data are chosen jointly, much of the generation ability. There are three key takeaways from our work. First, the quantizer's representation should be fixed before the data. Second, calibration and recovery data interact, so they should be selected together rather than tuned in isolation. Third, 2-bit models should be compared at equal effective bits, with codebooks and scales counted, and reported separately on QA and generation, because QA averages hide most of the damage. Promising directions for future work include compute-matched comparisons with trained scalar methods, recovery that targets instruction following and reasoning, calibration--recovery studies across more domains and larger budgets, broader coverage of MoE architectures, and low-bit kernels that turn memory savings into faster decoding (Appendix~\ref{app:limitations}).}

%% file: sections/appendix_v2.tex
\providecommand{\na}{\textcolor{gray}{--}}
\setcounter{topnumber}{3}\setcounter{totalnumber}{4}\renewcommand{\topfraction}{0.85}\renewcommand{\textfraction}{0.1}\renewcommand{\floatpagefraction}{0.75}

\makeatletter\providecommand{\AppFloatBarrier}{\par\begingroup\let\@elt\relax\edef\@tempa{\@botlist\@deferlist\@dbldeferlist}\ifx\@tempa\@empty\endgroup\else\endgroup\clearpage\fi}\makeatother
\makeatletter\providecommand{\AppNeedSpace}[1]{\par\penalty-100\begingroup\setlength{\dimen@}{#1}\dimen@ii\pagegoal\advance\dimen@ii-\pagetotal\ifdim\dimen@>\dimen@ii\break\fi\endgroup}\makeatother
\clearpage
\begingroup\setlength{\parskip}{0pt}
\providecommand{\appsec}[1]{\par\vspace{4pt plus 1fil}\noindent\textbf{\makebox[2.2em][l]{\ref{#1}}\nameref{#1}}\hfill\textbf{\pageref{#1}}\par\vspace{1pt plus 0.15fil}}
\providecommand{\appsub}[1]{\noindent\hspace*{2.2em}\makebox[2.6em][l]{\ref{#1}}\nameref{#1}\,\dotfill\,\pageref{#1}\par\vspace{0pt plus 0.15fil}}
\section*{{Appendix contents}}
\appsec{app:experiments}
\appsub{app:protocol}
\appsub{app:method-configs}
\appsub{app:compute-accounting}
\appsub{app:profiling}
{\appsub{app:rev-deploy-sec}}
\appsec{app:sec-repr}
{\appsub{app:rev-rd-sec}}
\appsub{app:cross-paradigm}
\appsub{app:best-recipes}
\appsub{app:vq-domains}
{\appsub{app:rev-vptq-rot}}
\appsub{app:trained-baselines}
{\appsub{app:rev-eqat}}
{\appsub{app:rev-checks}}
{\appsub{app:rev-sigtests}}
\appsec{app:sec-calib}
\appsub{app:calibration-full}
{\appsub{app:rev-crossed}}
{\appsub{app:rev-robust}}
{\appsub{app:rev-sig-calrec}}
\appsec{app:sec-recovery}
\appsub{app:recovery-ablation}
{\appsub{app:rev-provenance}}
\appsub{app:lora-recovery}
{\appsub{app:rev-topk}}
\appsub{app:qwen-recovery}
\appsec{app:sec-models}
\appsub{app:qwen3-8b}
\appsub{app:olmoe}
{\appsub{app:rev-bits}}
\appsub{app:router}
\appsec{app:sec-gen}
\appsub{app:analyses}
\appsub{app:gen-cliff}
\appsub{app:compute-figs}
\appsub{app:ifeval}
\appsub{app:overthinking}
{\appsub{app:rev-math500}}
\appsec{app:limitations}
\endgroup
\pagebreak
\makeatletter\setlength{\@fptop}{0pt}\makeatother %

\section{\texorpdfstring{{Experimental setup and cost}}{Experimental setup and cost}}
\label{app:experiments}
\suppressfloats[t]

\subsection{Quantization and evaluation details}
\label{app:protocol}

\paragraph{Effective precision.}
All reported scalar runs use the scale-only group-128 format ($2.125$ bits per
weight, rounded to $2.12$). The vector index rates (VPTQ $2.00$, YAQA
${\approx}2.26$) exclude codebook storage{; Table~\ref{tab:app-method-configs} lists effective bits including codebooks and scales}.

\paragraph{Evaluation and reproducibility.}
We score all tasks through an \texttt{lm-evaluation-harness}-based split-task
orchestrator~\citep{gao2023evalharness}. Multiple-choice tasks (ARC-Easy, ARC-Challenge, HellaSwag, PIQA,
WinoGrande, LAMBADA) use log-likelihood scoring; GSM8K and HumanEval are decoded greedily (temperature $0$, no
sampling) and scored by strict-match and pass@1, respectively. Calibration and
recovery sample counts are the audited per-method values used in the compute
accounting (\S\ref{app:compute-accounting}). Scalar calibration uses $128$ sequences, and the vector Hessians use $128$--$2{,}048$ sequences per domain.
Training and recovery runs use a fixed random seed ($42$).
{``P-bf16'' in a caption marks results evaluated in bfloat16 with this harness and these metrics.}

\subsection{Per-method quantization configurations}
\label{app:method-configs}

This section consolidates the configuration of every quantization and recovery
method we evaluate. Table~\ref{tab:app-method-configs} summarizes the key
settings; the paragraphs below give the details. Unless noted, all methods
target 2-bit weights at group size $128$, share the calibration and evaluation
protocol of \S\ref{app:protocol}, and use a fixed random seed ($42$).

\begin{table*}[t]
\centering
\caption{\textbf{Per-method configuration summary.} Paradigm, effective
bits/weight (\emph{Eff.}{; indices, codebooks, and scales}), the parameters
each method trains, and the calibration or recovery data. All methods use group
size $128$ and the shared evaluation protocol (\S\ref{app:protocol}). {VPTQ uses $2.33$, $2.27$, $2.30$, $2.09$, and $2.14$ bits on Phi-4-mini, Llama, Qwen3-8B, Qwen3-MoE, and OLMoE; on the MoEs it shares one codebook per layer and projection type across experts.}}
\label{tab:app-method-configs}
\small
\setlength{\tabcolsep}{3.5pt}
\begin{tabular}{ll c l l}
\toprule
Method & Paradigm & Eff.\ b/w & Trained parameters & Calibration / recovery data \\
\midrule
QuaRot       & scalar PTQ (rotation) & 2.12 & {none (GPTQ)}         & $128$ sequences \\
SpinQuant    & scalar PTQ (rotation) & 2.12 & {none (GPTQ)}         & $128$ sequences \\
GSQ          & learned scalar        & 2.12 & grid assignments           & $0.1$B-token, $10$ epochs \\
BitDistiller & full QAT              & 2.12 & all weights                & $1$-epoch mixture \\
{EfficientQAT} & {block-wise + E2E QAT} & {${\le}2.25$} & {all weights, then scales} & {our calibration + recovery data} \\
VPTQ         & vector PTQ            & {2.09--2.33} & codebook (post-hoc)        & per-domain Hessians \\
YAQA         & vector PTQ            & {2.26--2.31} & codebook (post-hoc)        & per-domain Hessians \\
PTQ+RFT      & codebook recovery     & {2.26--2.31} & codebook only              & task-aligned mixtures \\
\bottomrule
\end{tabular}
\end{table*}

\paragraph{QuaRot and SpinQuant (scalar PTQ).}
Both apply {a fixed orthogonal rotation and then quantize with GPTQ} at group
size $128$ with half-precision per-group scale and zero-point. QuaRot uses {randomized Hadamard rotations of the residual stream with norm fusion; our SpinQuant runs use fixed randomized Hadamard matrices at SpinQuant's R1 and R2 positions instead of learned rotations}; on Qwen3-MoE we apply the rotation expert-wise. Both consume $128$ calibration sequences and
add no trained weights, giving our no-recovery scalar reference.

\paragraph{VPTQ (vector PTQ).}
VPTQ maps every $8$-dimensional weight sub-vector to one of $65{,}536$ ($2^{16}$)
codebook entries ($2.00$ index bits/weight) through channel-independent
second-order (input-Hessian) optimization{, with a Hadamard transform of each layer's input. Dense VPTQ runs without YAQA's model-wide rotation (Appendix~\ref{app:rev-vptq-rot}).} We collect input
Hessians per calibration domain. The bit-matched control of
\S\ref{app:vq-domains} instead uses a $262{,}144$ ($2^{18}$) codebook
($2.25$ bits/weight).

\paragraph{YAQA (vector PTQ).}
YAQA quantizes onto a trellis codebook using Fisher-weighted, Kronecker-factored
curvature derived from the full-model KL objective, at ${\sim}2.26$ index
bits/weight, and retains a separate, fine-tuning-ready codebook under the same
incoherence rotation. We estimate curvature from $128$--$2{,}048$ sequences per
calibration domain.

\paragraph{GSQ (learned scalar).}
GSQ learns discrete grid assignments through a Gumbel-softmax relaxation,
initialized with GPTQ ($512$ sequences). It optimizes one transformer block at a
time with detached inputs for $10$ epochs over a $0.1$B-token calibration set
($48{,}828$ sequences of length $2048$) at group size $128$, using a Lion
optimizer. Block-wise processing keeps its peak memory low despite a high total
FLOP count (Tables~\ref{tab:app-compute-all} and~\ref{tab:app-deploy-all}).

\paragraph{BitDistiller (full QAT).}
BitDistiller performs full-model quantization-aware training with asymmetric
activation clipping and confidence-aware self-distillation (CAKLD,
$\beta\approx0.84$) on top of a Hadamard-rotated $2$-bit model at group size
$128${ (unlike the original BitDistiller)}. We train for one epoch with FSDP full-shard across $8$ GPUs, sequence
length $1024$, per-device batch $1$ and gradient accumulation $8$ (effective
batch $64$), learning rate $2\times10^{-5}$ with cosine decay and $10\%$ warmup.
The recovery mixture holds ${\sim}500$k sequences for the dense models and
$47{,}616$ for Qwen3-MoE (the step-$800$ checkpoint; \S\ref{app:trained-baselines}).

\paragraph{Codebook-only recovery (PTQ+RFT).}
Our recovery stage freezes the quantized backbone and the discrete assignments
and updates only the continuous codebook (or the scale and zero-point for scalar
formats), leaving the effective bitwidth unchanged. It distills from a
\textsc{bf16} (or disposable $4$-bit) teacher with a temperature-$1$ forward-KL
term plus next-token cross-entropy over the full vocabulary, using per-device
batch $1$, gradient accumulation $8$, learning rate $10^{-4}$, a cosine schedule
with $500$ warmup steps, and sequence length $1024$. Recovery data are the
task-aligned mixtures of \S\ref{app:recovery-ablation}; the optional LoRA variant
adds rank-$16$ ($\alpha{=}32$) adapters (\S\ref{app:lora-recovery}).

\subsection{Training-compute accounting for the recovery comparison}
\label{app:compute-accounting}

Table~\ref{tab:app-compute} reports the training-compute budget used in our
recovery-efficiency comparison. We count the dominant algorithmic FLOPs
from the completed run configurations, with $N=3.84\times10^{9}$ parameters.
For GSQ, layer inputs are detached, so each optimization token requires a layer
forward and weight/logit backward (approximately $4ND$), not a full-model
$6ND$ update; we additionally count one activation-propagation pass and GPTQ
initialization. BitDistiller costs approximately $8ND$, which covers the frozen-teacher forward pass
plus the student forward and backward passes. This gives $16{,}136$ and $15{,}717$ PFLOP,
respectively. The difference is only 2.7\% and is smaller than the uncertainty
of this accounting; accordingly, we treat the methods as compute-comparable
rather than claiming a strict ordering.

YAQA includes the Fisher-Hessian pipeline omitted by a calibration-only count.
The completed run collects both input and output covariances for 128 linear
maps. Besides model forward/backward passes, each covariance contributes an
outer product; using the actual Phi-4 layer dimensions gives
$2\sum_\ell(d_{\mathrm{in},\ell}^2+d_{\mathrm{out},\ell}^2)=
2.617\times10^{10}$ FLOPs per token. We collected four domain Hessians (code, instruction,
reasoning, and web) concurrently on one 8-GPU node and reused
to produce eight base/mixture checkpoints. Their observed mean sequence lengths
(406, 216, 3455, and 934 tokens) give 426 PFLOP in total, or 53 PFLOP amortized
per output. A standalone reason+web checkpoint charged both source Hessians
would instead cost approximately 373 PFLOP. The plot uses the amortized value
because that is how we produced the reported checkpoint suite. The headline
``half the compute of QAT'' comparison contrasts the recovery stages directly
(PTQ$+$RFT $8{,}349$ vs BitDistiller $15{,}717$ PFLOP) and is therefore
unaffected by this base-quantization Hessian-amortization choice. Matrix
factorization and trellis search add less than 0.1 PFLOP at these dimensions.

\begin{table}[htbp]
\centering
\caption{\textbf{Compute and weight-memory accounting for the recovery
comparison} (Phi-4-mini). FLOPs include the dominant operations
described above. Weight GB is the deployed quantized-weight footprint implied
by effective bits/weight; it excludes shared non-quantized tensors and runtime
buffers. AvgQA is the 7-task average. {YAQA and PTQ+RFT are the rw64 checkpoint of Table~\ref{tab:recovery-summary} before and after recovery.}}
\label{tab:app-compute}
\small
\setlength{\tabcolsep}{5pt}
\begin{adjustbox}{max width=\columnwidth}
\begin{tabular}{l l r r c}
\toprule
Method & Repr. & FLOPs (PFLOP) & Weight GB & AvgQA \\
\midrule
SpinQuant 4-bit    & scalar & 2       & 1.920 & 69.2 \\
SpinQuant 2-bit    & scalar & 2       & 1.020 & 29.3 \\
YAQA 2-bit {(rw64)} & vector & 53      & 1.085 & 61.8 \\
{EfficientQAT 2-bit} & {scalar} & {526} & {1.080} & {57.3} \\
PTQ+RFT 2-bit {(rw64)} & vector & 8{,}349 & 1.085 & \textbf{63.8} \\
BitDistiller 2-bit & scalar & 15{,}717 & 1.020 & 53.0 \\
GSQ 2-bit          & scalar & 16{,}136 & 1.020 & 52.6 \\
\bottomrule
\end{tabular}
\end{adjustbox}
\end{table}

The same accounting extends to the other model families. Table~\ref{tab:app-compute-all}
lists the audited per-method training compute for Phi-4-mini, Llama-3.1-8B and
Qwen3-30B-A3B; the ordering is stable across models; rotation and forward-only
vector PTQ cost tens to hundreds of PFLOP, whereas the learning-based scalar
baselines cost over $10{,}000$ PFLOP. For the MoE BitDistiller entry we count
activated-parameter passes ($N=3.34\times10^{9}$). The reported
Qwen3-30B-A3B step-$800$ checkpoint processes $800\times64=51{,}200$ sequences of
$1{,}024$ tokens, so $8ND\approx1{,}401$~PFLOP.

\begin{table}[htbp]
\centering
\caption{\textbf{Audited training compute per method (PFLOP).} One-time
algorithmic cost from the completed run configurations
(\S\ref{app:compute-accounting}), counting the dominant model passes. Scalar
rotations (SpinQuant) and forward-only vector PTQ (VPTQ, YAQA) are orders of
magnitude cheaper than the learning-based baselines (GSQ, BitDistiller).
\emph{Per token}, BitDistiller's full-model forward+backward ($8N$) costs twice
GSQ's block-wise, detached-input update ($4N$); the two nonetheless reach
comparable dense-model totals because GSQ optimizes a $0.1$B-token calibration
set for $10$ epochs whereas BitDistiller trains a single epoch over the recovery
mixture (${\sim}500$k sequences for the dense models). The Qwen3-MoE
BitDistiller entry is likewise one epoch, over its smaller $47{,}616$-sequence
mixture (the step-$800$ checkpoint that produced its accuracy row); its budget is
correspondingly smaller.
PTQ+RFT adds a codebook-only recovery pass (frozen backbone, $6N$/token) to the
YAQA base and remains cheaper than BitDistiller on every model
($1.9\times$ on Phi-4-mini, $3.7\times$ on Llama, $54\times$ on Qwen3-MoE). MoE
rows use activated-parameter passes. Dashes mark configurations not run for that
model. {EfficientQAT is counted at $6ND$ over its Block-AP and E2E-QP token passes (Appendix~\ref{app:rev-eqat}).}}
\label{tab:app-compute-all}
\small
\setlength{\tabcolsep}{6pt}
\begin{adjustbox}{max width=\columnwidth}
\begin{tabular}{l r r r}
\toprule
Method & Phi-4-mini & Llama-3.1-8B & Qwen3-MoE \\
\midrule
SpinQuant (rotation)        & 2        & 4        & 2 \\
VPTQ (vector PTQ)           & 36       & 69       & 36 \\
YAQA (vector PTQ)           & 53       & 728      & 15 \\
PTQ+RFT (codebook recovery) & 8{,}349  & 8{,}938  & 24 \\
GSQ (scalar QAT)            & 16{,}136 & 33{,}743 & 14{,}035 \\
BitDistiller (QAT)          & 15{,}717 & 32{,}866 & 1{,}401 \\
{EfficientQAT (QAT)}        & {526} & {1{,}101} & {458} \\
\bottomrule
\end{tabular}
\end{adjustbox}
\end{table}

\subsection{Measured deployment and training-cost profile}
\label{app:profiling}

Table~\ref{tab:app-deploy-all} contrasts the methods along the axes that the
representation actually changes. These axes are storage (on-disk and packed footprints), the
peak GPU memory the quantization or recovery procedure needs, and the one-time
training compute. Three trends stand out. First, the training-free methods
(scalar SpinQuant and forward vector PTQ, VPTQ/YAQA) only need the \textsc{bf16}
model resident, so their peak memory equals a forward pass ($11$--$63$~GB across
our models). Second, the learning-based baselines diverge sharply. GSQ refines
one transformer block at a time and is the \emph{lowest}-memory method
($2$--$17$~GB), whereas BitDistiller runs full QAT at ${\sim}14P$ and is the
highest ($54$--$427$~GB, exceeding one $80$~GB GPU on the MoE). {EfficientQAT, which trains block by block and then only the step sizes end to end, sits in between at a measured $8$--$20$~GB.} Third, our
PTQ+RFT recovery adds a frozen-backbone codebook pass whose peak memory stays
{at most $1.14\times$} that of the YAQA base ({$9$}--$72$~GB), while remaining far
cheaper in compute than QAT (Table~\ref{tab:app-compute-all}). Runtime
throughput and bandwidth are architecture-bound (the harness dequantizes every
checkpoint to \textsc{bf16}), so we report the deployable low-bit footprint via
the Packed column rather than a dense-runtime measurement.

\begin{table*}[t]
\centering
\caption{\textbf{Per-method cost profile for all 2- and 4-bit baselines.}
Effective bits/weight (\emph{Eff.}{, Table~\ref{tab:app-method-configs}}), on-disk checkpoint size, \emph{Packed} low-bit footprint (\textsc{bf16} embeddings kept), peak GPU memory of the quantization or recovery procedure, and one-time \emph{Train PFLOP} (Table~\ref{tab:app-compute-all}). GSQ, QAT, and RFT peak memory is analytic (\S\ref{sec:method-ptqqat}). {$^\S$Measured with \texttt{torch.cuda.max\_memory\_allocated} as the larger of EfficientQAT's Block-AP and E2E-QP stages, with the batch size and context length of our runs.} $^\ddagger$Dense-\textsc{bf16} serialization estimate.}
\label{tab:app-deploy-all}
\small
\setlength{\tabcolsep}{6pt}
\begin{adjustbox}{max width=\textwidth}
\begin{tabular}{ll c ccc r}
\toprule
& & \shortstack{Eff.\\b/w} & \shortstack{On-disk\\(GB)} & \shortstack{Packed\\(GB)} & \shortstack{Peak mem\\(GB)} & \shortstack{Train\\PFLOP} \\
\midrule
\multirow{7}{*}{\shortstack[l]{Phi-4-mini\\(3.84B)}}
 & SpinQuant 4-bit    & 4.25 & 7.67$^\ddagger$  & 2.94 & 11.0  & 2 \\
 & SpinQuant 2-bit    & 2.12 & 7.67$^\ddagger$  & 2.08 & 11.0  & 2 \\
 & GSQ 2-bit          & 2.12 & 4.20        & 2.25 & 2.2   & 16{,}136 \\
 & BitDistiller 2-bit & 2.12 & 7.67        & 2.08 & 53.8  & 15{,}717 \\
 & {EfficientQAT 2-bit} & {2.25} & {2.09} & {2.13} & {8.0$^\S$} & {526} \\
 & YAQA 2-bit         & 2.26 & 9.82        & 2.31 & 12.2  & 53 \\
 & PTQ+RFT 2-bit      & 2.26 & 9.82        & 2.31 & {9.0}  & 8{,}349 \\
\midrule
\multirow{8}{*}{\shortstack[l]{Llama-3.1-8B\\(8.03B)}}
 & SpinQuant 4-bit    & 4.25 & 16.06$^\ddagger$ & 5.04 & 17.7  & 4 \\
 & SpinQuant 2-bit    & 2.12 & 16.06$^\ddagger$ & 3.04 & 17.7  & 4 \\
 & GSQ 2-bit          & 2.12 & 1.99$^\ddagger$  & 3.04 & 5.0   & 33{,}743 \\
 & BitDistiller 2-bit & 2.12 & 16.06$^\ddagger$ & 3.04 & 112.4 & 32{,}866 \\
 & {EfficientQAT 2-bit} & {2.25} & {3.97} & {3.16} & {8.7$^\S$} & {1{,}101} \\
 & VPTQ 2-bit         & {2.27} & 19.79       & {3.17} & 17.7  & 69 \\
 & YAQA 2-bit         & 2.26 & 18.04       & 3.17 & 17.7  & 728 \\
 & PTQ+RFT 2-bit      & 2.26 & 18.04       & 3.17 & 18.8  & 8{,}938 \\
\midrule
\multirow{8}{*}{\shortstack[l]{Qwen3-30B-A3B\\(30.5B, MoE)}}
 & SpinQuant 4-bit    & 4.25 & 61.06$^\ddagger$ & 16.68 & 62.8  & 2 \\
 & SpinQuant 2-bit    & 2.12 & 61.06$^\ddagger$ & 8.65  & 62.8  & 2 \\
 & GSQ 2-bit          & 2.12 & 8.03$^\ddagger$  & 8.65  & 17.0  & 14{,}035 \\
 & BitDistiller 2-bit & 2.12 & 61.06$^\ddagger$ & 8.65  & 427.0 & 1{,}401 \\
 & {EfficientQAT 2-bit} & {2.25} & {9.40} & {9.12} & {19.5$^\S$} & {458} \\
 & VPTQ 2-bit         & {2.09} & 95.55       & {8.52}  & 62.8  & 36 \\
 & YAQA 2-bit         & 2.26 & 69.70       & 9.16  & 62.8  & 15 \\
 & PTQ+RFT 2-bit      & 2.26 & 69.70       & 9.16  & 71.5  & 24 \\
\bottomrule
\end{tabular}
\end{adjustbox}
\end{table*}

\subsection{Deployment cost}
\label{app:rev-deploy-sec}

Table~\ref{tab:rev-deploy} measures weight memory and decode throughput on one H100 for public checkpoints of each format in one software stack, under Hugging Face \texttt{generate} and under a compiled, static-cache batch-1 decoder (QTIP's own generator). 2-bit weights cut weight memory $3.2$--$4.2\times$; decode speed depends on the kernels and the batch size, and we make no latency claim beyond this table.

\begin{table}[htbp]
\centering
\caption{Weight memory and decode throughput (tokens/s), Llama-3.x-8B on one H100, one software stack. \emph{Static+graph} uses batch 1, a static KV cache, and a compiled decode step ($^{\dagger}$kernel cannot be graph-captured). The EfficientQAT row uses the public Llama-3-8B w2g128 GPTQ checkpoint.}
\label{tab:rev-deploy}
\scriptsize
\setlength{\tabcolsep}{3.5pt}
\begin{tabular}{l c ccc c}
\toprule
& & \multicolumn{2}{c}{Batch 1} & Batch 8 & \\
\cmidrule(lr){3-4}\cmidrule(lr){5-5}
Format & Weights (GiB) & generate & static+graph & generate & Peak, bs8 (GiB) \\
\midrule
\textsc{bf16}                      & 14.96 & 56.4 & 118.4 & 445.3 & 18.5 \\
QTIP 2-bit trellis (YAQA's format) & 3.59  & 2.4  & \textbf{145.3} & 65.4 & 7.1 \\
VPTQ v8-k65536                     & 4.62  & 55.4 & 55.4$^{\dagger}$ & 284.4 & 8.1 \\
EfficientQAT w2g128 (GPTQ kernels) & 3.70  & 8.4  & 14.3$^{\dagger}$ & 33.6 & 7.2 \\
\bottomrule
\end{tabular}
\end{table}

\section{\texorpdfstring{{Vector versus scalar quantization}}{Vector versus scalar quantization}}
\label{app:sec-repr}

\subsection{Measured rate--distortion}
\label{app:rev-rd-sec}

We test the rate--distortion argument of Section~\ref{sec:method-vector} on 20 linear layers from five depths of each dense model. After two-sided incoherence processing (random signs and block Hadamard), we quantize them at exactly 2 bits per weight with $k$-means codebooks of dimension 1--8, fit on each matrix, and with the group-wise rounding of scalar PTQ (Table~\ref{tab:rev-rd}). The rotated weights are near-Gaussian, the errors follow the Gaussian theory, and the error falls as the codebook dimension grows. Codebook storage is counted in Table~\ref{tab:app-method-configs}.

\subsection{Cross-paradigm 2-bit comparison}
\label{app:cross-paradigm}

Table~\ref{tab:app-cross-paradigm} compares the trained scalar checkpoints
directly with the strongest available vector PTQ result on the same model in
AvgQA. On Phi-4-mini the vector advantage is {10.7} points over GSQ and {10.3}
points over BitDistiller; on Llama it is {8.1} points over GSQ and {4.8} points over
BitDistiller; and on Qwen3-MoE it is {10.1 points over BitDistiller}. The comparison is deliberately not presented as a
compute-matched efficiency result, because BitDistiller performs full-model QAT, GSQ
performs layer-wise discrete optimization, and the vector methods are
no-recovery PTQ.

\begin{table}[htbp]
\centering
\caption{\textbf{Vector PTQ versus trained scalar quantization at 2 bits.}
AvgQA ($\uparrow$). ``Best VQ'' is the strongest task-complete
VPTQ or YAQA checkpoint for each model{, as in Table~\ref{tab:vq-architecture}}. $\Delta$ is Best VQ minus the scalar
baseline.}
\label{tab:app-cross-paradigm}
\small
\setlength{\tabcolsep}{3pt}
\begin{adjustbox}{max width=\columnwidth}
\begin{tabular}{ll c lc r}
\toprule
Model & Scalar baseline & AvgQA & Best VQ & AvgQA & $\Delta$ \\
\midrule
Phi-4-mini   & GSQ          & 52.6 & YAQA & {63.3} & {+10.7} \\
Phi-4-mini   & BitDistiller & 53.0 & YAQA & {63.3} & {+10.3} \\
Llama-3.1-8B & GSQ          & 57.0 & YAQA & {65.1} & {+8.1} \\
Llama-3.1-8B & BitDistiller & 60.3 & YAQA & {65.1} & {+4.8} \\
Qwen3-MoE    & BitDistiller & {55.9} & {YAQA} & {66.0} & {+10.1} \\
\bottomrule
\end{tabular}
\end{adjustbox}
\end{table}

\subsection{Best calibration recipes}
\label{app:best-recipes}

This appendix reports, for every method and model, the \emph{best} 2-bit
checkpoint over its calibration recipes, selecting QA and generation
independently, and records which recipe was selected in each case.{ VPTQ and YAQA are selected from the same recipes.}

\begin{table*}[t]
\centering
\caption{\textbf{Cross-paradigm comparison at 2 bits.} This reproduces the
main-paper Table~\ref{tab:vq-architecture} with the per-metric best-recipe
selection made explicit (Table~\ref{tab:app-recipe-key}). Best 2-bit
AvgQA (\subref{tab:app-vq-qa}) and AvgGen (\subref{tab:app-vq-gen}) per model,
each selecting the best calibration recipe for its metric independently
(Table~\ref{tab:app-recipe-key}). GSQ is scalar PTQ with layer-wise discrete
optimization; BitDistiller is full-model QAT with self-distillation; {EfficientQAT (EQAT) is block-wise and end-to-end scalar QAT;} VPTQ
and YAQA are vector PTQ. Bold
marks the best per model; {$^{*}$GSQ without rotation.}}
\label{tab:app-vq-architecture}
\begin{subtable}[t]{0.49\textwidth}
\centering
\caption{QA (AvgQA $\uparrow$).}
\label{tab:app-vq-qa}
\small
\setlength{\tabcolsep}{3pt}
\begin{adjustbox}{max width=\linewidth}
\begin{tabular}{l c ccc cc}
\toprule
& & \multicolumn{3}{c}{Trained scalar} & \multicolumn{2}{c}{Vector PTQ} \\
\cmidrule(lr){3-5}\cmidrule(lr){6-7}
Model & \textsc{bf16} & GSQ & BitDist. & {EQAT} & VPTQ & YAQA \\
\midrule
Phi-4-mini    & 71.0 & 52.6 & 53.0 & {57.3} & {57.6} & \textbf{63.3} \\
Llama-3.1-8B  & 73.1 & 57.0 & 60.3 & {62.4} & {63.3} & {\textbf{65.1}} \\
Qwen3-MoE     & 72.6 & {65.1$^{*}$} & 55.9 & {\textbf{69.0}} & {64.8} & {66.0} \\
OLMoE-1B-7B   & {68.1} & {61.0$^{*}$} & 49.1 & {56.2} & {63.0} & {\textbf{64.0}} \\
\bottomrule
\end{tabular}
\end{adjustbox}
\end{subtable}
\hfill
\begin{subtable}[t]{0.49\textwidth}
\centering
\caption{Generation (AvgGen $\uparrow$).}
\label{tab:app-vq-gen}
\small
\setlength{\tabcolsep}{3pt}
\begin{adjustbox}{max width=\linewidth}
\begin{tabular}{l c ccc cc}
\toprule
& & \multicolumn{3}{c}{Trained scalar} & \multicolumn{2}{c}{Vector PTQ} \\
\cmidrule(lr){3-5}\cmidrule(lr){6-7}
Model & \textsc{bf16} & GSQ & BitDist. & {EQAT} & VPTQ & YAQA \\
\midrule
Phi-4-mini    & 73.2 & 21.6 & 18.4 & {30.3} & {29.5} & \textbf{37.4} \\
Llama-3.1-8B  & 42.9 & 20.0 & {20.3} & {22.3} & {19.7} & \textbf{26.6} \\
Qwen3-MoE     & 61.8 & {50.4$^{*}$} & 45.7 & {59.0} & {53.0} & {\textbf{61.8}} \\
OLMoE-1B-7B   & 32.4 & {13.6$^{*}$} & 0.9  & {9.7} & {15.2} & {\textbf{20.6}} \\
\bottomrule
\end{tabular}
\end{adjustbox}
\end{subtable}
\end{table*}

\paragraph{Selection protocol.}
For every (model, method) pair we evaluate a set of calibration recipes. {VPTQ and YAQA are both selected from the five single domains (\texttt{web}, \texttt{instruction}, \texttt{reasoning}, \texttt{code}, \texttt{mixed}; Table~\ref{tab:app-vq-domain}), the recipes we ran for both methods, and from the reason+web mixture on OLMoE. YAQA's Hessian mixtures are reported separately (Tables~\ref{tab:calibration-main} and~\ref{tab:app-calibration-complete}), as are VPTQ's on the MoEs (Table~\ref{tab:rev-vptq-shared}). The trained scalar baselines are selected from their own data grids, which include a larger \texttt{Mixed-500k} corpus.} We then pick the recipe with the
highest AvgQA for Table~\ref{tab:app-vq-qa} and, separately, the highest AvgGen
for Table~\ref{tab:app-vq-gen}.

\begin{table}[htbp]
\centering
\caption{\textbf{Selected calibration recipe per method.} The recipe that
maximizes AvgQA and, separately, AvgGen in
Tables~\ref{tab:app-vq-qa}--\ref{tab:app-vq-gen}. Single words denote base
domains; fractions denote convex mixtures (e.g.\ \texttt{r/w\,.5/.5} is
$0.5$~reasoning\,$+\,0.5$~web). \texttt{Mixed-500k} is the enlarged
trained-baseline corpus. Dashes denote methods not run on a model. {EfficientQAT recipes are in Table~\ref{tab:rev-eqat}, the MoE GSQ runs in Table~\ref{tab:rev-gsq}, and the MoE VPTQ recipes in Table~\ref{tab:rev-vptq-shared}.}}
\label{tab:app-recipe-key}
\small
\setlength{\tabcolsep}{4pt}
\begin{adjustbox}{max width=\columnwidth}
\begin{tabular}{ll ll}
\toprule
Model & Method & Best-QA recipe & Best-Gen recipe \\
\midrule
\multirow{4}{*}{Phi-4-mini}
 & GSQ          & Mixed-500k     & Mixed-500k \\
 & BitDistiller & Mixed-500k     & Mixed-500k \\
 & VPTQ         & instruction    & code \\
 & YAQA         & web            & reasoning \\
\midrule
\multirow{4}{*}{Llama-3.1-8B}
 & GSQ          & Mixed-500k     & Mixed-500k \\
 & BitDistiller & mixture        & mixture \\
 & VPTQ         & {web}            & code \\
 & YAQA         & {web}     & reasoning \\
\midrule
\multirow{4}{*}{Qwen3-MoE}
 & {GSQ}    & {Mixed-500k} & {Mixed-500k} \\
 & BitDistiller & mixture        & mixture \\
 & VPTQ & {web, instruction (tie)} & code \\
 & YAQA         & {web, mixed (tie)} & {reasoning} \\
\midrule
\multirow{3}{*}{OLMoE-1B-7B}
 & {GSQ}    & {Mixed-500k} & {Mixed-500k} \\
 & VPTQ & {w$+$r}  & w$+$r \\
 & YAQA         & w$+$r          & w$+$r \\
\bottomrule
\end{tabular}
\end{adjustbox}
\end{table}

\paragraph{QA is recipe-stable; generation is recipe-sensitive.}
Across YAQA recipes the QA average spans only $3.9$--{$7.4$} points while the
generation average spans $13.5$--$25.2$ points
(Figure~\ref{fig:app-calib-gen}{; $13.5$--$26.7$ over the full sweep, Table~\ref{tab:calibration-main}}), so recipe choice moves generation far more
than QA. Consequently the QA-optimal recipe is consistently a broad, general
corpus, \texttt{web} or the $50/50$ reasoning/web mixture on the dense and MoE
models, whereas the generation-optimal recipe shifts toward task-aligned,
reasoning- and code-heavy data. On both dense models and Qwen3-MoE, VPTQ's
generation peaks under \texttt{code} calibration, and YAQA's generation peaks
under {\texttt{reasoning} calibration on all three models (and, among its Hessian mixtures, under the reasoning/web/instruction mixture RWI442 on
Qwen3-MoE)}. The QA winner and generation winner therefore rarely coincide.

\paragraph{A robust default.}
Because QA is nearly flat in the recipe and generation prefers reasoning-aligned
data, the $50/50$ reasoning/web mixture is a strong single-recipe default. It is
the QA optimum for YAQA on Llama-3.1-8B and Qwen3-MoE and stays within roughly a
point of the best QA elsewhere, while retaining most of the generation benefit
of reasoning-heavy calibration. This matches the main-paper recommendation to
treat calibration as a generation knob and to blend a small, task-aligned
Hessian mixture rather than tuning a single narrow domain. The trained scalar
baselines (GSQ, BitDistiller) instead depend on \emph{scale}. Their base-domain
checkpoints collapse to near-random AvgQA, and only the enlarged
\texttt{Mixed-500k} / domain-mixture corpora recover usable accuracy, which is
the recipe we report for them.

\subsection{Per-domain vector-quantizer comparison}
\label{app:vq-domains}

Table~\ref{tab:app-vq-domain} {compares VPTQ and YAQA} at each base calibration domain. YAQA wins {nine of ten} completed dense-model
comparisons {and four of five Qwen3-MoE domains}. Phi-4-mini YAQA reasoning and code completed after the older aggregate
table and reach AvgQA scores of 58.5 and 55.9, respectively. \paragraph{YAQA's advantage survives bit-matching.}
Because YAQA spends ${\sim}13\%$ more index bits than VPTQ ($2.26$ versus
$2.00$ bits/weight), we ran {bit-matched controls} on Phi-4-mini {with a $2^{18}$-entry VPTQ codebook (Table~\ref{tab:app-bitmatch}). Shared across all linear layers, it costs $2.26$ effective bits, equal to YAQA, and YAQA still leads clearly; stored per linear layer (${\approx}3.6$ effective bits), it narrows the gap but still trails YAQA.} The residual YAQA advantage reflects
its trellis representation and differing curvature,
not the bit budget{.}

\begin{table}[htbp]
\centering
\caption{\textbf{Bit-matched vector-quantizer control (Phi-4-mini).}
{VPTQ with a $2^{18}$-entry codebook, stored per linear layer (${\approx}3.6$ effective bits) or shared model-wide ($2.26$, equal to YAQA). YAQA leads in both cases}, so the trellis \emph{representation}, not bit count alone, explains the
residual gap (curvature and optimization also differ). AvgQA is the
seven-task mean; AvgGen is the mean of GSM8K and HumanEval (pass@1). Bold marks the best
per column.}
\label{tab:app-bitmatch}
\small
\setlength{\tabcolsep}{5pt}
\begin{tabular}{l cc cc}
\toprule
 & \multicolumn{2}{c}{web} & \multicolumn{2}{c}{reasoning} \\
\cmidrule(lr){2-3}\cmidrule(lr){4-5}
Method (bits/weight) & AvgQA & AvgGen & AvgQA & AvgGen \\
\midrule
VPTQ ($2.25${, per-linear, ${\approx}3.6$ eff.}) & $60.7$ & $\mathbf{15.4}$ & $53.6$ & $24.3$ \\
{VPTQ ($2.25$, model-wide, $2.26$ eff.)} & {$58.4$} & {$4.3$} & {$47.5$} & {$11.8$} \\
YAQA ($2.26$) & $\mathbf{63.3}$ & $12.2$ & $\mathbf{58.5}$ & $\mathbf{37.4}$ \\
\bottomrule
\end{tabular}
\end{table}

\begin{table}[htbp]
\centering
\caption{\textbf{AvgQA by vector quantizer and calibration domain.}
Bold marks the better method within each model/domain. {The five domains are the recipes run for both methods; Table~\ref{tab:vq-architecture} takes each method's best over them. ``Mixed'' is the fifth calibration domain of \S\ref{app:best-recipes}, not a Hessian mixture. ``Rotated'' marks VPTQ in YAQA's model-wide rotated frame (Appendix~\ref{app:rev-vptq-rot}).}}
\label{tab:app-vq-domain}
\small
\setlength{\tabcolsep}{4pt}

\begin{tabular}{ll ccccc}
\toprule
Model & Method & Web & Instruct & Reason & Code & {Mixed} \\
\midrule
\multirow{3}{*}{Phi-4-mini}
 & VPTQ & {57.4} & {\textbf{57.6}} & {47.8} & {51.0} & {56.2} \\
 & {VPTQ, rotated} & {51.5} & {53.9} & {41.0} & {44.4} & {47.8} \\
 & YAQA & \textbf{63.3} & {56.8} & \textbf{58.5} & \textbf{55.9} & \textbf{58.8} \\
\midrule
\multirow{3}{*}{Llama-3.1-8B}
 & VPTQ & {63.3} & {62.4} & {57.8} & {57.6} & {62.0} \\
 & {VPTQ, rotated} & {57.6} & {59.8} & {45.7} & {53.4} & {54.5} \\
 & YAQA & \textbf{65.1} & \textbf{63.9} & \textbf{62.1} & \textbf{60.8} & \textbf{64.3} \\
\midrule
\multirow{2}{*}{Qwen3-30B-A3B}
 & VPTQ & {64.8} & {64.8} & {59.8} & {\textbf{63.8}} & {63.7} \\
 & YAQA & {\textbf{66.0}} & {\textbf{64.9}} & {\textbf{64.5}} & 62.8 & {\textbf{66.0}} \\
\bottomrule
\end{tabular}
\end{table}

\subsection{VPTQ and the model-wide rotation}
\label{app:rev-vptq-rot}

YAQA quantizes in a model-wide rotated frame (randomized Hadamard rotations of the residual stream with norm fusion) on top of its own incoherence processing; VPTQ's original code does not use a model-wide rotation. In YAQA's frame, VPTQ loses accuracy in every dense model--domain cell (Table~\ref{tab:app-vq-domain}) and in its best checkpoints (Table~\ref{tab:rev-vptq-rot}), so Table~\ref{tab:vq-architecture} uses VPTQ without it; YAQA's lead shrinks but every paired interval still excludes zero (Table~\ref{tab:rev-sig}). Removing the model-wide rotation from YAQA changes its AvgQA by $-0.4$ (Phi-4-mini, web). The MoE, Qwen3-8B, and bit-matched VPTQ runs use the rotated frame.

\begin{table}[htbp]
\centering
\caption{VPTQ with and without YAQA's model-wide rotation (best calibration recipe per metric, as in Table~\ref{tab:vq-architecture}; all rows scored with the same evaluation harness).}
\label{tab:rev-vptq-rot}
\scriptsize
\setlength{\tabcolsep}{4pt}
\begin{tabular}{ll cc cc}
\toprule
& & \multicolumn{2}{c}{Phi-4-mini} & \multicolumn{2}{c}{Llama-3.1-8B} \\
\cmidrule(lr){3-4}\cmidrule(lr){5-6}
Method & Model-wide rotation & AvgQA & AvgGen & AvgQA & AvgGen \\
\midrule
VPTQ & yes & $53.9$ & $11.1$ & $59.8$ & $16.1$ \\
VPTQ & no  & $57.6$ & $29.5$ & $63.3$ & $19.7$ \\
YAQA & yes & $\mathbf{63.3}$ & $\mathbf{37.4}$ & {$\mathbf{65.1}$} & $\mathbf{26.6}$ \\
\bottomrule
\end{tabular}
\end{table}

\subsection{Trained scalar baseline details}
\label{app:trained-baselines}

We additionally evaluate two methods that optimize quantized weights. They are
BitDistiller~\citep{du2024bitdistiller}, a group-128 QAT method with
self-distillation, and GSQ, a group-128 scalar method that learns discrete
assignments after GPTQ initialization. These methods consume training or layer-optimization compute
and are therefore not direct no-recovery PTQ baselines. We report only
task-complete checkpoints in Table~\ref{tab:app-trained-baselines}. Phi-4-mini
BitDistiller and GSQ are the mixed-500k checkpoints, consistent with the
headline Table~\ref{tab:vq-architecture}.

\paragraph{BitDistiller QAT configuration.}
Our BitDistiller runs apply asymmetric autoclip, convert every quantized
projection to a $2$-bit group-$128$ \textsc{QuantLinear} with a Hadamard input
rotation, and fine-tune the student with CAKLD self-distillation against the
frozen \textsc{bf16} teacher. Optimization uses AdamW at learning rate
$2\times10^{-5}$ with a cosine schedule, $10\%$ warmup, gradient accumulation
$8$, per-device batch size $1$ across an eight-GPU H100 node, a $1{,}024$-token
sequence length, gradient clipping at $1.0$, and one epoch over the calibration
mixture (seed $42$). Qwen3-30B-A3B is FSDP-sharded to fit optimizer state on a
single node; its reported step-$800$ checkpoint corresponds to $51{,}200$
training sequences (\S\ref{app:compute-accounting}).

\begin{table*}[t]
\centering
\caption{\textbf{Task-complete trained 2-bit baselines.}
BitDistiller and GSQ use 2-bit, group-128 scalar weights. Results use the same
task metrics as the main evaluation. ``Best complete'' selects over complete
calibration-domain checkpoints only. {For Llama-3.1-8B BitDistiller, the best-QA and best-generation checkpoints differ; Table~\ref{tab:vq-architecture} takes AvgQA from the first and AvgGen from the second.} MoE GSQ results are in Table~\ref{tab:rev-gsq}.}
\label{tab:app-trained-baselines}
\scriptsize
\setlength{\tabcolsep}{2.5pt}
\begin{adjustbox}{max width=\textwidth}
\begin{tabular}{lll ccccccccc cc}
\toprule
Model & Method & Domain & MMLU & WG & ARC-e & ARC-c & HS & LMB & PIQA
& GSM8K & HEval & AvgQA & AvgGen \\
\midrule
Phi-4-mini & BitDistiller & Mixed-500k
 & 39.2 & 59.8 & 66.3 & 39.7 & 51.9 & 45.0 & 69.3
 & 20.3 & 16.5 & 53.0 & 18.4 \\
Phi-4-mini & GSQ & Mixed-500k
 & 41.0 & 60.1 & 63.3 & 40.2 & 47.2 & 47.0 & 69.2
 & 31.5 & 11.6 & 52.6 & 21.6 \\
Llama-3.1-8B & BitDistiller & Mixture{, best QA}
 & 43.9 & 66.1 & 70.2 & 42.2 & 65.3 & 59.7 & 75.0
 & 9.8 & 22.6 & 60.3 & 16.2 \\
{Llama-3.1-8B} & {BitDistiller} & {Mixture, best gen.}
 & {42.8} & {64.4} & {67.5} & {39.4} & {65.5} & {59.1} & {73.4}
 & {16.8} & {23.8} & {58.9} & {20.3} \\
Llama-3.1-8B & GSQ & Mixed-500k
 & 41.1 & 65.4 & 67.6 & 38.6 & 57.3 & 56.6 & 72.4
 & 25.3 & 14.6 & 57.0 & 20.0 \\
Qwen3-30B-A3B & BitDistiller & Mixture
 & 48.8 & 57.6 & 62.8 & 42.0 & 62.8 & 43.6 & 73.8 & 52.3 & 39.0 & 55.9 & 45.7 \\
\bottomrule
\end{tabular}
\end{adjustbox}
\end{table*}

Calibration remains consequential for trained baselines
(Table~\ref{tab:app-trained-domain}). The strongest complete Llama
BitDistiller run is the later mixture checkpoint, which improves AvgQA by
4.2 points and HumanEval by 20.8 points over its web checkpoint. The five base
GSQ domains collapse to roughly $30$--$32$ AvgQA, but a larger mixed-500k GSQ
checkpoint recovers to 57.0 AvgQA and 14.6 HumanEval, which is the row we report
in Table~\ref{tab:app-trained-baselines}. We do not merge newer MMLU-only
retries with older task-complete checkpoints because doing so would create
synthetic rows from different model artifacts.

\begin{table}[htbp]
\centering
\caption{\textbf{Calibration-domain sensitivity of trained 2-bit baselines
on Llama-3.1-8B.} AvgQA and HumanEval. BitDistiller and GSQ report their five
base-domain checkpoints plus the larger mixed-500k recipe.}
\label{tab:app-trained-domain}
\small
\setlength{\tabcolsep}{4pt}
\begin{adjustbox}{max width=\columnwidth}
\begin{tabular}{ll cc}
\toprule
Method & Domain & AvgQA & HEval \\
\midrule
\multirow{6}{*}{BitDistiller}
 & Web       & 56.1 & 1.8 \\
 & Instruct  & 35.9 & 0.0 \\
 & Reasoning & 38.0 & 0.0 \\
 & Code      & 32.4 & 0.6 \\
 & Mixture   & \textbf{60.3} & \textbf{22.6} \\
 & Mixed-500k& 54.9 & 15.9 \\
\midrule
\multirow{6}{*}{GSQ}
 & Web       & 31.3 & 0.0 \\
 & Instruct  & 32.4 & 14.6 \\
 & Reasoning & 29.9 & 11.6 \\
 & Code      & 29.7 & 14.0 \\
 & Mixture   & 31.7 & 14.6 \\
 & Mixed-500k& \textbf{57.0} & \textbf{14.6} \\
\bottomrule
\end{tabular}
\end{adjustbox}
\end{table}

\subsection{EfficientQAT with matched data}
\label{app:rev-eqat}

\paragraph{Choice of trained scalar baseline.} Published EfficientQAT~\citep{chen2024efficientqat} and ParetoQ~\citep{liu2025paretoq} results use their own data, budgets, and protocols. ParetoQ's full-model QAT budget is far larger than the budgets compared here (Table~\ref{tab:app-compute-all}), whereas EfficientQAT's stages can use exactly our calibration and recovery data, so we re-run EfficientQAT under our protocol.

\paragraph{Setup.} We use the upstream EfficientQAT code and its 2-bit hyperparameters; only the data change. Block-AP draws from the YAQA calibration pools (web, math, or web+math) and, in one more arm, from the all-4 mixture of the strongest GSQ and BitDistiller entries; E2E-QP is skipped or uses the exact recovery sample of the crossed design ($3.3$M tokens), giving an EfficientQAT counterpart for every cell of Table~\ref{tab:rev-interaction}. EfficientQAT receives more calibration tokens than YAQA ($16.8$M token passes in Block-AP alone).

\begin{table}[!t]
\centering
\caption{EfficientQAT grid (P-bf16). \emph{Calib.} is the Block-AP data, where all-4 is the mixture of the best GSQ and BitDistiller entries. \emph{E2E} is the E2E-QP data, where none means Block-AP only. w2g128 is at or below YAQA's $2.26$ bits; w2g64 is above it.}
\label{tab:rev-eqat}
\scriptsize
\setlength{\tabcolsep}{3pt}
\begin{tabular}{lll l ccccc}
\toprule
Model & Format (bits) & Calib. & E2E & AvgQA & GSM8K & HEval & AvgGen & IFEval \\
\midrule
\multirow{15}{*}{Phi-4-mini} & w2g128 (${\le}2.25$) & web & none & 53.3 & 5.7 & 0.0 & 2.8 & 13.5 \\
 & w2g128 (${\le}2.25$) & web & web & 57.3 & 11.1 & 0.0 & 5.6 & 12.4 \\
 & w2g128 (${\le}2.25$) & web & math & 56.9 & 30.3 & 0.0 & 15.2 & 12.6 \\
 & w2g128 (${\le}2.25$) & math & none & 45.0 & 26.5 & 0.0 & 13.2 & 15.5 \\
 & w2g128 (${\le}2.25$) & math & web & 52.5 & 31.4 & 0.0 & 15.7 & 14.8 \\
 & w2g128 (${\le}2.25$) & math & math & 48.3 & 45.8 & 0.0 & 22.9 & 13.7 \\
 & w2g128 (${\le}2.25$) & all-4 & none & 50.9 & 22.9 & 13.4 & 18.2 & 18.1 \\
 & w2g128 (${\le}2.25$) & all-4 & web & 56.2 & 32.9 & 20.7 & 26.8 & 16.5 \\
 & w2g128 (${\le}2.25$) & all-4 & math & 54.0 & 44.2 & 16.5 & 30.3 & 14.2 \\
\cmidrule(lr){2-9}
 & w2g64 (${\le}2.50$) & web & none & 55.2 & 8.5 & 0.0 & 4.2 & 13.9 \\
 & w2g64 (${\le}2.50$) & web & web & 58.5 & 13.9 & 0.0 & 6.9 & 12.2 \\
 & w2g64 (${\le}2.50$) & web & math & 58.5 & 36.2 & 1.8 & 19.0 & 14.4 \\
 & w2g64 (${\le}2.50$) & math & none & 47.8 & 33.9 & 0.0 & 16.9 & 17.2 \\
 & w2g64 (${\le}2.50$) & math & web & 55.0 & 36.7 & 0.0 & 18.3 & 12.4 \\
 & w2g64 (${\le}2.50$) & math & math & 50.2 & 52.2 & 0.0 & 26.1 & 13.5 \\
\midrule
\multirow{15}{*}{Llama-3.1-8B} & w2g128 (${\le}2.25$) & web & none & 58.1 & 3.3 & 0.6 & 2.0 & 5.9 \\
 & w2g128 (${\le}2.25$) & web & web & 62.4 & 6.6 & 0.6 & 3.6 & 10.2 \\
 & w2g128 (${\le}2.25$) & web & math & 62.0 & 17.4 & 2.4 & 9.9 & 15.9 \\
 & w2g128 (${\le}2.25$) & math & none & 48.0 & 22.7 & 0.6 & 11.6 & 6.3 \\
 & w2g128 (${\le}2.25$) & math & web & 58.7 & 24.3 & 1.2 & 12.8 & 11.5 \\
 & w2g128 (${\le}2.25$) & math & math & 51.6 & 34.4 & 0.0 & 17.2 & 12.9 \\
 & w2g128 (${\le}2.25$) & all-4 & none & 55.3 & 21.8 & 15.2 & 18.5 & 7.9 \\
 & w2g128 (${\le}2.25$) & all-4 & web & 61.7 & 25.5 & 14.6 & 20.1 & 10.5 \\
 & w2g128 (${\le}2.25$) & all-4 & math & 59.2 & 30.5 & 14.0 & 22.3 & 15.7 \\
\cmidrule(lr){2-9}
 & w2g64 (${\le}2.50$) & web & none & 60.2 & 4.3 & 0.6 & 2.5 & 3.9 \\
 & w2g64 (${\le}2.50$) & web & web & 63.9 & 9.2 & 3.0 & 6.2 & 9.8 \\
 & w2g64 (${\le}2.50$) & web & math & 63.9 & 17.6 & 4.9 & 11.2 & 15.9 \\
 & w2g64 (${\le}2.50$) & math & none & 51.3 & 24.8 & 6.1 & 15.4 & 8.1 \\
 & w2g64 (${\le}2.50$) & math & web & 60.1 & 25.6 & 7.3 & 16.4 & 9.6 \\
 & w2g64 (${\le}2.50$) & math & math & 54.1 & 35.3 & 3.7 & 19.5 & 13.5 \\
\midrule
\multirow{9}{*}{OLMoE-1B-7B} & w2g128 (${\le}2.25$) & web & none & 53.5 & 2.4 & 2.4 & 2.4 & 12.9 \\
 & w2g128 (${\le}2.25$) & web & web & 56.2 & 3.3 & 0.6 & 2.0 & 14.4 \\
 & w2g128 (${\le}2.25$) & web & math & 55.9 & 5.9 & 1.2 & 3.6 & 14.4 \\
 & w2g128 (${\le}2.25$) & math+web & none & 51.3 & 7.9 & 0.0 & 3.9 & 15.0 \\
 & w2g128 (${\le}2.25$) & math+web & web & 54.8 & 6.5 & 0.0 & 3.3 & 16.8 \\
 & w2g128 (${\le}2.25$) & math+web & math & 53.1 & 11.4 & 0.0 & 5.7 & 15.3 \\
 & w2g128 (${\le}2.25$) & all-4 & none & 52.6 & 8.9 & 7.9 & 8.4 & 16.8 \\
 & w2g128 (${\le}2.25$) & all-4 & web & 55.9 & 8.3 & 4.9 & 6.6 & 16.3 \\
 & w2g128 (${\le}2.25$) & all-4 & math & 54.3 & 10.3 & 9.1 & 9.7 & 15.7 \\

\midrule
\multirow{12}{*}{Qwen3-30B-A3B} & w2g128 (${\le}2.25$) & web & none & 64.9 & 49.4 & 7.3 & 28.3 & 17.7 \\
 & w2g128 (${\le}2.25$) & web & web & \textbf{69.0} & 56.3 & 7.9 & 32.1 & 19.8 \\
 & w2g128 (${\le}2.25$) & web & math & 68.7 & 74.6 & 18.3 & 46.4 & 21.4 \\
 & w2g128 (${\le}2.25$) & math & none & 57.2 & 76.0 & 14.6 & 45.3 & 13.3 \\
 & w2g128 (${\le}2.25$) & math & web & 65.3 & 77.3 & 21.3 & 49.3 & 19.0 \\
 & w2g128 (${\le}2.25$) & math & math & 61.3 & 80.0 & 17.7 & 48.8 & 14.2 \\
 & w2g128 (${\le}2.25$) & math+web & none & 61.3 & 76.6 & 22.6 & 49.6 & 15.0 \\
 & w2g128 (${\le}2.25$) & math+web & web & 67.0 & 79.0 & 28.0 & 53.5 & 19.4 \\
 & w2g128 (${\le}2.25$) & math+web & math & 65.2 & 80.9 & 34.1 & 57.5 & 14.6 \\
 & w2g128 (${\le}2.25$) & all-4 & none & 62.4 & 73.1 & 40.2 & 56.7 & 18.5 \\
 & w2g128 (${\le}2.25$) & all-4 & web & 67.8 & 78.0 & 32.9 & 55.5 & 22.4 \\
 & w2g128 (${\le}2.25$) & all-4 & math & 66.6 & 78.9 & 39.0 & \textbf{59.0} & 20.3 \\
\bottomrule
\end{tabular}
\end{table}

\paragraph{Findings.} EfficientQAT avoids the scalar collapse and beats BitDistiller on every model (Table~\ref{tab:rev-eqat}). It trails YAQA on generation on every model (within noise on Qwen3-MoE) and on QA on the dense models and OLMoE; on Qwen3-MoE its best-QA checkpoint leads YAQA but keeps little generation (Table~\ref{tab:rev-sig}). Unlike YAQA, its best E2E-QP data does not flip with the calibration data on the dense models.

\paragraph{Representation $\times$ calibration.} Without recovery, switching from web to math calibration raises AvgGen more and costs less QA with YAQA than with EfficientQAT (w2g128). The paired-bootstrap interaction is $15.2$ [$11.8$, $18.6$] AvgGen on Phi-4-mini and $7.7$ [$4.6$, $10.9$] on Llama.

\subsection{GSQ on the MoEs}
\label{app:rev-checks}

GSQ starts from a rotated model (rotation and norm fusion). Evaluating the unquantized rotated models (Table~\ref{tab:rev-gsq}) shows that this step breaks OLMoE (random-floor QA before any quantization; most likely the pre-MoE norm fusion, Appendix~\ref{app:olmoe}). For a uniform protocol we run GSQ without rotation on both MoEs, on the checkpoints every other method uses, with the dense GSQ recipe unchanged. GSQ then does not collapse and trails YAQA on both MoEs (Tables~\ref{tab:vq-architecture}, \ref{tab:rev-sig}).

\begin{table}[htbp]
\centering
\caption{GSQ invariance check on OLMoE and no-rotation runs on both MoEs (core QA suite and GSM8K, P-bf16). ``Rotated'' is the unquantized model after GSQ's rotation and norm fusion, which must match \textsc{bf16} if the rotation is function-preserving. ``No rotation'' is GSQ without rotation, using the dense recipe.}
\label{tab:rev-gsq}
\scriptsize
\setlength{\tabcolsep}{4pt}
\begin{tabular}{ll cc}
\toprule
Model & Checkpoint & AvgQA & GSM8K \\
\midrule
OLMoE-1B-7B-Instruct & \textsc{bf16} & 68.1 & {33.7} \\
 & GSQ rotated, unquantized (transformers 5 / 4.53) & 29.5 / 29.5 & 0.0 / 0.0 \\
 & GSQ 2-bit, no rotation (Mixed-500k) & 61.0 & 13.8 \\
Qwen3-30B-A3B & GSQ 2-bit, no rotation (Mixed-500k) & 65.1 & 75.2 \\
\bottomrule
\end{tabular}
\end{table}

\subsection{Significance of the headline comparisons}
\label{app:rev-sigtests}

We compare YAQA with each baseline by a paired item-level bootstrap ($5{,}000$ resamples within each task; Table~\ref{tab:rev-sig}), each method entering with its best checkpoint for the metric. Checkpoints without per-item logs are re-scored under one protocol (P-bf16), with each baseline--YAQA pair evaluated using identical software and settings. Best-QA checkpoints are compared on AvgQA and best-generation checkpoints on AvgGen; re-scored values match Table~\ref{tab:vq-architecture} within $1.3$ points, except Qwen3-MoE VPTQ generation ($55.8$ vs.\ $53.0$). YAQA leads every baseline with intervals that exclude zero, except EfficientQAT on Qwen3-MoE, which leads on QA and is tied with YAQA on generation ($+3.2$ [$-0.8$, $7.5$], $p=0.13$).

\section{\texorpdfstring{{Calibration data and its interaction with recovery}}{Calibration data and its interaction with recovery}}
\label{app:sec-calib}

\subsection{Complete cross-model calibration sweep}
\label{app:calibration-full}

Table~\ref{tab:app-calibration-complete} retains recipes evaluated on all
three main model families. We exclude model-specific mixtures from the
matrix to avoid interpreting missing runs as poor results. Over
task-complete rows, the QA/Gen ranges are {7.4}/26.7 for Phi-4-mini, 4.9/17.0
for Llama, and 3.9/13.5 for Qwen3-MoE.

\begin{table*}[t]
\centering
\caption{\textbf{Task-complete 2-bit YAQA calibration sweep.}
QA and generation means from rows with complete per-task evaluations.
Blank weight entries mean that the source is absent; dashes in result columns
mean that only aggregate AvgQA/HEval, not the full task row, was reported.
$\Delta_{\rm complete}$ excludes dashed result cells.}
\label{tab:app-calibration-complete}
\scriptsize
\setlength{\tabcolsep}{3pt}
\begin{tabular}{ll cccc cc cc cc}
\toprule
& & \multicolumn{4}{c}{Hessian mixture weight}
& \multicolumn{2}{c}{Phi-4-mini}
& \multicolumn{2}{c}{Llama-3.1-8B}
& \multicolumn{2}{c}{Qwen3-30B-A3B} \\
\cmidrule(lr){3-6}\cmidrule(lr){7-8}\cmidrule(lr){9-10}\cmidrule(lr){11-12}
Group & Recipe & Code & Web & Reason & Instr
& QA & Gen & QA & Gen & QA & Gen \\
\midrule
\multirow{4}{*}{Single}
 & Web       &   & 1 &   &   & {\textbf{63.3}} & 12.2 & 65.1 & 9.6 & 66.0 & 54.0 \\
 & Instruct  &   &   &   & 1 & 56.8 & 10.7 & 63.9 & 11.8 & 64.9 & 49.8 \\
 & Reasoning &   &   & 1 &   & 58.5 & 37.4 & 62.1 & \textbf{26.6} & 64.5 & 61.8 \\
 & Code      & 1 &   &   &   & 55.9 & 18.7 & 60.8 & 22.6 & 62.8 & 48.9 \\
\midrule
\multirow{8}{*}{Standard mix}
 & reason+web      &      & 0.5  & 0.5  &      & 62.6 & 24.7 & \textbf{65.7} & 17.8 & \textbf{66.7} & 56.5 \\
 & webreason       &      & 0.7  & 0.3  &      & 63.0 & 27.7 & 65.5 & 15.1 & 65.8 & 56.2 \\
 & web+instruct    &      & 0.6  &      & 0.4  & 58.7 & 14.6 & \textbf{65.7} & 16.5 & 66.6 & 51.4 \\
 & reason+instruct &      &      & 0.6  & 0.4  & 57.3 & 18.2 & 64.6 & 17.6 & 64.2 & 53.8 \\
 & code+web        & 0.6  & 0.4  &      &      & 58.0 & 18.2 & 64.3 & 21.6 & 65.2 & 59.4 \\
 & code+reason     & 0.6  &      & 0.4  &      & 56.2 & 19.4 & 61.6 & 23.5 & 63.9 & 53.7 \\
 & code+reason+web & 0.4  & 0.3  & 0.3  &      & 57.8 & 23.4 & 64.6 & 22.0 & 65.9 & 61.8 \\
 & all-4           & 0.25 & 0.25 & 0.25 & 0.25 & 58.8 & 19.9 & 65.2 & 20.1 & 66.0 & 59.3 \\
\midrule
Targeted
 & rwi442 & & 0.4 & 0.4 & 0.2 & 59.5 & 17.1 & 62.7 & 12.2 & 66.5 & \textbf{62.4} \\
\midrule
\multicolumn{6}{l}{$\Delta_{\rm complete}$}
& {7.4} & 26.7 & 4.9 & 17.0 & 3.9 & 13.5 \\
\bottomrule
\end{tabular}
\end{table*}

\subsection{Crossed calibration--recovery design and held-out test}
\label{app:rev-crossed}

\paragraph{Design.} We cross the calibration data of the 2-bit YAQA base (web data from FineWeb-Edu or math data from OpenR1-Math) with short codebook-only recovery on the same two sources, or no recovery. Each recovered cell uses $3{,}200\times1{,}024$-token recovery ($3.3$M tokens) with a 4-bit teacher and three recovery seeds. The interaction is (math$-$web recovery on the math base) minus (math$-$web recovery on the web base); a negative value means that the best recovery data flips with the calibration data. Table~\ref{tab:rev-interaction} summarizes Phi-4-mini and Llama-3.1-8B (Figure~\ref{fig:rev-interaction}); Tables~\ref{tab:rev-llama} and~\ref{tab:rev-q8b} give Llama and the held-out Qwen3-8B in full.

\begin{table}[htbp]
\centering
\caption{\textbf{The best recovery data depends on the calibration data} (2-bit YAQA; Figure~\ref{fig:rev-interaction}). Recovered cells show the mean${\pm}$standard deviation over three recovery seeds{; no-recovery cells are the bases of Table~\ref{tab:calibration-main}}. The interaction (AvgGen) is given with a $95\%$ hierarchical bootstrap interval over seeds and test items.}
\label{tab:rev-interaction}
\scriptsize
\setlength{\tabcolsep}{4pt}
\begin{tabular}{ll cc cc}
\toprule
& & \multicolumn{2}{c}{Phi-4-mini} & \multicolumn{2}{c}{Llama-3.1-8B} \\
\cmidrule(lr){3-4}\cmidrule(lr){5-6}
Calibration & Recovery & AvgQA & AvgGen & AvgQA & AvgGen \\
\midrule
web  & none & {$63.3$} & {$12.2$} & $65.1$ & $9.6$ \\
web  & web  & $63.3{\pm}0.1$ & $18.9{\pm}0.5$ & $66.0{\pm}0.1$ & $11.0{\pm}0.3$ \\
web  & math & $62.8{\pm}0.1$ & $24.0{\pm}0.4$ & $66.1{\pm}0.1$ & $14.1{\pm}0.3$ \\
math & none & {$58.5$} & {$37.4$} & {$62.1$} & {$26.6$} \\
math & web  & $60.0{\pm}0.1$ & $\mathbf{45.7}{\pm}0.1$ & $64.0{\pm}0.0$ & $\mathbf{30.3}{\pm}0.5$ \\
math & math & $58.7{\pm}0.1$ & $42.3{\pm}1.0$ & $62.5{\pm}0.0$ & $28.4{\pm}0.4$ \\
\midrule
\multicolumn{2}{l}{Interaction (AvgGen)} & \multicolumn{2}{c}{$-8.5$ {\tiny$[-12.6,-4.3]$}} & \multicolumn{2}{c}{$-5.0$ {\tiny$[-8.4,-2.3]$}} \\
\multicolumn{2}{l}{Qwen3-8B (held-out)} & \multicolumn{4}{c}{$-9.4$ {\tiny$[-13.3,-5.3]$}; all three predictions hold} \\
\bottomrule
\end{tabular}
\end{table}

\begin{table}[htbp]
\centering
\caption{Llama-3.1-8B crossed design (2-bit YAQA, P-bf16). Recovered rows show the mean${\pm}$standard deviation over recovery seeds $\{42,1,2\}$, with $3{,}200\times1{,}024$-token codebook-only recovery and a 4-bit teacher. The held-out tasks, which are not used for any selection, are MBPP (3-shot), HumanEval+ (0-shot), and MATH-500 (4-shot). Domain perplexity is measured on held-out FineWeb-Edu (web) and OpenR1-Math (math) text ($64\times1{,}024$ tokens each), where \textsc{bf16} scores $6.86$ and $2.70$.}
\label{tab:rev-llama}
\scriptsize
\setlength{\tabcolsep}{2.2pt}
\begin{adjustbox}{max width=\linewidth}
\begin{tabular}{ll ccccc ccc cc}
\toprule
& & \multicolumn{5}{c}{Main suite} & \multicolumn{3}{c}{Held-out} & \multicolumn{2}{c}{Perplexity} \\
\cmidrule(lr){3-7}\cmidrule(lr){8-10}\cmidrule(lr){11-12}
Calib. & Recovery & AvgQA & GSM8K & HEval & AvgGen & IFEval & MBPP & HE+ & MATH & web & math \\
\midrule
web & none & $65.1$ & $11.9$ & $7.3$ & $9.6$ & $9.1$ & $12.4$ & $7.3$ & $1.2$ & $9.37$ & $4.03$ \\
web & web & $66.0{\pm}0.1$ & $13.1{\pm}0.6$ & $8.9{\pm}0.7$ & $11.0{\pm}0.3$ & $7.3{\pm}0.3$ & $15.8{\pm}0.2$ & $8.5{\pm}0.6$ & $1.7{\pm}0.3$ & $8.58$ & $3.79$ \\
web & math & $66.1{\pm}0.1$ & $17.2{\pm}0.1$ & $11.0{\pm}0.6$ & $14.1{\pm}0.3$ & $7.4{\pm}0.8$ & $18.3{\pm}0.2$ & $10.4{\pm}0.6$ & $2.2{\pm}0.3$ & $8.88$ & $3.53$ \\
math & none & {$62.1$} & {$32.5$} & {$20.7$} & {$26.6$} & $9.2$ & $30.8$ & $18.9$ & $12.6$ & -- & -- \\
math & web & $64.0{\pm}0.0$ & $39.1{\pm}0.4$ & $21.5{\pm}0.7$ & $30.3{\pm}0.5$ & $8.1{\pm}0.5$ & $35.9{\pm}0.5$ & $19.9{\pm}0.4$ & $15.5{\pm}0.8$ & $11.22$ & $3.08$ \\
math & math & $62.5{\pm}0.0$ & $36.4{\pm}0.2$ & $20.3{\pm}0.9$ & $28.4{\pm}0.4$ & $9.1{\pm}0.6$ & $34.3{\pm}0.6$ & $18.1{\pm}0.9$ & $11.9{\pm}0.4$ & $13.82$ & $3.13$ \\
\bottomrule
\end{tabular}
\end{adjustbox}
\end{table}

\paragraph{Results.} Short recovery raises Llama AvgQA and AvgGen on every cell, and the best recovery data flips with the calibration data (Table~\ref{tab:rev-llama}). The interaction persists on held-out tasks (Table~\ref{tab:rev-heldout}) and in held-out math perplexity. Calibration remains the larger factor ($+16.8$ AvgGen, $95\%$ interval $[13.9, 19.2]$).

\paragraph{Mechanism (held-out domain perplexity).} On Phi-4-mini, web recovery lowers web perplexity far more on the math-calibrated base ($16.08\to12.52$) than on the web-calibrated base ($12.18\to10.62$). The benefit of a recovery source thus depends on what calibration already covered.

\paragraph{Held-out model (Qwen3-8B).} On a model not used to develop the study, we test three predictions. We predict a negative interaction, web recovery at least as good as math recovery on the math-calibrated base, and a larger interaction on HumanEval than on GSM8K. All three hold (Table~\ref{tab:rev-q8b}).

\begin{table}[htbp]
\centering
\caption{Held-out Qwen3-8B crossed design (2-bit YAQA; mean${\pm}$standard deviation over three recovery seeds).}
\label{tab:rev-q8b}
\scriptsize
\setlength{\tabcolsep}{4pt}
\begin{tabular}{ll ccccc}
\toprule
Calib. & Recovery & AvgQA & GSM8K & HEval & AvgGen & IFEval \\
\midrule
\multicolumn{2}{l}{\textsc{bf16}} & {71.0} & {87.3} & {64.0} & {75.7} & 25.0 \\
web  & none & 63.3 & 53.0 & 7.9 & 30.5 & 15.2 \\
web  & web  & $65.1{\pm}0.1$ & $56.2{\pm}0.4$ & $14.6{\pm}0.0$ & $35.4{\pm}0.2$ & $16.2{\pm}0.6$ \\
web  & math & $64.8{\pm}0.0$ & $58.2{\pm}0.5$ & $19.9{\pm}0.7$ & $39.0{\pm}0.2$ & $16.8{\pm}0.9$ \\
math & none & 60.7 & 77.9 & 20.1 & 49.0 & 16.5 \\
math & web  & $63.9{\pm}0.1$ & $78.8{\pm}0.1$ & $34.6{\pm}0.4$ & $56.7{\pm}0.1$ & $16.8{\pm}0.9$ \\
math & math & $62.3{\pm}0.0$ & $77.0{\pm}0.2$ & $24.8{\pm}0.7$ & $50.9{\pm}0.4$ & $15.0{\pm}0.5$ \\
\bottomrule
\end{tabular}
\end{table}

\subsection{Robustness of the interaction and data checks}
\label{app:rev-robust}

These checks test whether the interaction is an artifact of one benchmark, one data sample, or one quantization run, and document how the data sources relate to the benchmarks.

\paragraph{Evaluation hygiene.} Report generation under the deployment prompt and check benchmark overlap. GSM8K, which carries the largest calibration gains, shares no $13$-gram with any calibration or recovery source; MATH-500 and HumanEval overlap slightly with the math and code sources, and overlap-free re-scoring leaves the Phi-4-mini interaction unchanged (Table~\ref{tab:rev-decontam}). Completion-style IFEval falls on most recovered Phi-4-mini models, but under the chat template recovery can raise IFEval and HumanEval (Table~\ref{tab:rev-chat}).

\begin{table}[htbp]
\centering
\caption{Held-out generation interaction (never used for any selection), $95\%$ hierarchical bootstrap over recovery seeds and test items.}
\label{tab:rev-heldout}
\scriptsize
\setlength{\tabcolsep}{4pt}
\begin{tabular}{l ccc}
\toprule
Benchmark & Phi-4-mini & Llama-3.1-8B & Qwen3-8B \\
\midrule
MBPP           & $-5.6$ {\tiny$[-9.7,-1.5]$} & $-4.1$ {\tiny$[-7.7,-0.7]$} & $-6.8$ {\tiny$[-10.8,-2.7]$} \\
HumanEval+     & $-12.2$ {\tiny$[-19.7,-4.7]$} & $-3.7$ {\tiny$[-9.1,1.6]$} & $-13.0$ {\tiny$[-20.5,-6.1]$} \\
MATH-500       & $-2.7$ {\tiny$[-6.7,1.1]$} & $-4.1$ {\tiny$[-6.9,-1.3]$} & $+1.7$ {\tiny$[-2.5,5.9]$} \\
MBPP+MATH-500  & $-4.2$ {\tiny$[-7.2,-1.3]$} & $-4.1$ {\tiny$[-6.3,-1.7]$} & $-2.6$ {\tiny$[-5.6,0.5]$} \\
All three      & $-6.8$ {\tiny$[-10.1,-3.6]$} & $-4.0$ {\tiny$[-6.3,-1.6]$} & $-6.0$ {\tiny$[-9.2,-3.0]$} \\
\bottomrule
\end{tabular}
\end{table}

\paragraph{Independently quantized bases and disjoint recovery samples (Phi-4-mini).} We re-quantized each calibration arm twice from independent calibration draws and drew two further recovery samples per domain, disjoint from the original. The interaction keeps its sign across bases ($-9.6$ [$-15.8$, $-3.4$]; Table~\ref{tab:rev-requant}) and recovery samples (Table~\ref{tab:rev-draws}).

\begin{table}[htbp]
\centering
\caption{Interaction across independently quantized Phi-4-mini bases (2-bit YAQA, P-bf16). \emph{Original} is the base of Table~\ref{tab:rev-interaction} (mean of three recovery seeds), and replicates use seed 42. The effect is math minus web recovery (AvgGen).}
\label{tab:rev-requant}
\scriptsize
\setlength{\tabcolsep}{3.5pt}
\begin{tabular}{ll ccc ccc c}
\toprule
& & \multicolumn{3}{c}{web recovery} & \multicolumn{3}{c}{math recovery} & \\
\cmidrule(lr){3-5}\cmidrule(lr){6-8}
Calib. & Base & GSM8K & HEval & AvgGen & GSM8K & HEval & AvgGen & Effect \\
\midrule
web & original & 24.2 & 13.6 & 18.9 & 33.8 & 14.2 & 24.0 & $+5.1$ \\
web & replicate 1 & 29.9 & 8.5 & 19.2 & 42.3 & 6.7 & 24.5 & $+5.3$ \\
web & replicate 2 & 25.7 & 10.4 & 18.0 & 39.6 & 17.1 & 28.3 & $+10.3$ \\
math & original & 56.9 & 34.6 & 45.7 & 60.1 & 24.6 & 42.3 & $-3.4$ \\
math & replicate 1 & 57.3 & 33.5 & 45.4 & 61.0 & 27.4 & 44.2 & $-1.2$ \\
math & replicate 2 & 61.4 & 32.9 & 47.2 & 64.1 & 23.2 & 43.6 & $-3.6$ \\
\bottomrule
\end{tabular}
\end{table}

\begin{table}[htbp]
\centering
\caption{Interaction across disjoint recovery-data samples (Phi-4-mini, original bases, P-bf16; GSM8K / HumanEval per cell). The $+20$k and $+40$k samples skip $20$k/$40$k source rows and share no rows with the original (seed 42).}
\label{tab:rev-draws}
\scriptsize
\setlength{\tabcolsep}{4pt}
\begin{tabular}{l cccc c}
\toprule
Sample & web$\to$web & web$\to$math & math$\to$web & math$\to$math & Interaction \\
\midrule
original & 24.2 / 13.6 & 33.8 / 14.2 & 56.9 / 34.6 & 60.1 / 24.6 & $-8.5$ \\
$+20$k rows & 23.2 / 13.4 & 33.6 / 12.2 & 56.3 / 35.4 & 60.1 / 25.0 & $-7.9$ \\
$+40$k rows & 25.1 / 14.6 & 34.4 / 14.6 & 58.4 / 31.7 & 61.5 / 27.4 & $-5.3$ \\
\bottomrule
\end{tabular}
\end{table}

\begin{table}[htbp]
\centering
\caption{Share of evaluation items (\%) sharing at least one $13$-gram with the first $20{,}000$ rows of each calibration/recovery source (covers the $3{,}200$-example recovery samples and the $2{,}048$-sequence Hessian sets).}
\label{tab:rev-decontam}
\scriptsize
\setlength{\tabcolsep}{5pt}
\begin{tabular}{l ccc}
\toprule
Source & GSM8K (test) & HumanEval & MATH-500 \\
\midrule
web (FineWeb-Edu)             & 0.0 & 0.0 & 0.0 \\
math (OpenR1-Math-220k)       & 0.0 & 0.6 & 4.8 \\
code (Evol-Instruct-Code-80k) & 0.0 & 5.5 & 1.0 \\
instruction (Alpaca-cleaned)  & 0.0 & 2.4 & 0.6 \\
\bottomrule
\end{tabular}
\end{table}

\begin{table}[htbp]
\centering
\caption{Recovered checkpoints under the chat template (bfloat16; GSM8K 5-shot, instruction-style HumanEval pass@1, IFEval prompt-strict).}
\label{tab:rev-chat}
\scriptsize
\setlength{\tabcolsep}{5pt}
\begin{tabular}{ll ccc}
\toprule
Calib. & Recovery & GSM8K & HumanEval & IFEval \\
\midrule
\multicolumn{5}{l}{\textit{Phi-4-mini (YAQA), with \textsc{bf16} at 81.4 / 73.8 / 69.7}} \\
web  & none & 13.0 & 15.2 & 43.1 \\
web  & web  & 24.4 & 18.9 & 43.8 \\
web  & math & 36.5 & 21.3 & 44.2 \\
math & none & 55.0 & 34.1 & 40.3 \\
math & web  & 58.9 & 43.9 & 44.5 \\
math & math & 61.5 & 37.2 & 37.9 \\
\midrule
\multicolumn{5}{l}{\textit{Qwen3-30B-A3B (YAQA, web), with \textsc{bf16} at 91.1 / 92.7 / 82.6}} \\
web  & none & 58.0 & 75.0 & 72.6 \\
web  & web (long) & 62.4 & 76.8 & 74.7 \\
\bottomrule
\end{tabular}
\end{table}

\subsection{Significance of the calibration and recovery effects}
\label{app:rev-sig-calrec}

Table~\ref{tab:rev-sig-calrec} applies the same bootstrap to the largest calibration effects of Table~\ref{tab:calibration-main} and to every recovery gain of Table~\ref{tab:recovery-summary}, with both checkpoints of each pair scored on one stack. All generation effects and all calibration effects on QA are significant; the recovery QA gain on the Phi-4-mini web base is not. HumanEval varies most across evaluation stacks (e.g., rw64 +rec.\ $19.5$ here vs.\ $28.7$ in Table~\ref{tab:recovery-summary}), but the sign and significance of every generation gain are unchanged.

\begin{table}[htbp]
\centering
\caption{Paired item-level bootstrap ($5{,}000$ resamples within each task) of the calibration effects ($\Delta_{\rm full}$ in Table~\ref{tab:calibration-main}) and the recovery gains (+rec.\ minus its own base, Table~\ref{tab:recovery-summary}). Values are in points, with $95\%$ intervals in brackets. Parentheses give the difference in the main-text table.}
\label{tab:rev-sig-calrec}
\scriptsize
\setlength{\tabcolsep}{3.5pt}
\begin{tabular}{lll ll}
\toprule
Model & Effect & Pair & $\Delta$AvgQA & $\Delta$AvgGen \\
\midrule
Phi-4-mini & calibration, QA & web $-$ code & $7.3$ [$6.6$, $8.1$] ({$7.4$}) & \\
 & calibration, Gen & reasoning $-$ instruction & & $26.5$ [$23.1$, $30.0$] ($26.7$) \\
Llama-3.1-8B & calibration, QA & reason+web $-$ code & $4.9$ [$4.3$, $5.6$] ($4.9$) & \\
 & calibration, Gen & reasoning $-$ web & & $17.3$ [$14.5$, $20.5$] ($17.0$) \\
Qwen3-MoE & calibration, QA & reason+web $-$ code & $3.6$ [$2.9$, $4.3$] ($3.9$) & \\
 & calibration, Gen & RWI442 $-$ code & & $13.5$ [$9.3$, $18.0$] ($13.5$) \\
\midrule
Phi-4-mini & recovery & web & $-0.2$ [$-0.7$, $0.3$] ($0.5$) & $20.7$ [$17.5$, $23.9$] ($17.8$) \\
 & recovery & reasoning & $2.4$ [$1.9$, $3.0$] ($2.5$) & $11.8$ [$8.3$, $15.6$] ($12.2$) \\
 & recovery & reason\_web & $0.5$ [$0.1$, $1.0$] ($1.1$) & $15.4$ [$12.2$, $18.5$] ($14.5$) \\
 & recovery & webreason & $0.5$ [$0.0$, $1.0$] ($1.1$) & $12.4$ [$9.1$, $15.7$] ($10.5$) \\
 & recovery & rw64 & $1.7$ [$1.2$, $2.2$] ($2.0$) & $12.4$ [$9.5$, $15.3$] ($16.5$) \\
Llama-3.1-8B & recovery & reason\_web & $1.8$ [$1.4$, $2.2$] ($1.7$) & $4.8$ [$2.5$, $7.1$] ($5.0$) \\
 & recovery & webreason & $1.8$ [$1.4$, $2.2$] ($1.8$) & $6.8$ [$4.5$, $9.3$] ($6.8$) \\
 & recovery & all-4 & $1.7$ [$1.3$, $2.2$] ($1.6$) & $6.7$ [$4.3$, $9.1$] ($5.4$) \\
Qwen3-MoE & recovery & web & $1.6$ [$1.2$, $2.1$] ($1.8$) & $8.2$ [$4.8$, $11.7$] ($6.6$) \\
\bottomrule
\end{tabular}
\end{table}

\section{\texorpdfstring{{Recovery fine-tuning}}{Recovery fine-tuning}}
\label{app:sec-recovery}

\subsection{Codebook-only recovery ablations}
\label{app:recovery-ablation}

The codebook-only ablation uses the Phi-4-mini 2-bit YAQA \texttt{rw64}
checkpoint. The reference configuration is forward KL, $\tau=1$,
full-vocabulary distillation, a \textsc{bf16} teacher, and web recovery data.
Each block varies one factor. Because this ablation fixes a single
\texttt{rw64} base, its web/forward-KL reference (GSM8K $49.6$) differs from the
per-recipe web base in Table~\ref{tab:recovery-summary} (GSM8K $40.5$); the two
should be read within their own base, not compared row-for-row. The report does
not contain a top-$p$ run, so we do not include one. The final row is the best validated run, not an untested
combination of independently optimal choices. {Top-$k$ rows are compared with a full-vocabulary baseline on the same platform (\emph{same platform}; Appendix~\ref{app:rev-topk}).}

\begin{table*}[t]
\centering
\caption{\textbf{Codebook-only recovery ablations.} Per-task accuracy on
Phi-4-mini 2-bit YAQA. AvgQA is the mean of the seven QA tasks; GSM8K and HEval
are reported separately.}
\label{tab:app-recovery-ofat}
\scriptsize
\setlength{\tabcolsep}{2.5pt}
\begin{adjustbox}{max width=\textwidth}
\begin{tabular}{ll ccccccccc c}
\toprule
Axis & Setting & MMLU & WG & ARC-e & ARC-c & HS & LMB & PIQA & GSM8K & HEval & AvgQA \\
\midrule
\multicolumn{12}{l}{\textit{References}} \\
 & \textsc{bf16} ceiling
 & 68.7 & 71.9 & 80.3 & 58.9 & 72.7 & 66.0 & 78.3 & 81.9 & 64.6 & 71.0 \\
 & 2-bit YAQA, no recovery
 & 51.7 & 63.6 & 73.6 & 48.5 & 61.7 & 59.8 & 73.7 & 44.9 & 8.5 & 61.8 \\
\midrule
\multicolumn{12}{l}{\textit{Loss}} \\
 & Forward KL
 & 53.7 & 66.1 & 75.0 & 50.4 & 63.9 & 60.5 & 75.1 & 49.6 & 19.5 & 63.5 \\
 & Reverse KL
 & 53.6 & 65.7 & 74.7 & 50.4 & 63.8 & 60.1 & 74.9 & 50.2 & 20.1 & 63.3 \\
 & Cross-entropy
 & 53.7 & 65.8 & 75.0 & 50.5 & 63.9 & 60.5 & 74.9 & 49.7 & 20.1 & 63.5 \\
\midrule
\multicolumn{12}{l}{\textit{Teacher precision}} \\
 & \textsc{bf16}
 & 53.7 & 66.1 & 75.0 & 50.4 & 63.9 & 60.5 & 75.1 & 49.6 & 19.5 & 63.5 \\
 & 4-bit matched
 & 53.7 & 66.5 & 74.6 & 50.4 & 63.8 & 61.2 & 75.1 & 48.7 & 18.9 & 63.6 \\
\midrule
\multicolumn{12}{l}{\textit{Temperature}} \\
 & $\tau=0.5$
 & 53.4 & 66.4 & 75.0 & 50.7 & 63.8 & 60.3 & 74.8 & 48.4 & 19.5 & 63.5 \\
 & $\tau=1.0$
 & 53.7 & 66.1 & 75.0 & 50.4 & 63.9 & 60.5 & 75.1 & 49.6 & 19.5 & 63.5 \\
 & $\tau=2.0$
 & 53.7 & 66.1 & 74.8 & 50.3 & 63.5 & 60.1 & 74.9 & 50.2 & 19.5 & 63.3 \\
\midrule
\multicolumn{12}{l}{\textit{Logit support} {(top-$k$ rows and their same-platform full-vocabulary baseline; App.~\ref{app:rev-topk})}} \\
 & {Full vocab., same platform}
 & {53.1} & {67.0} & {75.1} & {50.5} & {64.1} & {56.3} & {74.2} & {50.6} & {17.1} & {62.9} \\
 & $k=128$
 & {53.0} & {66.6} & {75.1} & {50.5} & {64.0} & {56.6} & {74.4} & {48.8} & {17.1} & {62.9} \\
 & $k=512$
 & {53.0} & {66.1} & {74.9} & {50.6} & {64.1} & {56.4} & {74.8} & {50.0} & {15.9} & {62.8} \\
 & $k=1024$
 & {53.3} & {65.9} & {74.9} & {50.7} & {64.0} & {56.5} & {75.0} & {49.7} & {15.9} & {62.9} \\
 & Full vocabulary
 & 53.7 & 66.1 & 75.0 & 50.4 & 63.9 & 60.5 & 75.1 & 49.6 & 19.5 & 63.5 \\
\midrule
\multicolumn{12}{l}{\textit{Recovery data}} \\
 & Web
 & 53.7 & 66.1 & 75.0 & 50.4 & 63.9 & 60.5 & 75.1 & 49.6 & 19.5 & 63.5 \\
 & Instruct
 & 53.2 & 66.7 & 75.0 & 50.2 & 63.9 & 59.8 & 75.1 & 49.4 & 14.6 & 63.4 \\
 & Reasoning
 & 53.6 & 65.3 & 73.2 & 48.7 & 63.3 & 60.0 & 74.2 & 58.4 & 10.4 & 62.6 \\
 & Code
 & 53.4 & 66.0 & 73.0 & 49.4 & 63.4 & 62.0 & 74.5 & 53.3 & 21.3 & 63.1 \\
 & Mixture
 & 53.7 & 66.1 & 74.8 & 50.3 & 64.0 & 60.3 & 74.8 & 55.0 & 17.7 & 63.4 \\
\midrule
\rowcolor{gray!12}
\multicolumn{2}{l}{\textbf{Selected run with mixture data (best QA/generation balance)}}
 & 53.7 & 66.1 & 74.8 & 50.3
 & 64.0 & 60.3 & 74.8 & 55.0
 & 17.7 & \textbf{63.4} \\
\bottomrule
\end{tabular}
\end{adjustbox}
\end{table*}

\subsection{Recovery runs and data card}
\label{app:rev-provenance}

\begin{table}[htbp]
\centering
\caption{Base, data, and budget of every recovery result. All recovery is codebook-only (LoRA rows add rank-16, $\alpha{=}32$ adapters), forward KL at $\tau{=}1$ over the full vocabulary, cosine schedule, learning rate $10^{-4}$, and uses the final checkpoint (no checkpoint selection).}
\label{tab:rev-provenance}
\scriptsize
\setlength{\tabcolsep}{2.6pt}
\begin{tabular}{@{}>{\raggedright\arraybackslash\leavevmode}p{0.15\linewidth}>{\raggedright\arraybackslash\leavevmode}p{0.22\linewidth}>{\raggedright\arraybackslash\leavevmode}p{0.24\linewidth}>{\raggedright\arraybackslash\leavevmode}p{0.23\linewidth}>{\raggedright\arraybackslash\leavevmode}p{0.08\linewidth}@{}}
\toprule
Result & Base checkpoint & Recovery data & Budget & Teacher \\
\midrule
Table~\ref{tab:recovery-summary} (long) & each row's own calibration recipe & same mixture as the calibration (an equal four-way mixture from a 500k-row pool for Llama) & Phi ${\approx}360$M; Llama $2{,}600\times16\times4{,}096$ (${\approx}170$M); Qwen ${\approx}1.2$M tokens & \textsc{bf16} \\
Table~\ref{tab:app-recovery-ofat} (ablations) & Phi rw64 (math+web $0.6/0.4$) & web unless the data axis varies & $3{,}200\times1{,}024$ & \textsc{bf16} / 4-bit \\
Tables~\ref{tab:rev-interaction}, \ref{tab:rev-llama}, \ref{tab:rev-q8b} & web or math calibration & FineWeb-Edu or OpenR1-Math, $3$ seeds & $3{,}200\times1{,}024$ & 4-bit \\
Tables~\ref{tab:app-lora-recovery}, \ref{tab:app-lora-ofat} (LoRA) & webreason ($0.7/0.3$) for Phi, reason+web ($0.5/0.5$) for Llama & web unless the data axis varies & $3{,}200\times1{,}024$ & \textsc{bf16} / 4-bit \\
Table~\ref{tab:app-qwen-recovery} & Qwen3-MoE YAQA web & per row & per row & \textsc{bf16} \\
\bottomrule
\end{tabular}
\end{table}

\paragraph{Relation between Tables~\ref{tab:recovery-summary} and~\ref{tab:app-recovery-ofat}.} The two tables differ in base and budget, not in method. Each row of Table~\ref{tab:recovery-summary} recovers the base calibrated on its own recipe with long recovery, whereas Table~\ref{tab:app-recovery-ofat} varies one factor at a time from the fixed rw64 base with a budget two orders of magnitude smaller (Table~\ref{tab:rev-provenance}).

\paragraph{Data card.} Calibration and recovery draw on four public sources. Web data come from FineWeb-Edu (\texttt{CC-MAIN-2024-10}), math data from OpenR1-Math-220k, code data from Evol-Instruct-Code-80k, and instruction data from Alpaca-cleaned. We apply no filtering beyond each source's own, and format each source with a fixed per-source formatter (plain text for web, prompt--response text otherwise). The web and math calibration pools hold $50{,}000$ documents each, from which the Hessians use $128$--$2{,}048$ sequences per domain; convex mixtures weight the per-domain Hessians. Recovery samples are drawn with a fixed seed from the same sources, disjoint from the $64$-document validation split. The crossed design uses $3{,}200$ examples of $1{,}024$ tokens ($3.3$M tokens). None of the evaluation benchmarks is a calibration or recovery source.

\subsection{LoRA-augmented recovery}
\label{app:lora-recovery}

Adding LoRA~\citep{hu2022lora} adapters on top of codebook recovery increases
attainable generation accuracy, but also changes the trainable and stored
parameterization. We therefore report each checkpoint separately rather than
combining per-metric maxima. All HumanEval scores use adapters folded into the
materialized weights. On Phi-4-mini, AvgQA is stable across recovery data
($63.6$--$64.4$); the specialization is in generation, where reasoning gives the
strongest GSM8K (63.3) and code gives the strongest HumanEval (32.9).
The same pattern appears on Llama. AvgQA stays near {$67$--$68$ while reasoning maximizes GSM8K (36.4) and code maximizes HumanEval (29.3)}. {The recovered bases are listed in Table~\ref{tab:rev-provenance}.} The best folded Phi
checkpoint reaches 34.8 IFEval
prompt-strict accuracy; the evaluation record does not identify that
checkpoint's recovery-data row, so we report it only as a best validated
outcome rather than attaching it to Table~\ref{tab:app-lora-recovery}.

\begin{table*}[t]
\centering
\caption{\textbf{LoRA + codebook recovery by recovery dataset.}
Per-task accuracy after folding the adapters into each materialized
checkpoint. AvgQA averages the seven QA tasks.
Phi-4-mini aggregates cover all reported configurations.}
\label{tab:app-lora-recovery}
\scriptsize
\setlength{\tabcolsep}{2.5pt}
\begin{adjustbox}{max width=\textwidth}
\begin{tabular}{ll cccccccccc}
\toprule
Model & Recovery data & MMLU & WG & ARC-e & ARC-c & HS & LMB & PIQA
& GSM8K & AvgQA & HEval \\
\midrule
\multirow{5}{*}{Phi-4-mini}
 & Web        & 55.8 & 65.7 & 76.3 & 51.0 & 65.5 & 61.5 & 75.3 & 45.3 & \textbf{64.4} & 17.7 \\
 & Instruct   & 55.6 & 65.3 & 75.5 & 51.0 & 65.7 & 60.7 & 75.5 & 50.7 & 64.2 & 30.5 \\
 & Reasoning  & 54.8 & 64.3 & 76.3 & 50.0 & 65.1 & 60.5 & 74.4 & \textbf{63.3} & 63.6 & 19.5 \\
 & Code       & 55.1 & 64.6 & 73.7 & 50.4 & 65.4 & 61.8 & 75.0 & 52.5 & 63.7 & \textbf{32.9} \\
 & Mixture    & 56.0 & 64.6 & 75.0 & 50.9 & 65.0 & 60.5 & 75.2 & 56.9 & 63.9 & 29.3 \\
\midrule
\multirow{5}{*}{Llama-3.1-8B}
 & Web        & {54.2} & {69.8} & {76.7} & {48.6} & {74.2} & {69.1} & {78.5} & {27.4} & {67.3} & {15.9} \\
 & {Instruct}   & {54.2} & {69.8} & {77.6} & {48.5} & {74.9} & {68.6} & {78.9} & {31.2} & {67.5} & {22.0} \\
 & Reasoning  & {54.5} & {68.2} & {77.8} & {49.6} & 73.3 & 69.2 & {77.7} & {\textbf{36.4}} & {67.2} & {14.0} \\
 & Code       & {54.8} & {69.9} & {78.3} & {49.4} & {74.1} & {68.9} & {77.9} & {33.5} & {\textbf{67.6}} & {\textbf{29.3}} \\
 & Mixture    & {54.6} & {69.9} & {78.1} & {49.5} & {74.6} & {68.2} & {78.5} & {34.1} & {\textbf{67.6}} & {23.2} \\
\bottomrule
\end{tabular}
\end{adjustbox}
\end{table*}

Table~\ref{tab:app-lora-ofat} reports the remaining one-factor-at-a-time
ablation axes. On Phi-4-mini, changing the loss, temperature, or teacher
precision moves AvgQA by at most 0.4 points, while HumanEval ranges
from 15.9 to 21.3. This mirrors codebook-only recovery, where optimization details
are comparatively stable, whereas recovery-data composition drives the
largest task-specific changes. Llama shows the same stability, with AvgQA
within 0.3 points across every axis{ (Appendix~\ref{app:rev-topk} covers top-$k$)}.

\begin{table}[htbp]
\centering
\caption{\textbf{LoRA + codebook one-factor-at-a-time ablations.}
Aggregate results for the non-data axes. The baseline uses forward KL,
$\tau=1$, full-vocabulary logits, a \textsc{bf16} teacher, and web recovery
data. Dashes denote unavailable complete task aggregates.}
\label{tab:app-lora-ofat}
\small
\setlength{\tabcolsep}{3pt}
\begin{adjustbox}{max width=\columnwidth}
\begin{tabular}{ll cc cc}
\toprule
& & \multicolumn{2}{c}{Phi-4-mini} & \multicolumn{2}{c}{Llama-3.1-8B} \\
\cmidrule(lr){3-4}\cmidrule(lr){5-6}
Axis & Setting & AvgQA & HEval & AvgQA & HEval \\
\midrule
Baseline & Forward KL, $\tau=1$, full vocab & 64.4 & 17.7 & {67.3} & {15.9} \\
\midrule
\multirow{2}{*}{Loss}
 & Reverse KL       & 64.1 & \textbf{21.3} & {67.3} & {\textbf{18.3}} \\
 & Cross-entropy    & 64.4 & 19.5 & {67.3} & {17.1} \\
\midrule
\multirow{2}{*}{Temperature}
 & $\tau=0.5$       & 64.1 & 16.5 & {67.1} & {15.9} \\
 & $\tau=2$         & 64.0 & 18.9 & {\textbf{67.4}} & {17.7} \\
\midrule
\multirow{4}{*}{Logit support}
 & {Full vocab, same platform} & {63.8} & {16.5} & {67.3} & {15.9} \\
 & $k=128$          & {63.8} & 17.7 & {67.5} & {16.5} \\
 & $k=512$          & {63.7} & 17.7 & {67.5} & {15.9} \\
 & $k=1024$         & {63.7} & {18.9} & {67.4} & {17.7} \\
\midrule
Teacher & 4-bit matched & 64.3 & 15.9 & {67.3} & {15.9} \\
\bottomrule
\end{tabular}
\end{adjustbox}
\end{table}

\subsection{Top-\texorpdfstring{$k$}{k} distillation}
\label{app:rev-topk}

The top-$k$ rows of Tables~\ref{tab:app-recovery-ofat} and~\ref{tab:app-lora-ofat} apply top-$k$ logit truncation in the distillation loss (the training loss changes with $k$). They are compared with a full-vocabulary baseline on the same platform. Truncation moves AvgQA by at most $0.2$ point, so distillation details matter little compared with the recovery data.

\subsection{Qwen3-MoE recovery-data sweep}
\label{app:qwen-recovery}

We additionally run a separate 2-bit Qwen recovery fine-tuning pipeline. It is
not the headline VPTQ/YAQA model and does not include MMLU, GSM8K, or HumanEval, so its
five-task average, AvgQA5, is not directly comparable to the seven-task AvgQA used
elsewhere. Within that controlled
sweep, however, broad web and mixture data outperform narrow instruction,
reasoning, and code recovery by 16--20 points.

\begin{table}[htbp]
\centering
\caption{\textbf{Qwen3-MoE recovery fine-tuning data sweep.}
AvgQA5 averages ARC-Easy, ARC-Challenge, PIQA, HellaSwag, and WinoGrande;
LAMBADA is shown but excluded. MMLU, GSM8K, and HumanEval are not evaluated in this controlled sweep.}
\label{tab:app-qwen-recovery}
\small
\setlength{\tabcolsep}{4pt}

\begin{tabular}{l cccccc c}
\toprule
Data & ARC-e & ARC-c & PIQA & HS & WG & LMB & AvgQA5 \\
\midrule
Web         & 69.1 & 42.0 & 70.7 & 58.4 & 56.3 & 37.7 & \textbf{59.3} \\
Mixture     & 67.4 & 43.4 & 70.4 & 56.6 & 57.6 & 37.5 & 59.1 \\
Instruct    & 34.0 & 25.6 & 59.4 & 44.3 & 52.6 & 23.3 & 43.2 \\
Reasoning   & 29.4 & 25.3 & 55.3 & 45.1 & 54.9 & 17.9 & 42.0 \\
Code        & 31.1 & 23.9 & 54.8 & 37.5 & 51.3 & 12.9 & 39.7 \\
\bottomrule
\end{tabular}
\end{table}

\section{\texorpdfstring{{Additional models and MoE details}}{Additional models and MoE details}}
\label{app:sec-models}

\subsection{A third dense model (Qwen3-8B)}
\label{app:qwen3-8b}

We additionally evaluate Qwen3-8B~\citep{yang2025qwen3}, a third dense model, at 2 bits under the
reason+web calibration mixture. Consistent with the dense-model ordering in the
main paper, YAQA outperforms VPTQ on AvgQA by 2.9 points (63.7 versus 60.8), on
generation by 5.5 HumanEval points (7.9 versus 2.4), and on nine of the ten
evaluated tasks, with ARC-Easy the only exception. Only the reason+web
calibration mixture is currently available for this model.

\begin{table}[htbp]
\centering
\caption{\textbf{Qwen3-8B (dense) at 2 bits, reason+web calibration.}
Per-task accuracy; AvgQA averages the seven QA tasks. HumanEval
reports pass@1. Bold marks the better of VPTQ and YAQA per column.}
\label{tab:app-qwen3-8b}
\small
\setlength{\tabcolsep}{2.5pt}
\begin{adjustbox}{max width=\textwidth}
\begin{tabular}{l ccccccccc c}
\toprule
Method & MMLU & WG & ARC-e & ARC-c & HS & LMB & PIQA & GSM8K & AvgQA & HEval \\
\midrule
\textsc{bf16} & 75.0 & 67.8 & 80.9 & 56.4 & 74.9 & 64.6 & 77.7 & 87.3 & 71.0 & 64.0 \\
VPTQ & 56.7 & 62.4 & \textbf{75.7} & 46.2 & 59.6 & 52.0 & 73.0 & 64.4 & 60.8 & 2.4 \\
YAQA & \textbf{63.7} & \textbf{65.7} & 74.1 & \textbf{48.3} & \textbf{65.4} & \textbf{55.6} & \textbf{73.2} & \textbf{68.6} & \textbf{63.7} & \textbf{7.9} \\
\bottomrule
\end{tabular}
\end{adjustbox}
\end{table}

\subsection{A second sparse MoE}
\label{app:olmoe}

{All OLMoE results use OLMoE-1B-7B-0924-Instruct~\citep{muennighoff2024olmoe}, for the \textsc{bf16} reference and every quantized checkpoint.} An incomplete norm fusion initially made OLMoE appear to collapse. After
folding the pre-MoE RMSNorm into all 64 experts, the fp32 invariance check
passes and 2-bit VPTQ recovers approximately 95\% of the \textsc{bf16} AvgQA
score {(with per-expert codebooks; in budget, $92\%$)}. We include this result because it guards against attributing a graph
transformation bug to quantization. {The same failure mode affects GSQ's rotation on OLMoE (Appendix~\ref{app:rev-checks}).}

\begin{table}[htbp]
\centering
\caption{\textbf{OLMoE-1B-7B-Instruct at 2 bits.} The model has 64 non-fused
experts. {$^{\dagger}$Per-expert codebooks ($5.92$ effective bits); expert-shared VPTQ is in Table~\ref{tab:rev-vptq-shared}.}}
\label{tab:app-olmoe}
\small
\setlength{\tabcolsep}{4pt}
\begin{tabular}{l cccc}
\toprule
Configuration & AvgQA & GSM8K & HEval & PTB ppl \\
\midrule
\textsc{bf16} ceiling & {68.1} & 33.7 & 31.1 & 14.7 \\
VPTQ web{$^{\dagger}$}              & {65.0} & 18.2 & 15.9 & 18.6 \\
VPTQ web+reason{$^{\dagger}$}       & {64.6} & 22.7 & 20.1 & 18.4 \\
YAQA web+reason       & 64.0 & 26.0 & 15.2 & 19.3 \\
\bottomrule
\end{tabular}
\end{table}

\subsection{MoE VPTQ recipes}
\label{app:rev-bits}

Table~\ref{tab:rev-vptq-shared} reports every MoE VPTQ calibration recipe, including chat-template HumanEval.

\begin{table}[htbp]
\centering
\caption{MoE VPTQ by calibration recipe (P-bf16). HumanEval$^{c}$ uses the chat template; parentheses give retention relative to \textsc{bf16} under the same protocol, where \textsc{bf16} scores $92.7$ on Qwen3-MoE and $34.8$ on OLMoE.}
\label{tab:rev-vptq-shared}
\scriptsize
\setlength{\tabcolsep}{3.4pt}
\begin{tabular}{ll cccc}
\toprule
Model & Calibration & AvgQA & GSM8K & HumanEval & HumanEval$^{c}$ \\
\midrule
\multirow{8}{*}{Qwen3-MoE} & web                  & 64.8 & 69.5 & 25.6 & 64.6 (70\%) \\
 & instruction          & 64.8 & 67.5 & 19.5 & 67.7 (73\%) \\
 & code                 & 63.8 & 66.9 & 39.0 & 76.2 (82\%) \\
 & math                 & 59.8 & 72.2 & 17.7 & 59.8 (65\%) \\
 & all-4 mixture        & 63.7 & 75.1 & 18.3 & 76.2 (82\%) \\
 & web/math $0.7/0.3$   & 66.0 & 76.3 & 22.6 & 67.1 (72\%) \\
 & math+web $0.5/0.5$   & 65.0 & 74.4 & 24.4 & 64.6 (70\%) \\
 & web/mixed $0.7/0.3$  & 67.2 & 75.6 & 34.8 & 72.6 (78\%) \\
\midrule
\multirow{2}{*}{OLMoE-1B-7B} & web                  & 62.7 & 11.8 & 7.9  & 7.3 (21\%)  \\
 & math+web             & 63.0 & 18.7 & 11.6 & 12.2 (35\%) \\
\bottomrule
\end{tabular}
\end{table}

\subsection{MoE router precision}
\label{app:router}

Quantizing the Qwen3-MoE router does not explain the model's strong 2-bit
retention. Across five checkpoints, 2-bit routers stay within 0.2 points
of fp16 routers.

\begin{table}[htbp]
\centering
\caption{\textbf{Qwen3-MoE router-precision ablation.} Expert weights remain
at their vector-quantized 2-bit setting; only router precision changes. The
checkpoint labels are those in the router-ablation report and their absolute
scores differ from the headline domain sweep; only within-row precision
deltas should be compared. {The VPTQ rows use per-expert codebooks (above the 2-bit budget).}}
\label{tab:app-router}
\small
\setlength{\tabcolsep}{4pt}
\begin{tabular}{l cccc}
\toprule
Checkpoint & fp16 & 8-bit & 4-bit & 2-bit \\
\midrule
VPTQ instruction & 65.5 & 65.5 & 65.6 & 65.4 \\
VPTQ web         & 67.0 & 66.9 & 67.1 & 67.1 \\
VPTQ code        & 64.0 & 64.0 & 64.1 & 64.0 \\
YAQA reason+web  & 68.5 & 68.5 & 68.5 & 68.6 \\
YAQA RWI442      & 68.2 & 68.1 & 68.0 & 68.0 \\
\bottomrule
\end{tabular}
\end{table}

\section{\texorpdfstring{{Generation, reasoning, and further analyses}}{Generation, reasoning, and further analyses}}
\label{app:sec-gen}

\subsection{Additional analyses}
\label{app:analyses}

\begin{figure}[htbp]
\centering
\includegraphics[width=0.82\linewidth]{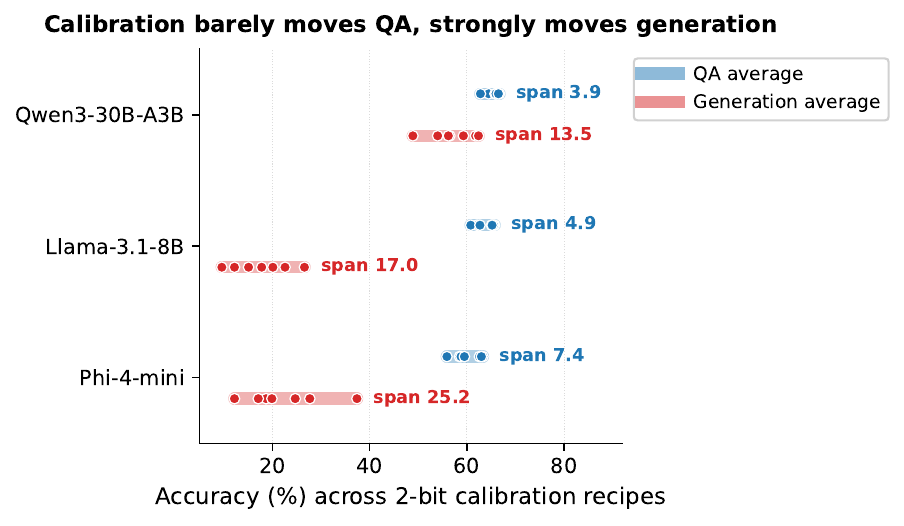}
\caption{\textbf{Calibration composition primarily moves generation, not QA.}
Across 2-bit YAQA calibration recipes, the QA average spans only $3.9$--{$7.4$}
points, whereas the generation average spans $13.5$--$25.2$ points {(plotted recipes; full sweep in Table~\ref{tab:calibration-main})}.}
\label{fig:app-calib-gen}
\end{figure}

\begin{figure}[htbp]
\centering
\includegraphics[width=0.82\linewidth]{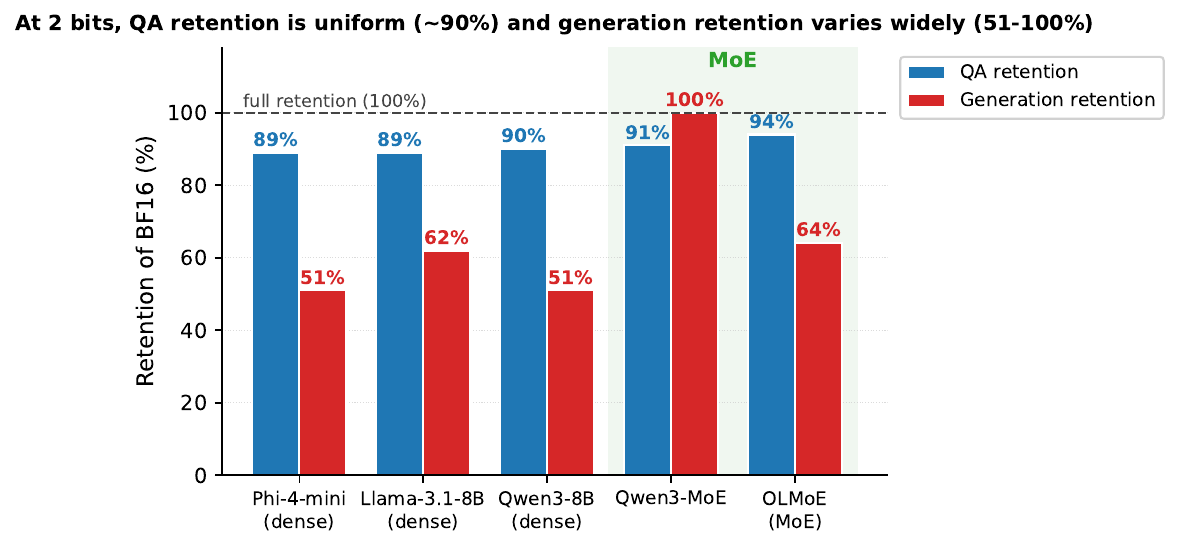}
\caption{\textbf{2-bit retention across models.} QA retention is
$\sim 90\%$ for every family {($89$--$94\%$)}, whereas generation retention varies widely ({$51$--$100\%$}).
Only Qwen3-MoE retains generation near-losslessly, while OLMoE ({$64\%$}) is closer to the dense models.
{Every bar uses the best vector checkpoint (YAQA; Table~\ref{tab:vq-architecture}).}}
\label{fig:app-retention}
\end{figure}

\begin{figure}[htbp]
\centering
\includegraphics[width=0.82\linewidth]{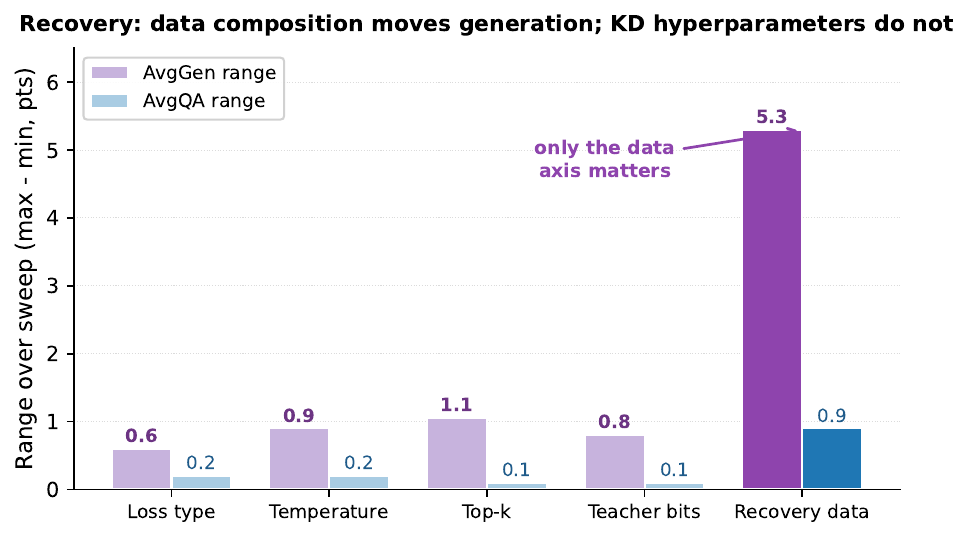}
\caption{\textbf{In recovery, data composition dominates; KD hyperparameters move it
little across the tested grid.} Loss type, temperature, top-$k$, and teacher precision each move AvgGen
by {$\le 1.1$} and AvgQA by $\le 0.2$; only the recovery-data axis moves
generation (AvgGen range $5.3$). {Top-$k$ bars use a same-platform baseline (Appendix~\ref{app:rev-topk}).}}
\label{fig:app-recovery-knobs}
\end{figure}

\begin{figure}[htbp]
\centering
\includegraphics[width=\linewidth]{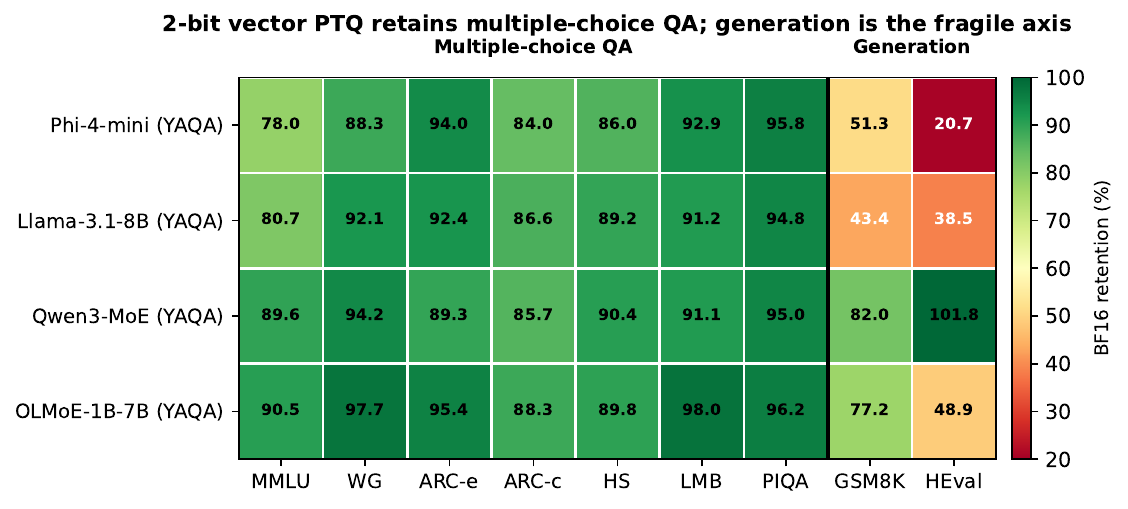}
\caption{\textbf{Which capabilities survive 2-bit vector PTQ.} Per-task
retention of {a 2-bit YAQA checkpoint per model (Phi-4-mini webreason, Qwen3-MoE web, Llama-3.1-8B and OLMoE reason+web)} relative to \textsc{bf16}.
Multiple-choice QA is largely retained ({$78$--$96\%$ on the dense models, $86$--$95\%$ on Qwen3-MoE, and $88$--$98\%$ on OLMoE}),
whereas generation (GSM8K, HumanEval) is the fragile axis. It falls to
$21$--$51\%$ on the dense models{, stays high on Qwen3-MoE ($82\%$ / $102\%$), and is mixed on OLMoE ($77\%$ / $49\%$).}}
\label{fig:app-pertask}
\end{figure}

\begin{figure}[htbp]
\centering
\includegraphics[width=\linewidth]{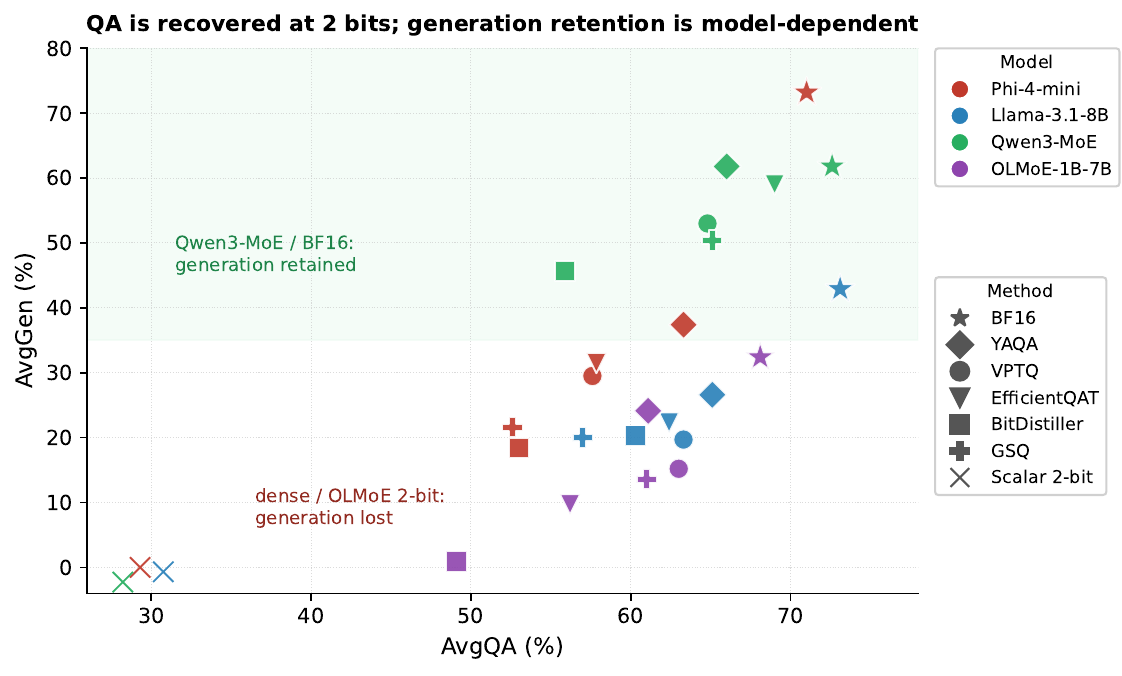}
\caption{\textbf{QA and generation are distinct axes.} AvgQA versus AvgGen for
every method and model at 2 bits {(Table~\ref{tab:vq-architecture}; scalar 2-bit from Table~\ref{tab:quantization-ladder})}. Every non-scalar method recovers QA, but
generation {separates Qwen3-MoE and \textsc{bf16} (high) from the dense models and OLMoE (low)}; scalar 2-bit collapses on both axes.
{MoE GSQ runs without rotation.}}
\label{fig:app-qa-gen}
\end{figure}

\begin{figure}[htbp]
\centering
\includegraphics[width=\linewidth]{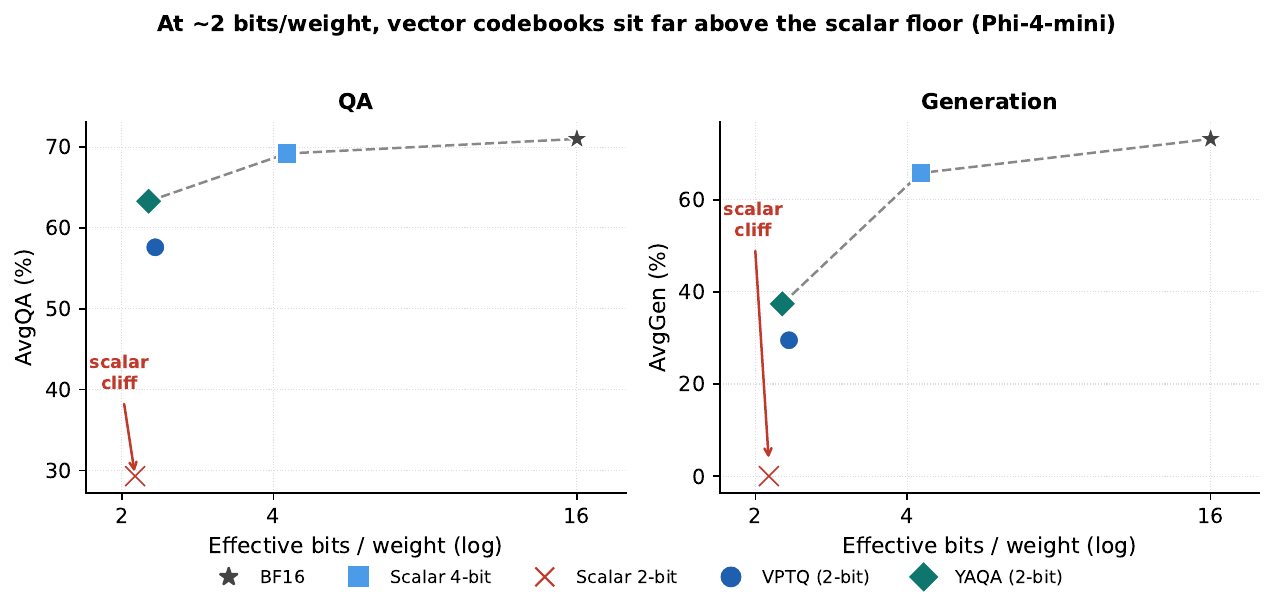}
\caption{\textbf{Representation, not bitwidth, sets the 2-bit frontier.}
Effective bits-per-weight {(indices, codebooks, and scales; Table~\ref{tab:app-method-configs})} versus AvgQA and AvgGen on Phi-4-mini. At $\sim 2$
bits/weight, a vector codebook (YAQA, $2.26$ b/w; VPTQ, {$2.33$} b/w) sits far
above the scalar 2-bit floor ({$2.125$} b/w) and near scalar 4-bit ({$4.25$} b/w), placing vector PTQ on the
accuracy--compression Pareto frontier. {Vector points are the best-QA and best-generation entries of Table~\ref{tab:vq-architecture}.}}
\label{fig:app-bpw}
\end{figure}

Figures~\ref{fig:app-calib-gen}--\ref{fig:app-recovery-knobs} distill three
cross-cutting trends from our 2-bit sweeps, and
Figures~\ref{fig:app-pertask}--\ref{fig:app-bpw} together with
Table~\ref{tab:app-scalar-cliff} add four analyses that reinforce the central
claims. They show that multiple-choice QA survives 2-bit vector PTQ while generation is the
fragile axis (Figure~\ref{fig:app-pertask}), that the two axes separate {Qwen3-MoE from the dense models and OLMoE}
(Figure~\ref{fig:app-qa-gen}), that vector PTQ lies on the
bits-per-weight Pareto frontier (Figure~\ref{fig:app-bpw}), and that the scalar
2-bit cliff is universal across rotation methods
(Table~\ref{tab:app-scalar-cliff}).

\begin{table}[htbp]
\centering
\caption{\textbf{The 2-bit scalar cliff is universal across rotation methods.}
Best-calibration-domain AvgQA at 16, 4, and 2 bits for two scalar PTQ
methods{ (Table~\ref{tab:quantization-ladder} uses web calibration only)}. Every method stays within $1$--$2$ points of \textsc{bf16} at 4 bits
and collapses to the ${\sim}29\%$ random-accuracy floor at 2 bits. QuaRot is
applied through expert-wise rotation on Qwen3-MoE.}
\label{tab:app-scalar-cliff}
\small
\setlength{\tabcolsep}{6pt}
\begin{tabular}{ll ccc}
\toprule
Model & Method & \textsc{bf16} & 4-bit & 2-bit \\
\midrule
\multirow{2}{*}{Phi-4-mini}
 & SpinQuant & 71.0 & 69.2 & 29.3 \\
 & QuaRot    & 71.0 & 69.3 & 29.8 \\
\midrule
\multirow{2}{*}{Llama-3.1-8B}
 & SpinQuant & 73.1 & 71.8 & 29.8 \\
 & QuaRot    & 73.1 & 71.7 & 29.3 \\
\midrule
\multirow{2}{*}{Qwen3-MoE}  & SpinQuant & 72.6 & 70.8 & 29.4 \\
 & QuaRot    & 72.6 & 70.3 & 29.3   \\
\bottomrule
\end{tabular}
\end{table}

\begin{figure}[htbp] \centering \begin{subfigure}{0.49\linewidth} \centering \includegraphics[width=\linewidth]{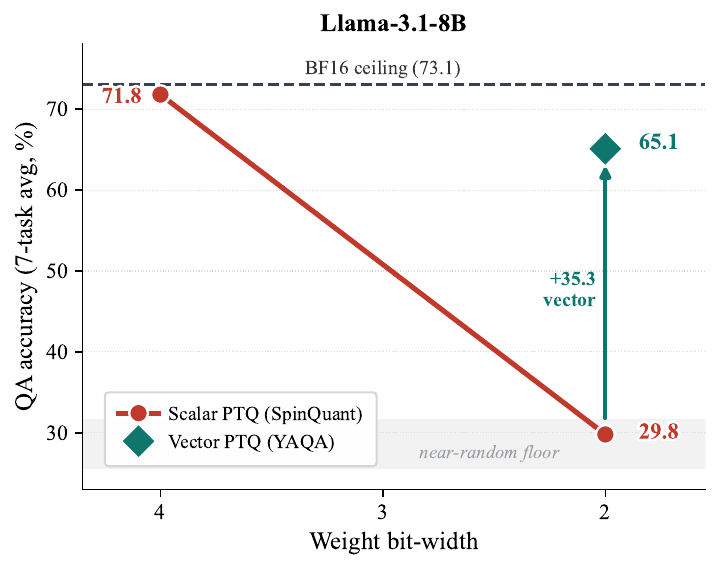} \caption{Llama-3.1-8B.} \end{subfigure} \hfill \begin{subfigure}{0.49\linewidth} \centering \includegraphics[width=\linewidth]{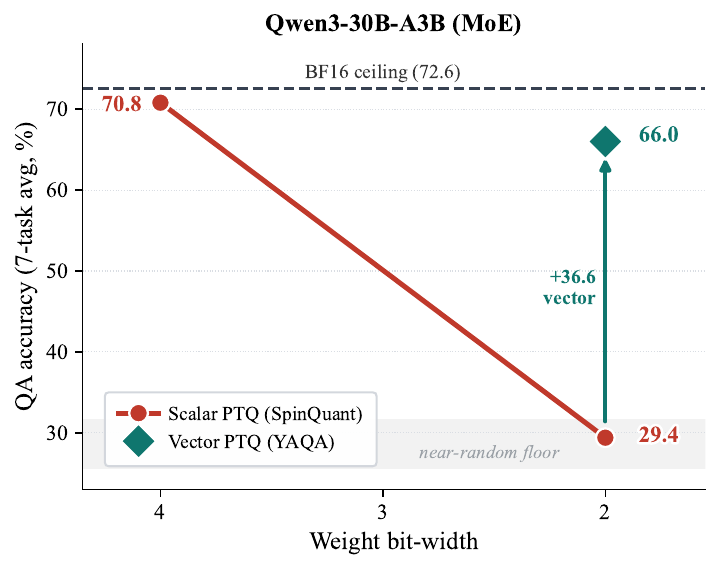} \caption{Qwen3-30B-A3B (MoE).} \end{subfigure} \caption{\textbf{The 2-bit scalar cliff on the other models} (analogs of Figure~\ref{fig:teaser}(a)). Scalar PTQ (SpinQuant) AvgQA collapses from near-lossless at 4 bits to the $\sim 29\%$ random-accuracy floor at 2 bits, whereas a 2-bit vector codebook (YAQA{, web calibration, on both models}) recovers most of the gap. AvgQA is the seven-task mean.} \label{fig:app-scalar-cliff-models} \end{figure}  \begin{table}[htbp]\centering\caption{\textbf{Qwen3-MoE 2-bit VPTQ domain-mixture sweep {(per-expert codebooks)}.} Convex-combination calibration mixtures of reasoning (r), web (w), and mixed (m) domains for a single 2-bit VPTQ Qwen3-30B-A3B checkpoint (one codebook per mixture). AvgQA is the seven-task mean; MMLU and GSM8K are 5-shot, HumanEval is pass@1, and IFEval is prompt-level strict. Calibration composition moves generation far more than QA (QA spans $2.7$ points; HumanEval spans $21$ points). {Per-expert codebooks ($7.23$ effective bits), so these rows show calibration sensitivity, not a 2-bit comparison; expert-shared results are in Table~\ref{tab:rev-vptq-shared}.} These finer convex mixtures lie \emph{outside} the common single-domain recipe grid selected in Tables~\ref{tab:vq-architecture} and~\ref{tab:app-vq-architecture}. We keep the headline tables restricted to the shared single-domain grid for a like-for-like cross-method comparison and report these finer mixtures here.}\label{tab:app-vptq-mixture-qwen}\small\setlength{\tabcolsep}{6pt}\begin{tabular}{l cccc}\toprule Calibration mixture & AvgQA & GSM8K & HumanEval & IFEval \\ \midrule reasoning .5 / web .5 & 67.0 & 82.0 & 31.7 & 23.8 \\ web .7 / reasoning .3 & 68.4 & 81.8 & \textbf{53.0} & 22.6 \\ web .5 / mixed .5 & 67.6 & \textbf{82.9} & 37.8 & 23.7 \\ web .7 / mixed .3 & \textbf{69.3} & 80.4 & 48.8 & 24.8 \\ reasoning .5 / mixed .5 & 66.6 & 82.1 & 44.5 & 23.7 \\ r .4 / w .4 / m .2 & 68.0 & 81.0 & 47.6 & 22.0 \\ r .2 / w .4 / m .4 & 68.3 & 80.4 & 43.3 & \textbf{27.0} \\ r .3 / w .4 / m .3 & 68.3 & 80.0 & 37.2 & 25.0 \\ \bottomrule \end{tabular}\end{table}

\subsection{Generation under the 2-bit cliff}
\label{app:gen-cliff}

\begin{figure}[htbp]
\centering
\begin{subfigure}{0.48\linewidth}
\centering
\includegraphics[width=\linewidth]{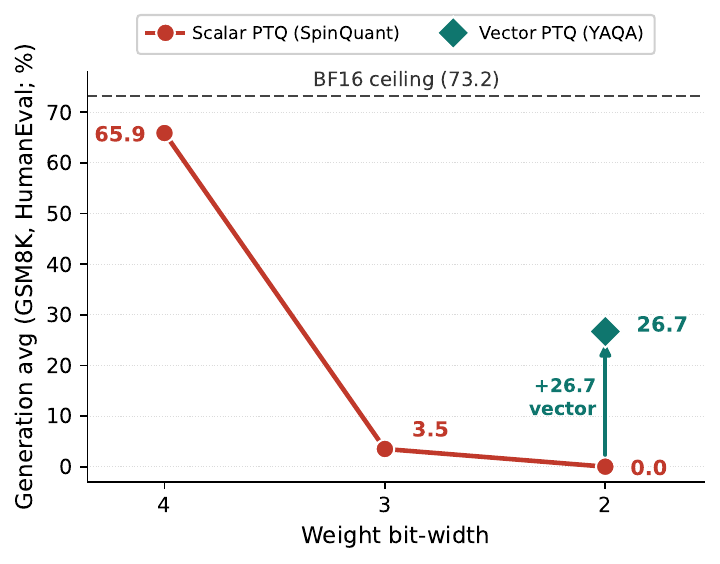}
\caption{Generation cliff and vector recovery.}
\label{fig:app-gen-a}
\end{subfigure}
\hfill
\begin{subfigure}{0.50\linewidth}
\centering
\includegraphics[width=\linewidth]{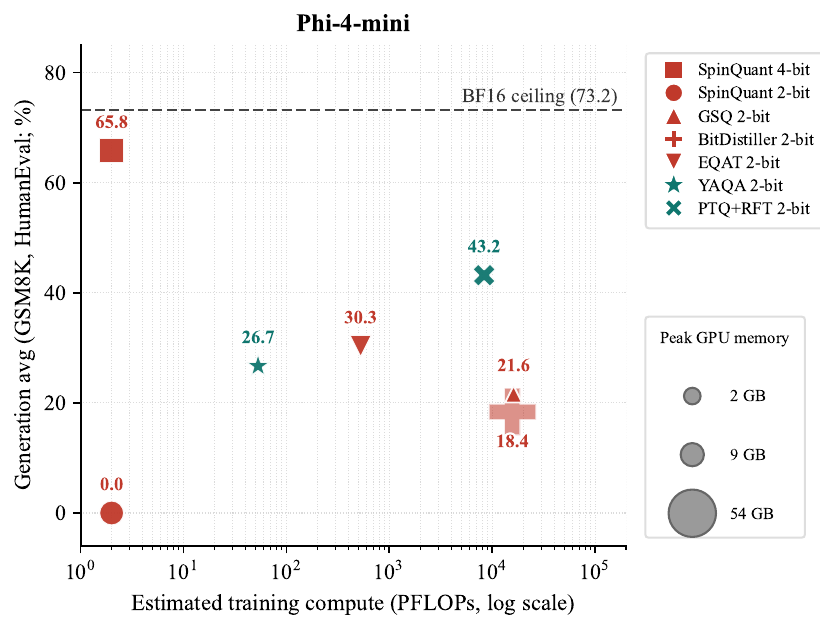}
\caption{Generation vs.\ estimated training compute at 2 bits.}
\label{fig:app-gen-b}
\end{subfigure}
\caption{\textbf{Generation mirrors the QA teaser.} (a) Scalar SpinQuant
generation collapses by 3-bit and vanishes at 2-bit; a vector codebook restores
$26.7$ AvgGen. (b) At a fixed 2-bit budget, data-centric recovery attains the
highest AvgGen {($43.2$) at about half the compute of the QAT baselines; EfficientQAT is labelled EQAT}. AvgGen is the mean of GSM8K and HumanEval on
Phi-4-mini; marker area is each algorithm's peak GPU memory {(Table~\ref{tab:app-deploy-all})}. Per-method {compute} in Table~\ref{tab:app-compute-all}.}
\label{fig:app-gen}
\end{figure}

Figure~\ref{fig:app-gen} reproduces the main-paper teaser for generation
(AvgGen, the mean of GSM8K and HumanEval) on Phi-4-mini. The scalar cliff is
even sharper than for QA. SpinQuant generation falls from $65.9$ at 4-bit to
$3.5$ at 3-bit and $0.0$ at 2-bit, and a vector codebook (YAQA, rw64 base)
restores $26.7$ at 2-bit (left). Under a fixed 2-bit budget, data-centric
recovery (PTQ+RFT on the 2-bit vector model, rw64 recommended checkpoint)
reaches $43.2$ AvgGen at a
fraction of the estimated compute of trained baselines such as GSQ
($21.6$) and BitDistiller ($18.4$) (right). These panels use the same checkpoints
and metric conventions as the QA teaser.

\subsection{Accuracy and generation vs.\ training compute (other models)}
\label{app:compute-figs}

\begin{figure}[htbp]
\centering
\begin{subfigure}{0.49\linewidth}
\centering
\includegraphics[width=\linewidth]{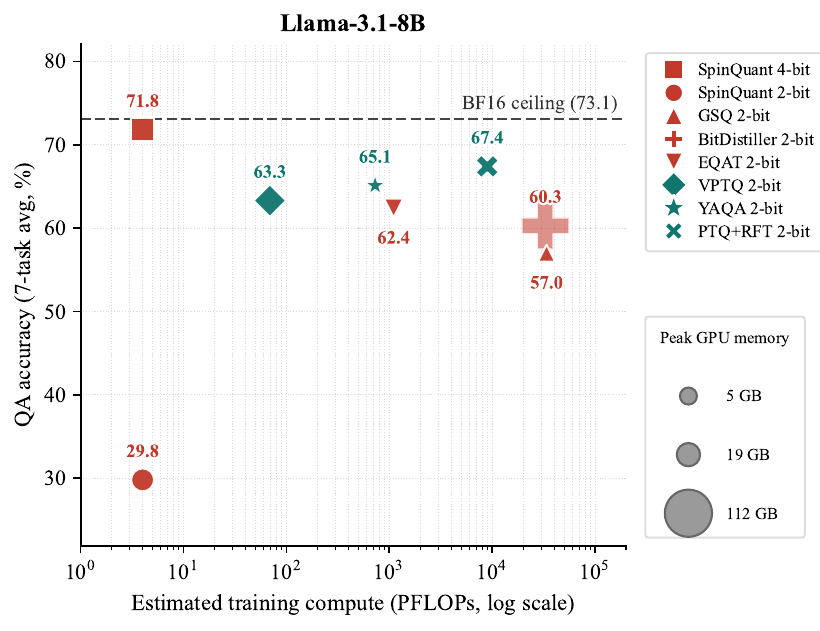}
\caption{AvgQA (7-task).}
\end{subfigure}
\hfill
\begin{subfigure}{0.49\linewidth}
\centering
\includegraphics[width=\linewidth]{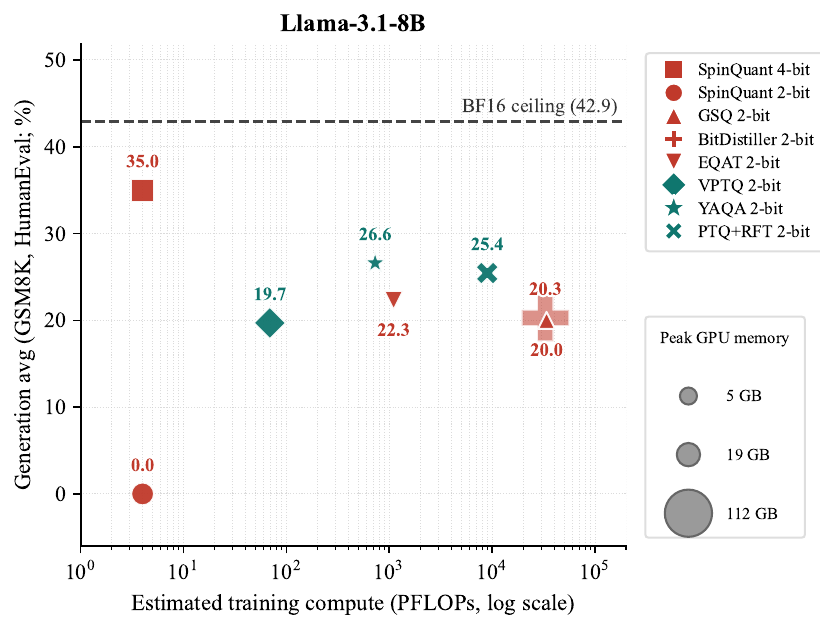}
\caption{AvgGen (GSM8K, HumanEval).}
\end{subfigure}
\caption{\textbf{Llama-3.1-8B accuracy and generation vs.\ training compute}
(analog of Figure~\ref{fig:teaser}(b)). Scalar $=$ crimson, vector $=$ teal;
$x$-axis is audited training compute (log scale); marker area is each algorithm's peak GPU memory {(Table~\ref{tab:app-deploy-all})}. Per-method {compute} in Table~\ref{tab:app-compute-all}{; accuracies from Tables~\ref{tab:vq-architecture} and~\ref{tab:recovery-summary}.}}
\label{fig:app-compute-llama}
\end{figure}

\begin{figure}[htbp]
\centering
\begin{subfigure}{0.49\linewidth}
\centering
\includegraphics[width=\linewidth]{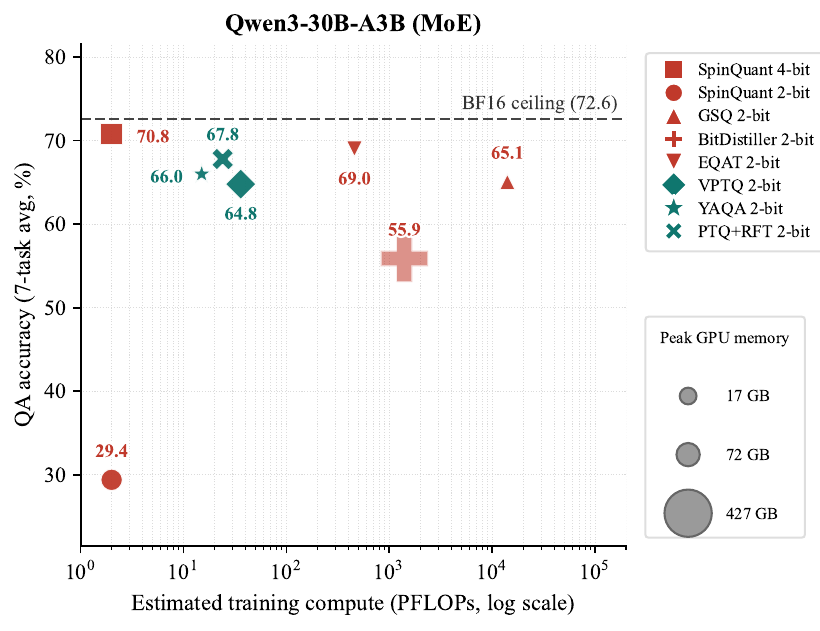}
\caption{AvgQA (7-task).}
\end{subfigure}
\hfill
\begin{subfigure}{0.49\linewidth}
\centering
\includegraphics[width=\linewidth]{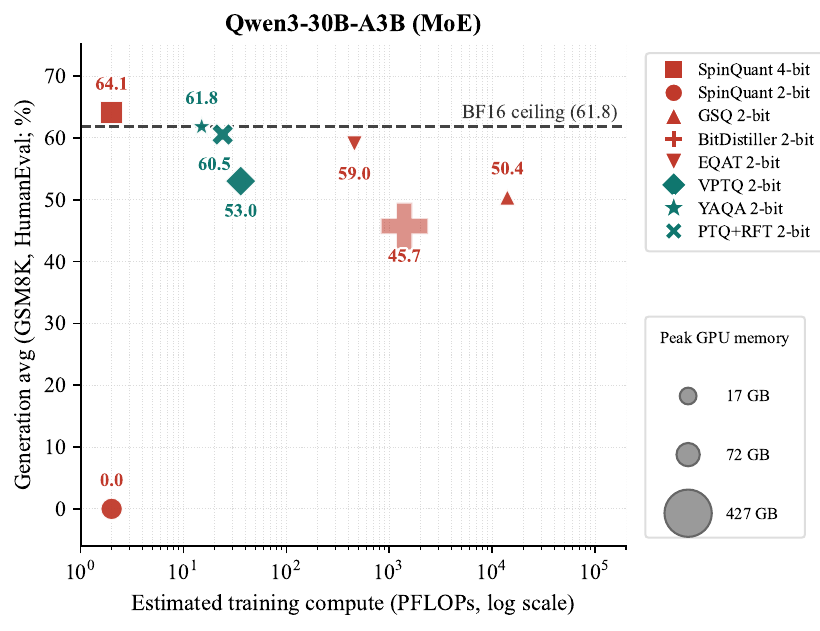}
\caption{AvgGen (GSM8K, HumanEval).}
\end{subfigure}
\caption{\textbf{Qwen3-30B-A3B (MoE) accuracy and generation vs.\ training
compute} (analog of Figure~\ref{fig:teaser}(b)). The MoE retains generation at 2
bits, unlike the dense models. Marker area is each algorithm's peak GPU memory {(Table~\ref{tab:app-deploy-all})}; per-method {compute} in Table~\ref{tab:app-compute-all}. {GSQ runs without rotation.}}
\label{fig:app-compute-qwen}
\end{figure}

Figures~\ref{fig:app-compute-llama} and~\ref{fig:app-compute-qwen} extend the
Figure~\ref{fig:teaser}(b) view to Llama-3.1-8B and the Qwen3-30B-A3B MoE. Each
2-bit method is plotted at its Table~\ref{tab:vq-architecture} score against its
audited training compute (Table~\ref{tab:app-compute-all}), with the 4-bit and
2-bit scalar ladder for reference. The QA panels reproduce the main ordering,
namely {YAQA} matches or beats the learning-based scalar
baselines (GSQ, BitDistiller) at $10^{2}$--$10^{3}\times$ less compute{, except EfficientQAT on Qwen3-MoE QA}. The
generation panels show the architecture split. The dense Llama loses most of
its \textsc{bf16} generation at 2 bits (best vector AvgGen $26.6$ versus a
\textsc{bf16} ceiling of $42.9$), whereas the Qwen3-MoE retains it (vector
AvgGen up to {$61.8$}, essentially lossless).

\subsection{Instruction following}
\label{app:ifeval}

IFEval is sensitive to whether the original checkpoint is instruction-tuned.
The Phi-4-mini instruct model has a meaningful ceiling and loses instruction
following at 2 bits; codebook-only recovery {does not restore it under completion-style prompts (Table~\ref{tab:recovery-summary}) but does under the chat template (Table~\ref{tab:rev-chat})}.
The base Llama checkpoint has a low IFEval ceiling, so its absolute values
should not be compared directly with Phi-4-mini. Across two independent MoE
families the 2-bit models remain usable instruction-followers. Qwen3-MoE holds
$24.4$ (from a \textsc{bf16} ceiling of $28.3$) and OLMoE-1B-7B-Instruct reaches
{$25.5$ at 2 bits (YAQA)}, so instruction following is not a casualty of the
2-bit transition.

\begin{table}[htbp]
\centering
\caption{\textbf{IFEval prompt-strict accuracy.} {Phi-4-mini YAQA uses reason+web calibration (Table~\ref{tab:recovery-summary}). $^{\dagger}$Per-expert codebooks, above the 2-bit budget.}}
\label{tab:app-ifeval}
\small
\begin{tabular}{ll c}
\toprule
Model & Configuration & IFEval \\
\midrule
Phi-4-mini-instruct & \textsc{bf16} & 56.6 \\
                    & 2-bit YAQA & 37.2 \\
                    & \quad + codebook recovery & {36.0} \\
\midrule
Llama-3.1-8B       & \textsc{bf16} & 8.9 \\
                    & 2-bit YAQA & 7.4 \\
\midrule
Qwen3-30B-A3B      & \textsc{bf16} & 28.3 \\
                    & 2-bit YAQA & 24.4 \\
                    & 2-bit VPTQ{$^{\dagger}$} & 15.3 \\
\midrule
OLMoE-1B-7B-Instruct & 2-bit YAQA & 25.5 \\
                    & 2-bit VPTQ{$^{\dagger}$} & 26.4 \\
\bottomrule
\end{tabular}
\end{table}

\AppNeedSpace{6\baselineskip}
\subsection{Overthinking under quantization}
\label{app:overthinking}

{The main study (all $262$ level-4--5 problems, $2.09$-bit checkpoint, every precision) is Figure~\ref{fig:app-overthinking-recovery} and Appendix~\ref{app:rev-math500}; this subsection reports a $40$-problem diagnostic run that also tracks hesitation markers and chain-of-thought length.}

We reproduce the overthinking failure mode of \citet{lotfi2026quantized} on our
own quantized checkpoints. On MATH-500~\citep{hendrycks2021math} (levels 4--5, greedy decoding) we compare
the \textsc{bf16} ceiling against the best 4-bit and best 2-bit checkpoint of a
reasoning model with an explicit thinking mode (Qwen3-30B-A3B; the best 2-bit is
the validated VPTQ \texttt{web\_mixed\_73} checkpoint{, which uses per-expert codebooks, $7.23$ effective bits}). Accuracy is exact match on
the final \verb|\boxed{}| answer; ``markers/1k'' counts hesitation and
branch-forking tokens (``Wait'', ``But'', ``Alternatively'', \dots) per $1{,}000$
generated tokens; the loop score is the fraction of repeated $4$-grams.

Table~\ref{tab:app-overthinking} shows that 2-bit quantization \emph{amplifies}
overthinking but does not collapse the model. Accuracy falls moderately
($0.675\to0.50$), while the hesitation-marker rate roughly doubles
($15.5\to32.2$ per $1$k), the loop score rises ($0.39\to0.58$), and the
length-cap hit rate grows ($25\to40\%$). {The repetition-penalty study, on all $262$ problems and at every precision, is in Figure~\ref{fig:app-overthinking-recovery} and Table~\ref{tab:rev-math500}.} The single-domain and
mixture checkpoints elsewhere in this paper are unaffected, because we always evaluated them on correctly rotated exports.

\begin{table}[htbp]
\centering
\caption{\textbf{Overthinking under 2-bit quantization} (Qwen3-30B-A3B,
MATH-500, greedy). 2-bit amplifies hesitation and looping and moderately lowers
accuracy, but does not collapse. ``Cap-hit'' is the length-cap hit rate;
``R-then-A'' is the reached-then-abandoned rate among wrong answers. {$40$ problems; 2-bit is the per-expert VPTQ checkpoint ($7.23$ effective bits).}}
\label{tab:app-overthinking}
\small
\setlength{\tabcolsep}{5pt}
\begin{tabular}{l cccccc}
\toprule
Precision & Accuracy & Mean CoT & Cap-hit & Markers/1k & Loop ($4$-gram) & R-then-A \\
\midrule
\textsc{bf16} & 0.675 & 4928 & 0.25 & 15.5 & 0.39 & 0.69 \\
4-bit         & 0.650 & 5265 & 0.25 & 12.4 & 0.41 & 0.50 \\
2-bit         & 0.500 & 5396 & 0.40 & 32.2 & 0.58 & 0.45 \\
\bottomrule
\end{tabular}
\end{table}

\paragraph{Qualitative examples.} Two representative MATH-500 generations from
the 2-bit Qwen3-MoE checkpoint {of the diagnostic run} illustrate the two regimes;
Figure~\ref{fig:app-overthinking-recovery} shows that a repetition penalty
{applied at both precisions narrows the gap to \textsc{bf16}}.

\begin{tcolorbox}[colback=green!3!white, colframe=green!45!black,
  title={\textbf{2-bit model, bounded and correct} (Level-4, gold $=\sqrt{5}$)},
  fonttitle=\bfseries, boxrule=0.5pt, fontupper=\small, left=5pt, right=5pt,
  top=3pt, bottom=3pt]
\emph{``I need the radius of a right cylindrical tank filled at $20\pi$~m$^3$/hr
while the level rises $4$~m/hr. Since $V=\pi r^2 h$, $\tfrac{dV}{dt}=\pi r^2
\tfrac{dh}{dt}$, so $20\pi=\pi r^2\cdot 4 \Rightarrow r^2=5 \Rightarrow
r=\sqrt{5}$.''} The model emits a bounded $1{,}282$-token chain (loop score
$0.09$) and terminates with the correct \texttt{\textbackslash boxed\{$\sqrt{5}$\}}.
\end{tcolorbox}

\begin{tcolorbox}[colback=red!3!white, colframe=red!45!black,
  title={\textbf{2-bit model, looping failure} (Level-5, gold $=6-5i$)},
  fonttitle=\bfseries, boxrule=0.5pt, fontupper=\small, left=5pt, right=5pt,
  top=3pt, bottom=3pt]
\emph{``I need to rotate $z$ around $c$ by $\pi/4$: $w=(z-c)e^{i\pi/4}+c$. Let me
compute $z-c$\dots''} The chain reasons coherently at first but then collapses
into a degenerate repetition (loop score $0.89$, $1{,}676$ hesitation markers),
filling the entire $8{,}192$-token budget \emph{without ever emitting a boxed
answer}, so it is scored wrong.
\end{tcolorbox}

\subsection{MATH-500 decoding study}
\label{app:rev-math500}

We repeat the overthinking study on all $262$ level-4--5 MATH-500 problems with the 2-bit ($2.09$-bit) Qwen3-MoE checkpoint, the best 4-bit checkpoint, and \textsc{bf16}, applying every decoding setting to every precision (greedy, $8{,}192$-token cap). A repetition penalty of $1.3$ helps both 2-bit ($+22.9$, $95\%$ interval $[17.2, 28.6]$) and \textsc{bf16}, so the 2-bit gap narrows but does not close (Table~\ref{tab:rev-math500}, Figure~\ref{fig:app-overthinking-recovery}).

\begin{table}[htbp]
\centering
\caption{MATH-500 levels 4--5 ($n{=}262$), Qwen3-30B-A3B, accuracy (\%) by decoding setting; loop is the repeated-4-gram fraction and cap the length-cap hit rate.}
\label{tab:rev-math500}
\scriptsize
\setlength{\tabcolsep}{3pt}
\begin{tabular}{l ccc ccc ccc}
\toprule
& \multicolumn{3}{c}{Greedy (no penalty)} & \multicolumn{3}{c}{Repetition penalty} & \multicolumn{3}{c}{Marker penalty} \\
\cmidrule(lr){2-4}\cmidrule(lr){5-7}\cmidrule(lr){8-10}
Precision & Acc. & Loop & Cap & 1.1 & 1.2 & 1.3 & 2 & 4 & 6 \\
\midrule
\textsc{bf16}         & 72.1 & 0.39 & 27\% & 76.0 & 77.1 & 77.1 & 77.1 & 79.0 & 79.0 \\
4-bit                 & 67.6 & 0.41 & 32\% & 69.8 & 72.1 & 72.1 & 73.7 & 74.0 & 77.9 \\
2-bit VPTQ ($2.09$ b) & 40.5 & 0.67 & 54\% & 56.1 & 60.3 & 63.4 & 51.1 & 52.7 & 48.9 \\
\bottomrule
\end{tabular}
\end{table}

\section{\texorpdfstring{{Limitations and future work}}{Limitations and future work}}
\label{app:limitations}

Our study compares 2-bit pipelines under one controlled protocol. The points below mark where its scope ends and where we see the most promising next steps.

\paragraph{Bit rates and compute.} Effective bits count indices, codebooks, and scales for every method (Table~\ref{tab:app-method-configs}), and YAQA keeps its lead at an exactly matched rate (Table~\ref{tab:app-bitmatch}). Calibration results select over several recipes per method, so we compare each method at the best recipe of its own grid. The GSQ, BitDistiller, and EfficientQAT comparisons are checkpoint comparisons, not compute-matched ones; the vector recipe already uses less training compute (Table~\ref{tab:app-compute-all}), and a sweep that gives every method the same budget is a natural extension.

\paragraph{Recovery.} Codebook-only recovery is a lightweight refinement that keeps the bit rate fixed; it complements rather than replaces full-precision training. Its gains are concentrated in generation. On the dense models it improves HumanEval substantially while AvgQA moves by about two points. A compute-matched ``more calibration data, no recovery'' control would isolate its QA contribution, and recovery for instruction following and on every architecture is a promising direction.

\paragraph{Reporting scope and conventions.}
{The study covers three primary models (Phi-4-mini, Llama-3.1-8B, Qwen3-30B-A3B) under the full protocol and two auxiliary models (OLMoE, Qwen3-8B) with narrower coverage. Bit rates include codebooks and scales (Table~\ref{tab:app-method-configs}); packed sizes also include retained high-precision tensors. Deployed packed weights, persistent training state, and per-device peak memory are distinct quantities, tabulated separately in Table~\ref{tab:app-deploy-all}. The decision guide (Table~\ref{tab:recipe}) is conditional on the deployment objective (QA, generation, or instruction following), the storage and compute budget, and low-bit kernel availability.}